\newif\ifICLR
\ICLRfalse % Plain/arXiv version; the ICLR submission is in main_ICLR.tex.

\ifICLR
  \documentclass{article}
  \usepackage{iclr2027_conference,times}
\else
  \documentclass[11pt]{article}
  \usepackage[margin=1in]{geometry}
\fi

\usepackage{amsmath,amssymb,amsthm,mathtools,bm}
\usepackage{algorithm}
\usepackage{algpseudocode}
\usepackage{booktabs}
\usepackage{tabularx}
\usepackage{array}
\usepackage{makecell}
\usepackage{enumitem}
\usepackage{graphicx}
\usepackage{subcaption}
\usepackage{titletoc}
\usepackage{placeins}
\usepackage{xcolor}
\usepackage{url}
\usepackage{xurl}
\usepackage{float}
\usepackage{afterpage}
\usepackage{aliascnt}
\usepackage{microtype}
\usepackage{tcolorbox} 
\usepackage{listings}
\usepackage{tikz}
\usetikzlibrary{arrows.meta,positioning,calc,fit,backgrounds}
\ifICLR
\else
  \usepackage[authoryear,round]{natbib}
\fi
\usepackage[
    colorlinks=true,
    linkcolor=blue,
    citecolor=blue,
    urlcolor=blue
]{hyperref}
\usepackage[
    capitalise,
    nameinlink,
    noabbrev
]{cleveref}

\definecolor{todored}{RGB}{190,30,45}
\definecolor{appendgreen}{RGB}{11,102,35}
\definecolor{attentionyellow}{RGB}{239, 183, 0}
\definecolor{morandiBlueFrame}{HTML}{4A7BB0} % greyblue
\definecolor{morandiBlueBg}{HTML}{EDF4FA}    % iceblue
\definecolor{directblue}{RGB}{43,103,167}
\definecolor{thinkingorange}{RGB}{224,112,54}
\definecolor{needlefill}{RGB}{245,194,66}
\definecolor{promptblue}{HTML}{0072B2}
\definecolor{thinkvermillion}{HTML}{D55E00}
\definecolor{ansgreen}{HTML}{009E73}

\newcolumntype{Y}{>{\raggedright\arraybackslash}X}

\newcommand\YZ[1]{\textcolor{teal}{[Yiqiao: #1]}}

\newcommand\LS[1]{\textcolor{red}{[Twist: #1]}}
\newcommand\TY[1]{\textcolor{blue}{[Tianyu: #1]}}

\usepackage[T1,OT1]{fontenc}
\tcbuselibrary{skins,breakable}
\definecolor{paperPromptInk}{HTML}{30312E}
\definecolor{paperPromptMuted}{HTML}{64665F}
\definecolor{paperPromptRule}{HTML}{DDDCD5}
\definecolor{paperPromptBorder}{HTML}{B1B0A7}
\definecolor{paperPromptFill}{HTML}{FAFAF7}
\definecolor{paperPromptHeader}{HTML}{F2F0E8}
\definecolor{paperPromptAccent}{HTML}{B87652}

\newtcolorbox{promptbox}[2][]{%
  enhanced,
  breakable,
  colback=paperPromptFill,
  colframe=paperPromptBorder,
  colbacktitle=paperPromptHeader,
  coltitle=paperPromptInk,
  coltext=paperPromptInk,
  boxrule=0.55pt,
  titlerule=0.4pt,
  arc=2pt,
  outer arc=2pt,
  boxsep=0pt,
  left=12pt,
  right=12pt,
  top=10pt,
  bottom=10pt,
  toptitle=8pt,
  bottomtitle=8pt,
  fonttitle=\normalfont\small,
  fontupper=\small\fontencoding{T1}\ttfamily\selectfont,
  before upper={\setlength{\parindent}{0pt}\setlength{\parskip}{3pt}\raggedright},
  before skip=10pt,
  after skip=8pt,
  pad at break*=3mm,
  title={\textcolor{paperPromptAccent}{\rule[-1pt]{2pt}{10pt}}\hspace{7pt}%
    \textbf{Prompt}\hspace{0.6em}\textcolor{paperPromptBorder}{\textbar}%
    \hspace{0.6em}#2},
  title after break={\textbf{Prompt}\hspace{0.6em}#2\hfill\textit{continued}},
  #1
}

\newcommand{\promptlabel}[1]{%
  \par{\normalfont\small\bfseries\color{paperPromptMuted}#1}\par\nobreak
}
\newcommand{\prompttag}[1]{\textcolor{paperPromptMuted}{\texttt{#1}}}
\newcommand{\promptvar}[1]{{\fontencoding{T1}\ttfamily\selectfont\{#1\}}}
\newcommand{\promptrecord}[1]{{\normalfont\bfseries #1}}
\newcommand{\promptdivider}{%
  \par\vspace{5pt}\noindent
  {\color{paperPromptRule}\rule{\linewidth}{0.4pt}}\par\nobreak\vspace{4pt}%
}

\theoremstyle{plain}

\newaliascnt{lemma}{theorem}

\aliascntresetthe{lemma}
\newaliascnt{proposition}{theorem}

\aliascntresetthe{proposition}
\newaliascnt{corollary}{theorem}

\aliascntresetthe{corollary}

\theoremstyle{definition}

\theoremstyle{remark}

\newcommand{\slcomment}[1]{%
  \textcolor{red}{\textbf{[SL: #1]}}%
}
\ifICLR
  \renewcommand{\YZ}[1]{}
  \renewcommand{\LS}[1]{}
  \renewcommand{\TY}[1]{}
  \renewcommand{\slcomment}[1]{}
\else
  \setlist[itemize]{leftmargin=*,topsep=3pt,itemsep=2pt}
  \setlist[enumerate]{leftmargin=*}
\fi
\graphicspath{{figures/}}
\title{Targeted Retrieval, Compact Representations:\\ How CoT Reasoning Improves Long-Context Counting}
\author{
Liang Twist Shan\textsuperscript{1,}\thanks{Correspondence to Liang Twist Shan (\href{mailto:lshan9@wisc.edu}{\texttt{lshan9@wisc.edu}}) and Yiqiao Zhong (\href{mailto:yiqiao.zhong@wisc.edu}{\texttt{yiqiao.zhong@wisc.edu}}).}
\qquad
Tianyu Hu\textsuperscript{2}
\qquad
Hao Yan\textsuperscript{1}
\qquad
Yiqiao Zhong\textsuperscript{1,2,}\footnotemark[2]
\\[0.75em]
{\small
\textsuperscript{1}Department of Statistics, University of Wisconsin-Madison
}\\
{\small
\textsuperscript{2}Department of Computer Sciences, University of Wisconsin-Madison
}\\[0.4em]
}
\date{}

\renewcommand{\YZ}[1]{}
\renewcommand{\LS}[1]{}
\renewcommand{\TY}[1]{}
\renewcommand{\slcomment}[1]{}

\begin{document}
% Use a dagger for the shared correspondence note; maketitle resets the counter.
\setcounter{footnote}{1}
\maketitle

\begin{abstract}
Large language models (LLMs) have been rapidly improving in long-context tasks, powered by Chain-of-Thought (CoT) reasoning. However, the internal mechanisms underlying this improvement remain unclear. We investigate these mechanisms through a needle-in-a-haystack (NIAH) counting task, where an LLM is asked to count the number of records dispersed in a long text. Across twelve model comparison groups, Thinking (or reasoning) improves counting accuracy over Non-thinking, with pronounced gains at larger counts. This motivates our mechanistic analysis, which identifies two contrasting mechanisms: (i) broad retrieval, where Non-thinking models broadly attend to multiple needles; (ii) targeted retrieval, where Thinking models use enumeration in CoT traces to successively retrieve needles. Targeted retrieval concentrates attention on individual needles and is accompanied by more compact internal representations. Moreover, causal intervention analysis suggests that Thinking models use the CoT trace to maintain and update an internal counter as needles are successively retrieved, even without explicit numbering. In small controlled experiments, both retrieval mechanisms and counter states emerge under standard autoregressive training. Together, our results connect long-context retrieval with representation geometry of counting, supporting a state-tracking account of CoT reasoning.
\end{abstract}

\section{Introduction}\label{sec:intro}

% Schedule the main figure and its note at the top of page 2.
\afterpage{%
\clearpage
\begin{figure}[H]
    \centering
    \includegraphics[trim=0 6bp 0 10bp,clip,width=\linewidth,keepaspectratio]{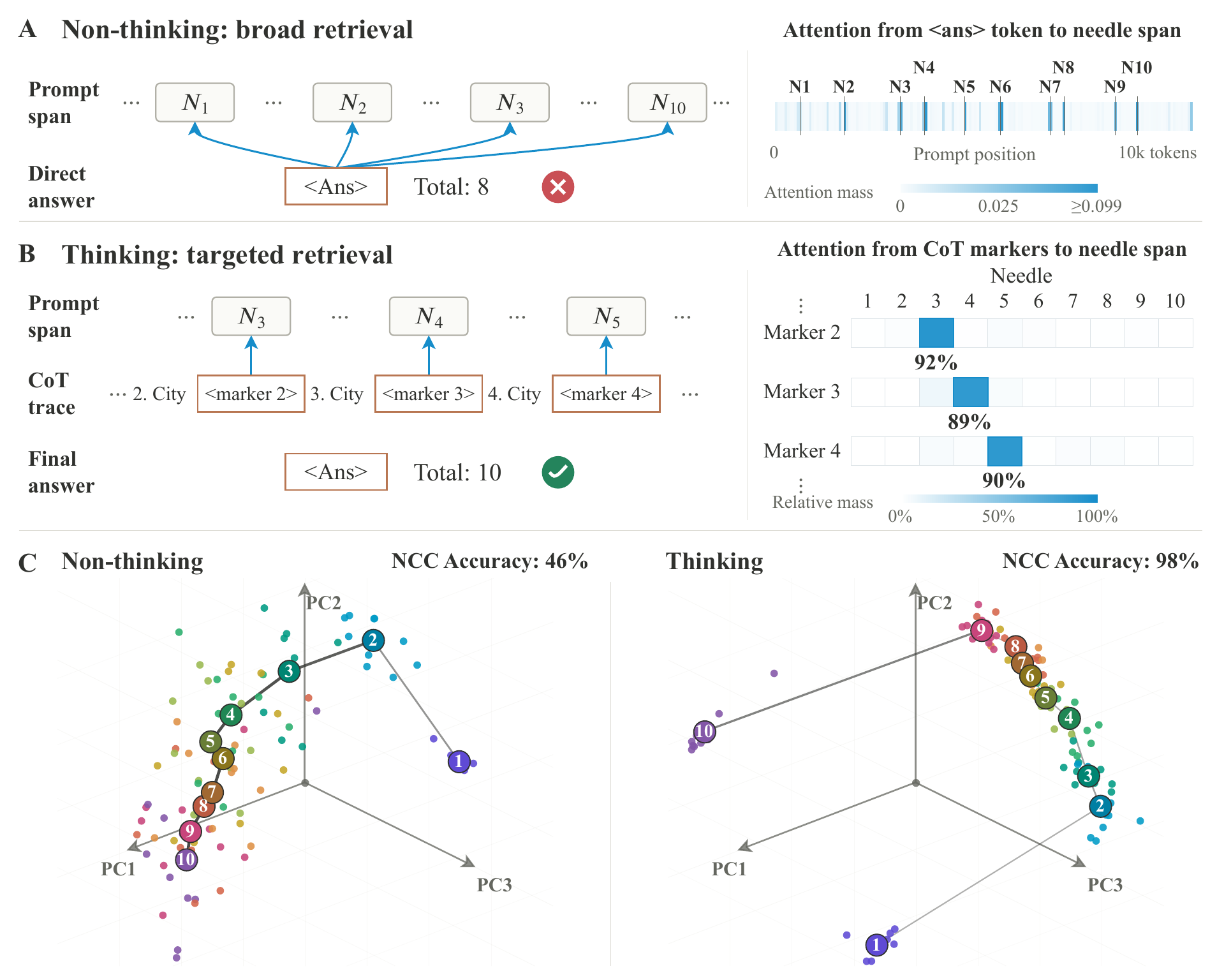}
    \caption{%\textbf{Mechanism and representation comparison of two modes.}
    \textbf{Two retrieval mechanisms and contrasting representations.}
    \textbf{A.} Non-thinking distributes final-query attention across prompt needles. \textbf{B.} Thinking targets successive needles as its trace progresses. Marker $k$ denotes the query before trace item $k+1$. \textbf{C.} PCA projections of states at prompt-needle endpoints (Non-thinking) and trace-item endpoints (Thinking), colored by running index, with numbered centroids. Held-out nearest-centroid classification (NCC) accuracy after whitened 16-component PCA is 46\% for Non-thinking and 98\% for Thinking. Attention and representations are based on Qwen3-8B with 10k-token inputs.\protect\footnotemark See details in App.~\ref{app:figure-details}.
    }
    \label{fig:main}
\end{figure}
\footnotetext{%
    An interactive 3D visualization is available at
    \url{https://github.com/Twist-Shan/count-state-geometry-supplement}.
    Download index.html and open it in a browser locally.%
}
}

Reasoning over long contexts requires large language models (LLMs) to retrieve relevant information while keeping track of progress. Chain-of-Thought (CoT) reasoning is a widely used inference paradigm in which models generate intermediate steps before producing a final answer \citep{wei2022cot}. As tasks become more complex, context engineering and agent memory address how information and task state are maintained and tracked during inference \citep{rajasekaran2025context,packer2023memgpt}. Despite these advances, a fundamental question remains: \textit{how do internal state-tracking mechanisms support long-context retrieval and reasoning?}

We present \textit{long-context counting} as a controlled testbed for studying state tracking, reducing much of the semantic ambiguity of open-ended reasoning. Adapted from needle-in-a-haystack (NIAH) benchmarks \citep{kamradt2023needle,hsieh2024ruler}, this task asks a model to count records (or needles, denoted by $N_1,N_2,N_3,\ldots$) scattered throughout a long passage (haystack), such as a passage drawn from Paul Graham essays. For example, the following input prompt contains two audit records followed by a query:
\par\begingroup\centering
``\texttt{<passage>} $\ldots$ \textit{In the 2024 city score audit, Paris received a score of 92.} $\ldots$
\textit{In the 2024 city score audit, Vienna received a score of 73.} $\ldots$ \texttt{</passage>}\\
How many city-score audit records are in the passage?''
\par\endgroup
A complete 1k-token passage example is provided in App.~\ref{app:behavioral-1k-example}. Accurate counting requires accounting for all relevant records without omissions or double counting. 
It thus combines retrieval and aggregation, two operations that language agents also need when working with long contexts, while providing explicit record and count labels for analyzing internal state tracking.  
%It thus combines retrieval and aggregation, operations also relevant to language agents that must work through evidence accumulated in context. 

%We compare Non-thinking and Thinking models (or modes). A Non-thinking model typically emits a final-answer count such as ``Total: 2'', while a Thinking model generates a CoT trace before the final count.\footnote{The Qwen3 and Gemma-4 checkpoints studied here support both Non-thinking and Thinking modes within the same checkpoint \citep{qwen3technicalreport,gemmateam2026gemma4}.} Our main analyses focus on native reasoning traces, delimited by \texttt{<think>} and \texttt{</think>} tags in Qwen. App.~\ref{app:enumeration} extends the analysis to instructed unnumbered enumeration with Thinking disabled, showing that key aspects of the observed pathway also arise outside a dedicated Thinking mode. We call records in the prompt \emph{needles} and their generated mentions \emph{trace items}.

We compare Non-thinking mode (or models), which directly outputs a count such as ``Total: 8'',
with Thinking mode (or models), which first generates a CoT trace.\footnote{
Qwen3 and Gemma-4 support both modes within the same checkpoint
\citep{qwen3technicalreport,gemmateam2026gemma4}.}
Our main analysis uses native Thinking traces. App.~\ref{app:enumeration} extends the analysis to instructed unnumbered enumeration with Thinking disabled, showing that key aspects of the observed pathway also arise outside a dedicated Thinking mode. We call records in the prompt \emph{needles} and their generated mentions \emph{trace items}.

Our study goes beyond existing mechanistic analysis, as standard in-context learning and steering often works for short contexts without explicit CoT traces \citep{hendel2023context,todd2024function,turner2024steering,rimsky2024steering}.
Here, counting provides a natural setting for studying this \textit{dynamic} process: which needle to retrieve next and how to keep track of the count. Starting from an input prompt $s_{1:t}$, a model emits tokens $s_{t+1},s_{t+2},\ldots$ one by one, iteratively growing the context $s_{1:t} \to s_{1:(t+1)} \to s_{1:(t+2)} \to \cdots$.

We examine how retrieval interacts with internal representations of counting progress, and whether these mechanisms guide subsequent reasoning generation.
Roughly speaking, representations at prompt-needle and trace-item positions follow a curve as the running index increases\footnote{We exclude self-reflection in CoT traces, as detailed in App.~\ref{app:native-prompts}.}. We call internal counting representations \emph{counter states}, and use \emph{running index} $k$ for the number of prompt needles encountered or trace items enumerated. The \emph{answer state} is the hidden representation at the answer query, labeled by total count $N$. For deeper understanding, we ask: 
\textbf{Q1} How do the two modes retrieve needles?
\textbf{Q2} How compact are their count representations?
\textbf{Q3} Do evolving counter states causally guide retrieval
and final count prediction?
%\textbf{Q1} (internal mechanism): what attention mechanisms do the two modes use for retrieving needles? \textbf{Q2} (representation geometry): how compact are representations at needle tokens?  \textbf{Q3} (counter-state evolution): do evolving counter states causally guide subsequent retrieval and final count prediction? 

%Overall, our key finding is that two distinct mechanisms exist. 
Our key finding is a contrast between two retrieval mechanisms. Non-thinking models depend on \textbf{broad retrieval}: they broadly attend to all prompt needles at the answer token but produce noisy representations and unreliable counts. Thinking models depend on \textbf{targeted retrieval}: enumeration traces couple successive record retrievals to compact, separable representations of progress, which may help avoid omissions and double counting, thereby improving the count accuracy.
Fig.~\ref{fig:main} illustrates both modes; see~App.~\ref{app:figure-details} for details. Below, we summarize our main contributions.

\begin{enumerate}
\item \textbf{A comprehensive comparison across model families.} %Across needle counts, prompt length, and model families, 
    %Across model families and scales,
    We characterize how counting accuracy varies with needle count and context length, and show that Thinking substantially improves accuracy.

\item \textbf{A mechanistic contrast between the two modes.} 
%We address Q1--Q3 by linking retrieval to representation compactness and testing the causal roles of counter and answer states (Secs.~\ref{sec:non-thinking}--\ref{sec:thinking}).
We answer Q1 and Q2 by showing that Non-thinking models collect noisy representations at prompt needles through broad retrieval, while Thinking models develop compact representations through targeted retrieval. For Q3, causal intervention results support evolving counter states in CoT traces (Secs.~\ref{sec:non-thinking}--\ref{sec:thinking}).

\item \textbf{A controlled experiment on the emergence of two mechanisms.} 
    We train small transformers from scratch under autoregressive losses, with or without CoT traces, which 
    %controls either broad retrieval emerges or targeted retrieval emerges. (Sec.~\ref{sec:emergence}。)
    reproduce key aspects of Thinking and Non-thinking mechanisms (Sec.~\ref{sec:emergence}).
\end{enumerate}

\subsection{Related work}
\label{sec:related}

Our mechanistic analysis of long-context counting connects three literatures and offers a unified account---how needles are retrieved from context, how counts are encoded geometrically, and how models maintain and update states during CoT reasoning. 

\paragraph{Long-context retrieval.} 
NIAH tests fact retrieval amid distractors \citep{kamradt2023needle}, and RULER, LongBench, and Counting-Stars extend evaluation to multiple facts and aggregation \citep{hsieh2024ruler,bai2024longbench,song2025countingstars}.
While mechanistic studies identify query-focused attention heads \citep{zhang2025qrhead} and retrieval heads \citep{wu2025retrieval}, how they support task execution is not fully explained, motivating our counting study. We contrast representation compactness between broad and targeted retrieval, providing a representation-level account of long-context retrieval.

%NIAH tests isolate fact retrieval from distractor text \citep{kamradt2023needle}, while RULER, LongBench, and Counting-Stars extend evaluation to multiple facts and aggregation \citep{hsieh2024ruler,bai2024longbench,song2025countingstars}. 
%At the mechanistic level, \citet{zhang2025qrhead} identify query-focused attention heads that retrieve information relevant to the question.
%While \citet{wu2025retrieval} identify retrieval heads, how these heads support task execution is not fully explained.
%This limitation motivates our counting study, where we contrast representation compactness between broad retrieval and targeted retrieval, providing a representation-level account of long-context retrieval.

\paragraph{Counting geometry.}
\citet{behrens2025counting} analyze how attention and feed-forward layers interact to count in small Transformers.
Other work focuses on the geometry of representations, such as low-dimensional curved manifolds that encode counts or circular calendar features \citep{gurnee2025when,engels2025nonlinear,kantamneni2025trigonometry,karkada2026symmetry}, identifies internal counters \citep{hasani2026countscope}, and examines aggregation of partial counts from explicitly partitioned inputs \citep{hasani2026mechanistic}.
However, connecting counting geometry to model behavior is challenging, as accurate linear decoding of counts does not imply correct output \citep{garcia2026rightanswer}. Our work combines layer-wise geometric analysis with causal interventions to detail the internal pathway for counting.

%\noindent\textbf{Counting geometry.}
%\citet{behrens2025counting} study how attention and feed-forward layers work together to count in small Transformers.
%Other work focuses on the geometry of representations, identifying low-dimensional curved manifolds that encode counts or circular calendar features \citep{gurnee2025when, engels2025nonlinear,kantamneni2025trigonometry, karkada2026symmetry}. Using causal interventions, \citet{hasani2026countscope} identify internal counters in text and vision, but their study is mostly limited to simple item-counting tasks. \citet{hasani2026mechanistic} use causal interventions to study how models represent and aggregate partial counts from explicitly partitioned inputs. However, connecting counting geometry to model behavior is challenging, as accurate linear decoding of counts does not imply correct output \citep{garcia2026rightanswer}. Our work combines layer-wise geometric analysis with causal interventions to detail the internal pathway for counting.

\paragraph{CoT reasoning and state-tracking mechanisms.}
Despite theoretical benefits of CoT reasoning  \citep{feng2023towards,merrill2024expressive,malach2024autoregressive}, the mechanism of CoT reasoning in practice remains unsettled, partly due to the difficulty of analyzing growing contexts.
Prior work examines Bayesian belief states in in-context learning \citep{shai2024belief,bigelow2025beliefdynamics,yan2026taskvector}, iterative retrieval in controlled CoT tasks \citep{cabannes2024iteration}, and state tracking in toy settings \citep{zhang2025fsa,li2025trackstate,huang2025lengthgeneralization,forner2026learningdynamics,wang2026unfaithful}.
Other work analyzes CoT at a coarser level \citep{yang2026howcot}.
Our work provides evidence that pretrained LLMs maintain and iteratively update count states while retrieving sparse records from long distractor passages.

%\noindent\textbf{CoT reasoning and state-tracking mechanism.} Despite theoretical benefits of CoT reasoning \citep{feng2023towards,merrill2024expressive, malach2024autoregressive}, the mechanism of CoT reasoning in practice remains unsettled, partly due to the difficulty of analyzing growing contexts. Prior work on Bayesian belief states for in-context learning \citep{shai2024belief, bigelow2025beliefdynamics, yan2026taskvector} is indicative of a general state-tracking mechanism.
%In controlled CoT tasks, \citet{cabannes2024iteration} identify iteration heads that retrieve successive inputs as the model updates an intermediate state.
%But current understanding of CoT reasoning is either limited to toy settings \citep{zhang2025fsa, li2025trackstate, huang2025lengthgeneralization,forner2026learningdynamics, wang2026unfaithful} or based on coarse-grained analyses \citep{yang2026howcot}. 
%Our work provides evidence that pretrained LLMs maintain and iteratively update count states while retrieving sparse records from long distractor passages.

\section{Thinking vs.~Non-thinking: performance across model families}
\label{sec:cross-model}

%To set up the behavioral comparison, we evaluate Non-thinking and Thinking on counting city--score records in distractor text, varying target count $N$ (1--20) and passage length $L$ (1k--20k tokens) across 12 comparison models.
%Most groups compare modes within one checkpoint; GLM and Ministral use separate checkpoints.
%Figure~\ref{fig:empirical-accuracy-laws} summarizes exact-answer accuracy and descriptive fits; models, prompts, and scoring are detailed in Appendix~\ref{app:empirical-law}.

\begin{figure}[!t]
    \centering
    \includegraphics[trim=0 1.5bp 0 1.5bp,clip,width=\linewidth]{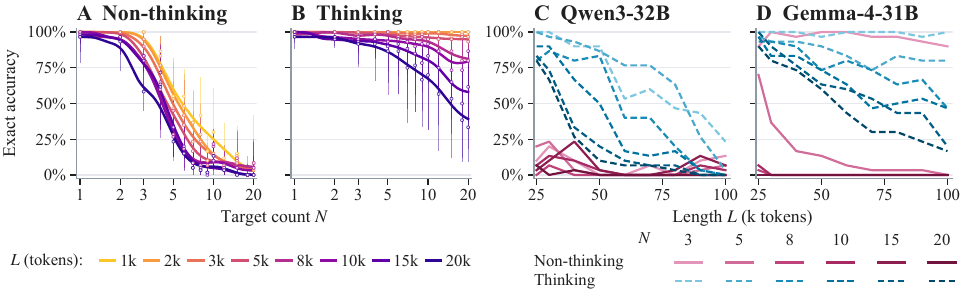}
    \caption{\textbf{Counting accuracy across counts and context lengths.}
    \textbf{A,B.} Median and interquartile range across 12 model groups; lines are smoothed guides.
    \textbf{C,D.} Accuracy over 30 paired seeds: Non-thinking (rose, solid) and Thinking (blue, dashed), with darker colors for larger $N$.
    Lines connect measured values. Qwen is evaluated with YaRN disabled.}
    \label{fig:empirical-accuracy-laws}
\end{figure}

We evaluate Non-thinking and Thinking on our NIAH counting task, varying target count $N$ (1--20) and passage length $L$ (1k--20k tokens) across 12 comparison groups. Most models have a Thinking/Non-thinking switch within one checkpoint; GLM and Ministral use separate checkpoints.
Figure~\ref{fig:empirical-accuracy-laws} summarizes observed accuracies and extends the comparison to 25k--100k passage length for Qwen3-32B and Gemma-4-31B. See App.~\ref{app:empirical-law} for prompts, evaluation, and per-model results. %This behavioral contrast between Non-thinking and Thinking motivates our mechanistic studies in later sections.

\begin{table}[H]
\centering
\setlength{\tabcolsep}{1.2pt}
\renewcommand{\arraystretch}{0.88}
\caption{\textbf{Thinking improves counting accuracy across models.} Table shows parsed exact accuracy (\%), $N=5$, over 30 paired seeds. %Large context lengths hurt both modes .
$^\dagger$Separate checkpoints.}
\label{tab:empirical-n5}
\begin{tabular}{@{}l@{\hspace{5pt}}c@{\hspace{8pt}}*{12}{c}@{}}
\toprule
& & \multicolumn{4}{c}{Qwen3}
& \multicolumn{4}{c}{Gemma-4}
& \multicolumn{2}{c}{Nemotron}
& GLM$^\dagger$
& Ministral$^\dagger$ \\
\cmidrule(lr){3-6}
\cmidrule(lr){7-10}
\cmidrule(lr){11-12}
\cmidrule(lr){13-13}
\cmidrule(lr){14-14}
$L$ & Mode
& 4B & 8B & 14B & 32B
& E4B & 12B & 26B-A4B & 31B
& v2-9B & 3-4B
& 4/Z1-9B & 3-8B \\
\midrule
1k & Non-thinking
& 43.3 & 60 & 50 & 96.7
& 36.7 & 73.3 & 100 & 100
& 40 & 46.7 & 20 & 73.3 \\
   & Thinking
& 100 & 100 & 100 & 100
& 100 & 100 & 100 & 100
& 100 & 96.7 & 100 & 96.7 \\
\midrule
20k & Non-thinking
& 3.3 & 30 & 30 & 46.7
& 33.3 & 10 & 43.3 & 63.3
& 3.3 & 13.3 & 13.3 & 16.7 \\
    & Thinking
& 90 & 96.7 & 100 & 100
& 43.3 & 83.3 & 93.3 & 100
& 33.3 & 73.3 & 46.7 & 70 \\
\bottomrule
\end{tabular}
\end{table}

\paragraph{Non-thinking is unreliable at large counts.}
Even in short passages, Non-thinking accuracy drops sharply with large count. At $L=1$k, median accuracy across the 12 model groups falls from 100\% at $N=3$ to 30\% at $N=10$ and 0\% at $N=20$ (Fig.~\ref{fig:empirical-accuracy-laws}A). 

%\paragraph{Thinking accuracy drops at large counts + long context.} While Thinking alleviates large-count difficulty, its accuracy drops when both factors are present. In Figure...\YZ{describe.}

%\paragraph{Thinking remains sensitive to count and context length.}
\paragraph{Thinking is better at large counts, yet still worsens at long context.}
Thinking maintains high accuracy over a wider count range (Fig.~\ref{fig:empirical-accuracy-laws}B). Averaged over the complete 1k--20k grid, it improves accuracy by 28--63 percentage points across the 12 groups. While Thinking alleviates large-count difficulty, its accuracy drops when large counts are combined with longer contexts (Fig.~\ref{fig:empirical-accuracy-laws}C,D).

%\paragraph{A quantitative explanation.}
\paragraph{A heuristic explanation.} For Non-thinking, %Fig.~\ref{fig:main}C suggests 
a plausible explanation is that broad retrieval accumulates noise in the representations as needle counts increase, and thus, the signal strength and count accuracy deteriorate quickly as $N$ increases. Indeed, at 10k tokens, neighboring counts are harder to distinguish in Qwen3-8B and Gemma-4-E4B at larger $N$, mainly because their mean representations are closer together, yielding an increase in relative noise (App.~\ref{app:empirical-relative-noise}).
%This shows an increase in relative noise, but does not establish an increase in absolute noise. %\LS{Fig.~\ref{fig:main}C shows representations at fixed $N=10$, so it does not establish that noise grows with $N$? Could we frame this as a possible explanation?} 
For Thinking, targeted retrieval still requires successful retrieval at each step. If each of the $N$ steps fails independently with probability $\gamma_L$ at context length $L$, counting accuracy is approximately 
\[
(1-\gamma_L)^N \approx 1-N\gamma_L, \qquad N\gamma_L\ll1.
\] Thus, any increase in per-step retrieval error at longer contexts is amplified over multiple steps.

%For Thinking, targeted retrieval in Fig.~\ref{fig:main}B,C may imply that the interaction of $N$ and $L$ is crucial. Indeed, if correct counting requires $N$ successful retrieval steps with independent failure probability $\gamma_L$, then the accuracy is approximately $[1-\gamma_L]^N \approx 1-N\gamma_L$ if $N\gamma_L \ll 1$. %\LS{Does this approximation $(1-\gamma_L)^N \approx 1-N\gamma_L$ require $N\gamma_L \ll 1$? We could keep only the product form maybe?} 
See App.~\ref{app:empirical-regression} for a detailed quantitative analysis.
%fits and generally support our heuristics. \LS{The fits describe the accuracy trends, but do not establish independent retrieval failures or the proposed noise mechanism, maybe we should claim more conservative?}%This contrast between Non-thinking and Thinking motivates a mechanistic analysis of the retrieval mechanisms, representation geometry, and the evolution of counter-states over CoT traces.
%This contrast motivates our mechanistic analysis of how
Motivated by this contrast, we examine how the two modes retrieve needles, represent counts, and track counting progress through CoT traces.

\section{Non-thinking mode: form, retrieve, and consolidate}
\label{sec:non-thinking}

%\YZ{I think a reasonable plan is to have two pages for this section. One subsection introduces basic definition and representation geometry (0.5 pages); then the second subsection details the three-step mechanism (1.5 pages).}

%Given time and computational constraints, we focus on Qwen3-8B and Gemma-4-E4B with approximately 10k-token passages for all the following mechanistic analysis. Layer and head indices are one-based throughout; layer 1 denotes the first Transformer block.

% Our mechanistic analysis in Secs.~\ref{sec:non-thinking}--\ref{sec:thinking} primarily focuses on Qwen3-8B and Gemma-4-E4B with approximately 10k-token prompts. Throughout, layer 1 denotes the first Transformer block. In both sections, we identify retrieval heads using one set of seeds and evaluate their causal effects on prompts generated with held-out seeds.
Our mechanistic analysis in Secs.~\ref{sec:non-thinking}--\ref{sec:thinking} primarily focuses on Qwen3-8B and Gemma-4-E4B with approximately 10k-token prompts.\footnote{Throughout, layer 1 denotes the first Transformer block.} In both sections, we identify retrieval heads using one set of seeds and evaluate their causal effects on prompts generated with held-out seeds.

%To avoid we use \emph{seeds} to select heads and layers and fit probes, then evaluate on held-out \emph{held-out data} with these choices fixed.  
We also test the robustness of our results by varying the settings: (i) simulated CoT reasoning using structured bullet-enumeration instructions with Thinking disabled, which provides further evidence for the retrieve--update--read pathway without explicit item numbering (App.~\ref{app:enumeration}), and (ii) other retrieval tasks such as retrieving the $k$th record and counting within a specified category (App.~\ref{app:additional-task}). 
Together, these results support our findings across reasoning settings
and retrieval tasks. Instructed enumeration further suggests a way to
study reasoning behavior in closed-source models through verifiable
intermediate outputs.
%Additional results under variant settings are consistent with our main findings.

\subsection{Broad attention, noisy representations}
\label{sec:broad-retrieval-score}

\begin{figure}[htb]
    \centering
    \includegraphics[trim=0 1.5bp 0 1.5bp,clip,width=\linewidth]{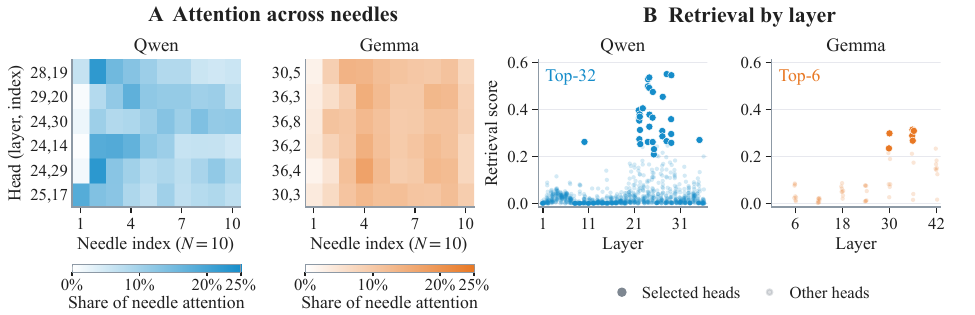}
    \caption{\textbf{Broad retrieval in Non-thinking mode.} \textbf{A.} Needle-attention shares for each model's six highest-ranked heads, normalized within each $N=10$ prompt and averaged over 20 seeds. \textbf{B.} Broad retrieval scores $B_h$ across counts 2--10; highlighted points mark the frozen Qwen Top-32 and Gemma Top-6 sets. Gemma includes global-attention layers only.}
    \label{fig:nonthinking-broad-retrieval}
\end{figure}

First, we examine attention patterns at the final-answer query---the last token of the supplied assistant prefix \texttt{Total:}---in our NIAH counting task. A subset of heads concentrates attention on needle spans (i.e., all tokens in a record), consistent with prior work on retrieval heads \citep{wu2025retrieval}. These heads distribute attention across multiple needles, although individual heads exhibit position preferences (Fig.~\ref{fig:nonthinking-broad-retrieval}A).

%\YZ{Write a quantitative sentence.} \YZ{I remember this is a plot that shows attention is on similar scales for broad retrieval head, but I can't find the plots.}

% Figure 3 follows the score definition.

To quantify both the amount and breadth of needle-directed attention, let $A_h(q,s)$ denote head $h$'s attention from query $q$ to key $s$. At the answer query $q_A$, the attention mass on needle span $\mathcal S_j$ is $m_{hj}=\sum_{s\in\mathcal S_j}A_h(q_A,s)$. Across $N$ needles, define the total mass $M_h=\sum_{j=1}^{N}m_{hj}$ and normalized shares $p_{hj}=m_{hj}/M_h$ for $M_h>0$. The \textit{broad retrieval score} is
\[
B_h=\frac{M_h\exp(H_h)}{N}, \qquad
H_h=-\sum_{j=1}^{N}p_{hj}\log p_{hj}.
\]
Here, $\exp(H_h)$ is the effective number of attended needles, equaling $k$ for uniform attention over $k$ needles; dividing by $N$ gives effective coverage, reaching 100\% when attention is uniform across all needles. Thus, $B_h$ rewards both substantial needle-directed attention and broad coverage. We set $0\log0=0$ and $B_h=0$ when $M_h=0$. Across prompts with 2--10 needles, mean effective coverage is 81.2\% for Qwen's Top-32 heads and 80.3\% for Gemma's Top-6 heads. High-scoring heads concentrate mainly at Qwen L22--29 and Gemma L30--36 (Fig.~\ref{fig:nonthinking-broad-retrieval}B).

%\YZ{explain why it measures broad retrieval.} 

%\YZ{Identify a few representative heads, make some small plots to show the spread of attention across prompt needles.}\\

%\YZ{Use a small plots to show the distribution of broad retrieval scores across layers, explain that they concentrate in middel layers.}

\paragraph{How well do prompt-needle representations encode counts?}
We next examine whether states at the final token of each needle span (needle end states) encode each needle's running index---the number of needles encountered so far. Across $N=1$--10, ridge regression reads out this index with high held-out $R^2$ (0.910/0.827 for Qwen/Gemma at L18/L13). However, predictive accuracy alone does not establish that the decoded direction controls the model's count. Motivated by the count-steering results of \citet{hasani2026countscope}, a separate analysis tests linear steering at the last needle span using ridge directions fitted on $N=10$ inputs. Across layers, it fails to reliably increase or decrease the answer in the intended direction (ten initially correct $N=3$ prompts per model; App.~\ref{app:nonthinking-steering}).

To characterize these representations, we use PCA to describe their geometry and nearest-centroid classification (NCC) to quantify running index separability, with NCC selecting Qwen L18/Gemma L13 on the $N=1$--10 fitting set (App.~\ref{app:nonthinking-geometry}). 
Class means are averages of needle end states with the same running index. The first three principal components of these means explain 98.1\% of their variance for Qwen and 97.5\% for Gemma, and NCC balanced accuracy on held-out individual states is only 46\% and 40\%, respectively. Individual states overlap despite the low-dimensional structure of their class means. Also, we tried removing the opening task instructions, which leaves similar running index centroid representations (App.~\ref{app:nonthinking-robustness}).  

%Thus, Non-thinking representations encode counting progress, but individual states are not cleanly separated by running index.

%Thus, low-dimensional centroid structure and high regression $R^2$ coexist with substantial overlap between index classes. PCA and NCC characterize this geometry; they do not by themselves establish causal use.

%\YZ{Explain the panel in the main figure, show good ridge regression but mediocre NCC accuracy.}\\
%\YZ{Explain the steering at prompt needle spans fails. Briefly connect to the existing counting paper.}\\
%\YZ{Show that changes in prompt formats and needles (city --> flower) introduce more noise in the representations, which shows prompt needle representations fail to reliably represent count-state.}

\subsection{Causal analysis of the mechanism}

We next test how needle span information contributes to the final count through causal interventions (e.g., head ablation, activation patching) at needle spans, retrieval heads, and the answer query. These experiments support a form--retrieve--consolidate pathway across layers.

%For broad retrieval, we detail a three-step mechanistic pathway connecting different locations and layers, supported by causal intervention analysis. %We present our analysis for Qwen3-8B here and defer a similar analysis for Gemma4-?B to Appendix~\ref{}. 

% Figure follows its explanatory text: nonthinking_form_retrieve_consolidate.pdf

\begin{figure}[htb]
    \centering
    \includegraphics[trim=0 8bp 0 8bp,clip,width=\linewidth,keepaspectratio]{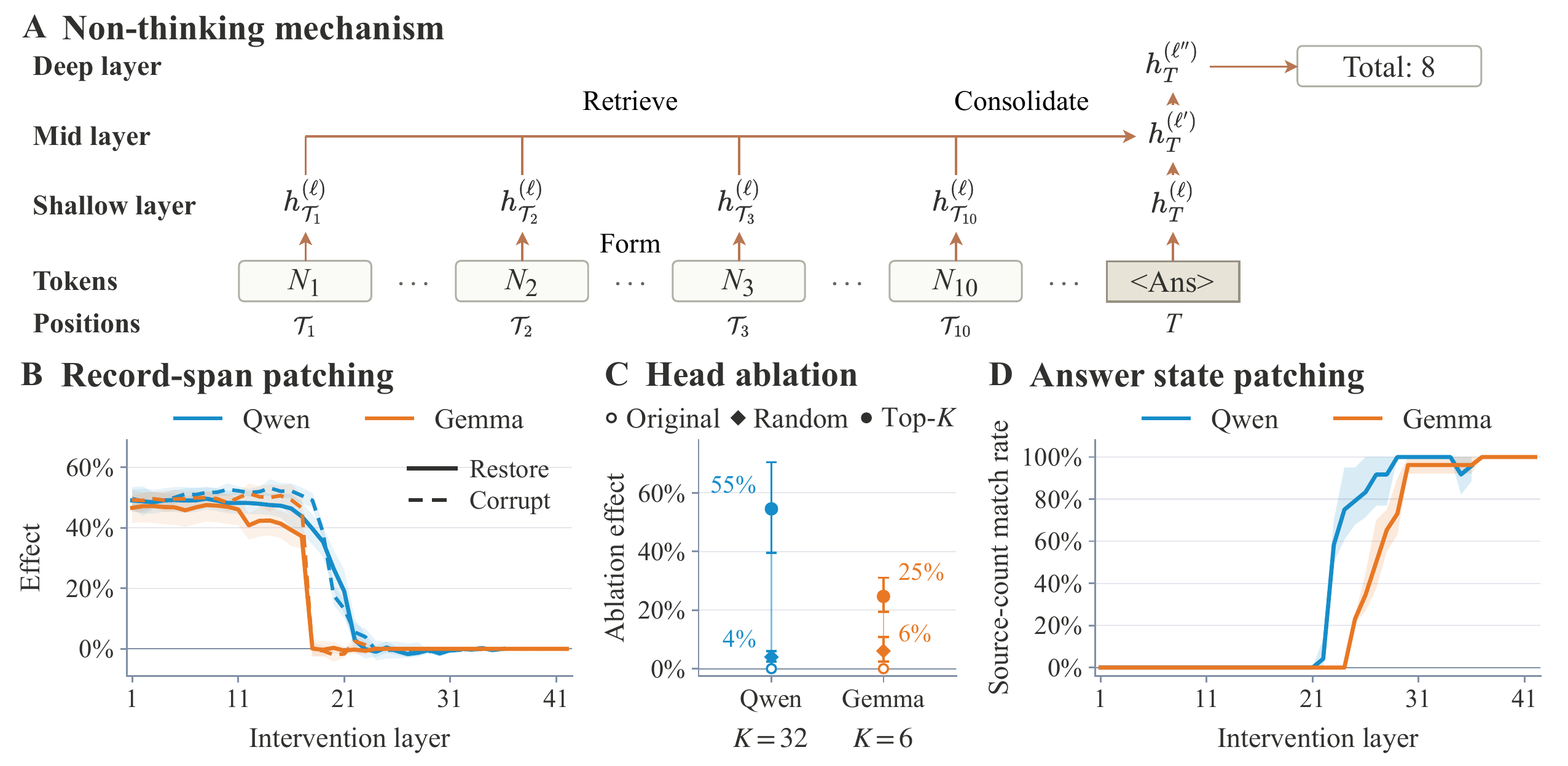}
% Use default caption spacing.
    \caption{\textbf{The form--retrieve--consolidate pathway.} \textbf{A.} Schematic. \textbf{B.} Clean-to-corrupted restoration and reverse patching of needle spans. \textbf{C.} Retrieval-head ablation vs.~layer-matched random and original controls. \textbf{D.} Proportion of predictions matching the source count after answer state patching.}
    \label{fig:nonthinking-form-retrieve-consolidate}
\end{figure}

\paragraph{Form: count-encoding representations form in shallow layers.}
% \YZ{State layerwise results: how representations encode count index, ridge regression, clustering quality, NCC, etc.}
% What layers have causal effects in model's final count? We consider causal intervention experiments: \YZ{Define ``donor''. ``receive'' here.} \YZ{Is it better to use the language ``source'', ``target''? Check the literature.}
Needle span representations carry information about each needle's running index---the number of needles encountered so far. These representations remain noisy and overlap across indices. 
To test their contribution, we replace needle spans with length-matched ordinary text to create corrupted inputs. We then replace hidden states at these positions in the corrupted run with those from the original clean run at the same layer, testing whether this restores the correct count~\citep{heimersheim2024patching}. We also reverse the replacement to test the effect of corrupting clean states. Here, ``clean'' denotes unmodified inputs, regardless of answer correctness. We measure restoration by the decrease in normalized count error $|E-N|/N$, and reverse patching by its increase (Eq.~\ref{eq:nonthinking-span-effects}). 
% Here $E$ is the expected count conditional on candidate answers 1--10 and $N$ is the true count.
Here $E$ averages counts 1--10 using normalized candidate probabilities, and $N$ is the true count. Restoring clean needle span states into corrupted targets, or reversing this direction, gives large effects at shallow layers (47--50\% at L1), falling below 5\% past Qwen L23 and Gemma L18 (Fig.~\ref{fig:nonthinking-form-retrieve-consolidate}B). 
%Thus, interventions at needle positions
%Needle-site interventions thusaffect the answer primarily before these drop-off layers.

%These effects are large in shallow layers (47--50\% at L1) and fall below 5\% by Qwen L23 and Gemma L18 (Fig.~\ref{fig:nonthinking-form-retrieve-consolidate}B).

\paragraph{Retrieve: broad retrieval collects needles in middle layers.} At the answer query, broad retrieval heads attend to multiple needle spans and aggregate their information into the representation at the answer position. To test this role, we ablate the selected retrieval heads identified in Figure~\ref{fig:nonthinking-broad-retrieval}B at the query token, producing normalized absolute count changes ($|\widehat N_{\mathrm{ablated}}-\widehat N_{\mathrm{clean}}|/N$, where $\widehat N$ denotes the generated count) of 54.5\% for Qwen and 24.7\% for Gemma, versus 4.0\% and 6.0\% for layer-matched random heads (Fig.~\ref{fig:nonthinking-form-retrieve-consolidate}C). 
%Top-k sweeps and initially correct-input controls are in Appendix~\ref{app:nonthinking-retrieval}.
We also varied the number of ablated heads; see App.~\ref{app:nonthinking-retrieval}.

% \noindent\textbf{\textbf{Consolidate}: answer states consolidate count information in deep layers.} We interpret the deeper layers as integrating the retrieved needle information into an answer state that guides the final count prediction.
\paragraph{Consolidate: answer states support final count prediction.} We test whether count information at the answer query causally influences the final prediction. 
To test this role, we patch answer states between different-count prompts with correct baseline answers, one layer at a time, and measure the proportion of predictions matching the source count (App.~\ref{app:nonthinking-causal}).
%Predictions shift toward the source count, with transfer effects
At the final layer, this proportion reaches 95.8\% for Qwen and 100\% for Gemma (Fig.~\ref{fig:nonthinking-form-retrieve-consolidate}D). 
% Subspace ablation further shows that a low-dimensional count-aligned subspace of the answer state contributes to the final count prediction (App.~\ref{app:nonthinking-causal}).
Subspace ablation further supports a causal role for count-related information in the answer state (App.~\ref{app:nonthinking-causal}). Together, these results are consistent with consolidation at the answer query.
%Through subspace ablation, we also find that count information tends to concentrate in a low-dimensional subspace of the answer state (Appendix~\ref{app:nonthinking-causal}).
%Removing count-aligned components at retrieval and answer sites also reduces restoration more than removing norm-matched orthogonal components (Appendix~\ref{app:nonthinking-causal}).

%\append{Non-thinking appendix: complete Top-$K$ sweeps, answer state removal, serial mediation, retrieval-head distributions, OV/QK observations, and per-seed and correct-only controls.}

\section{Thinking mode: retrieve, update, and read}
\label{sec:thinking}

For Thinking mode, we examine retrieval and state representations at each step as the model enumerates items in its thinking trace. Variation in enumeration formats (or grammars) complicates the analysis of thinking traces by affecting how trace items and their token positions are identified. We therefore restrict the analysis to successfully parsed trace items; see App.~\ref{app:native-prompts} for details.
%which we call \emph{item} in the following section.

\subsection{Targeted attention, compact representations}
\label{sec:targeted-retrieval-score}

\begin{figure}[htb]
    \centering
    \includegraphics[trim=0 1bp 0 1.5bp,clip,width=\linewidth]{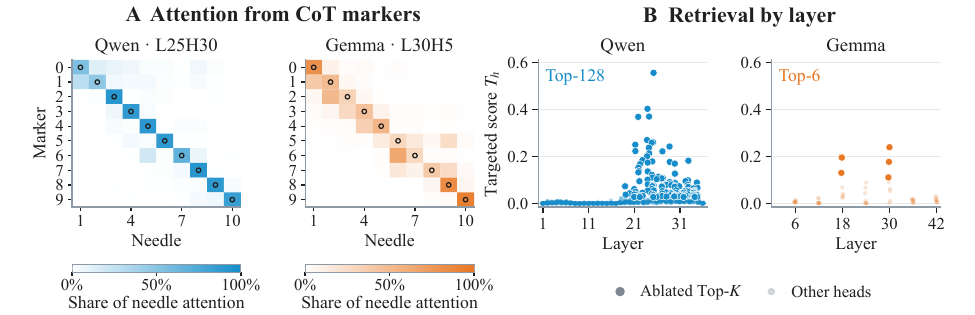}
% Use default caption spacing.
    \caption{\textbf{Targeted retrieval in Thinking.}
    \textbf{A.} Needle-normalized attention for one head per model ($N=10$); circles mark needle $k+1$ at Marker $k$.
    \textbf{B.} Mean $T_h$ across seeds; highlights mark Qwen Top-128 and Gemma Top-6.
    Gemma includes global-attention layers only.}
    \label{fig:native-targeted-retrieval}
\end{figure}

The thinking trace uses successive queries to retrieve needles in the prompt. We examine attention patterns at the token immediately before each trace item. 
The Qwen and Gemma heads shown here mainly attend to the next record as enumeration proceeds (Fig.~\ref{fig:native-targeted-retrieval}A).

% Figure follows its explanatory text: native_targeted_retrieval.pdf

To quantify this pattern, let $q_k$ (Marker $k-1$ in the plots) denote the query preceding trace item $k$ and $\mathcal S_k$ the complete span of its corresponding prompt needle. The \textit{targeted retrieval score} is 
\[
T_h(q_k,\mathcal S_k)=\sum_{s\in\mathcal S_k}A_h(q_k,s).
\]
The score ranges from zero to one and measures the fraction of head $h$'s attention assigned to the next needle. For the heatmaps in Fig.~\ref{fig:native-targeted-retrieval}A, we divide each needle's attention mass by the total attention mass received by all needles.
%share of the total attention assigned to all needles; while 
%We average scores over eligible queries within each seed and then weight seeds equally. 
In our calculation, we average scores over trace items that can be parsed, first within each seed and then across seeds (App.~\ref{app:cot-representations}).
We use the highlighted Qwen Top-128 and Gemma Top-6 heads for the subsequent causal analysis (Fig.~\ref{fig:native-targeted-retrieval}B).
%The highlighted Qwen Top-128 and Gemma Top-6 banks are fixed for the subsequent causal tests.

\paragraph{How well do trace-item states encode counting progress?} 
As in Non-thinking mode, we use PCA to visualize representation geometry and NCC to measure running index separability.
We first compare the two modes on the same 30 Qwen inputs with $N=10$, selecting complete Thinking traces with consecutive numbering. At the selected layers (Non-thinking L13; Thinking L31), representations in Thinking traces form more compact clusters by running index (Fig.~\ref{fig:main}C). Held-out NCC accuracy in this comparison is 98\% for trace-item states, compared with 46\% for prompt-needle states in Non-thinking mode. The main appendix analyses of running index and final count use $N=1$--10 in both modes (Appendices~\ref{app:nonthinking-geometry} and~\ref{app:cot-representations}), while Fig.~\ref{fig:main}C uses $N=10$.
%\YZ{Move to the appendix. Write something at the start of Section 4 about parsing and enumeration formats in CoT traces.}

Furthermore, we also compare Thinking with Non-thinking in terms of sensitivity to the needle domain. 
With answer state NCC classifiers fitted on city examples and held fixed, replacing city needles with flower needles reduces accuracy by 17 and 15 percentage points for Qwen and Gemma in Non-thinking. 
The drops are much smaller in Thinking: 2 percentage points for Qwen and none for Gemma (App.~\ref{app:nonthinking-robustness} and~\ref{app:cot-representations}). 
%This result also suggests that Thinking may act as a denoising process, making count representations more robust to variations.
This contrast shows that count readout from answer states is more robust to needle-domain variation in Thinking mode, consistent with the heuristics that reasoning acts as representation-level denoising.

%The higher sensitivity in Non-Thinking models is not due to the opening instructions, as removing the opening instructions retains similar running index centroid configurations at the selected layers. 

%Decodable count information does not guarantee robust transfer across needle types. Removing the opening instructions retains similar running index centroid configurations at the selected layers. 

%For Thinking, \YZ{add a few numbers }. In contrast, for Non-Thinking, replacing city needles with flower needles lowers frozen answer state nearest-centroid accuracy from 67\% to 50\% for Qwen and from 58\% to 43\% for Gemma.(Appendix~\ref{app:nonthinking-robustness}). 

\subsection{Causal analysis of the mechanism}
\begin{figure}[htb]
    \centering
    \includegraphics[trim=0 6bp 0 1.5bp,clip,width=\linewidth,keepaspectratio]{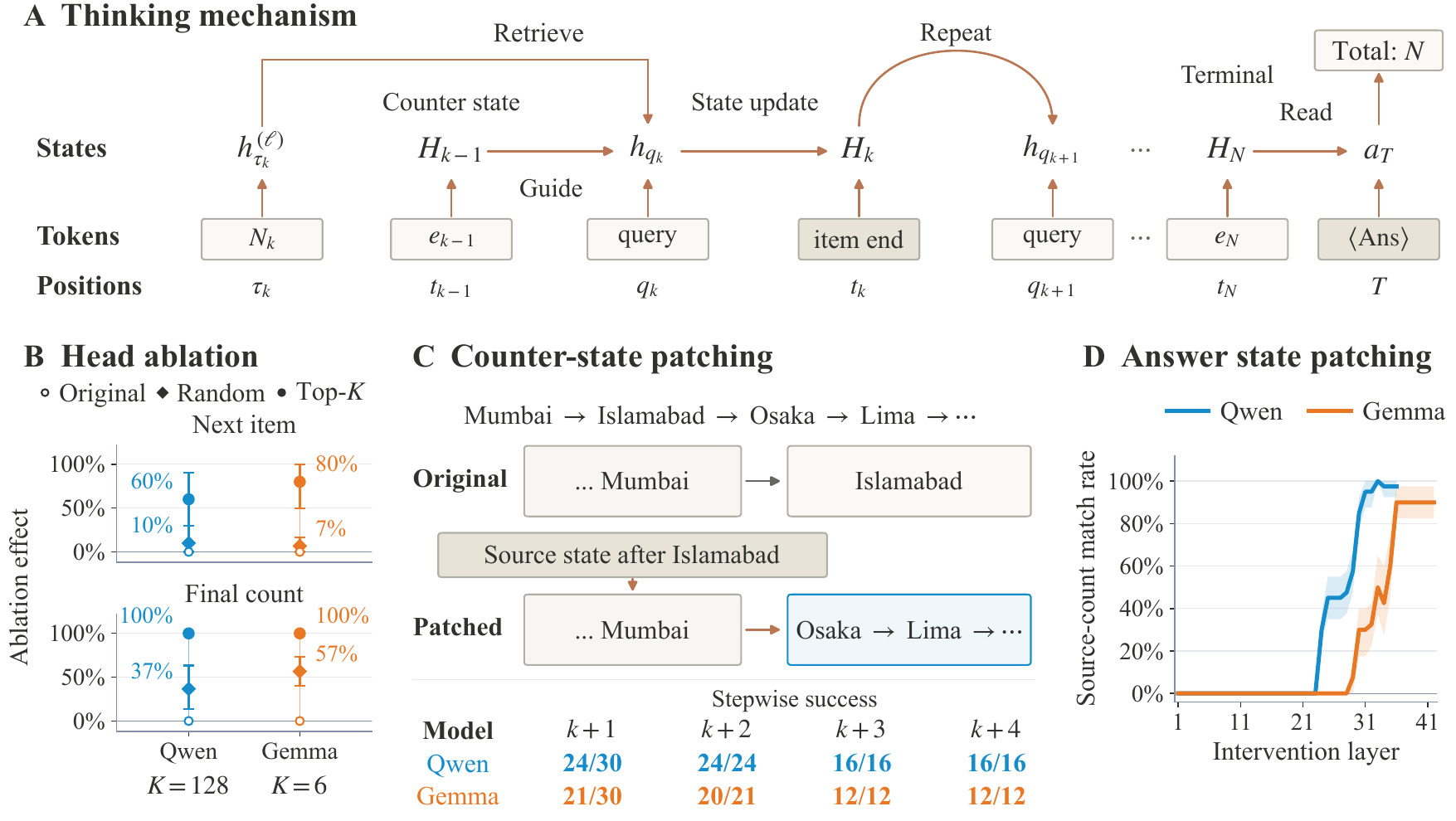}
% Use default caption spacing.
    \caption{\textbf{The retrieve--update--read pathway.} \textbf{A.} Schematic. \textbf{B.} Retrieval-head ablation vs.~layer-matched random and original controls. \textbf{C.} Counter state patching from a later trace item to an earlier one, using traces without indices (Qwen L19; prompted Gemma L21). Later columns report continuation success conditional on success at the preceding step and a remaining item. \textbf{D.} Proportion of predictions matching the source count after answer state patching.}
    \label{fig:native-retrieve-encode-count-loop}
\end{figure}

We use head ablation and state patching to test how retrieval heads, counter states, and answer states contribute to counting. The results support a retrieve--update--read pathway in Thinking (Fig.~\ref{fig:native-retrieve-encode-count-loop}A).

% Figure follows its explanatory text: native_retrieve_encode_count_loop.pdf

\paragraph{Retrieve: targeted retrieval precisely attends to the next needle.}
Targeted retrieval heads retrieve information from the next needle in the prompt as each trace item is generated. To test this role, we ablate the selected heads identified in Figure~\ref{fig:native-targeted-retrieval}B, starting at the selected marker and continuing at every subsequent token. Ablating these heads increases next-item failure rates by 60 percentage points for Qwen and 80 for Gemma, versus 10 and 6.7 for layer-matched random heads. Final count failure rate increases by 100 percentage points in both models, versus 36.7 and 56.7 for layer-matched random heads (Fig.~\ref{fig:native-retrieve-encode-count-loop}B). The selected heads therefore contribute to both the next lookup and completion of the count. We also varied the number of ablated heads; see more details in App.~\ref{app:cot-current-dose}.

\paragraph{Update: counter states guide continued enumeration.}
To avoid explicit numbering as a cue to counting progress, we use traces without indices: naturally generated traces for Qwen and the prompted \texttt{FOUND:} format for Gemma. We then test whether counter states guide the next retrieval by transplanting states from source item $k$ into target item $k-1$ \YZ{do we patch source $k+1$ to target $k$? Make it consistent with the later results?}, aligning the ends of the two item spans. We patch at the layers with the highest NCC accuracy: Qwen L19 and Gemma L21, restricting Gemma's candidates to L1--L22 (App.~\ref{app:cot-progress}). Across ten held-out seeds per model and $k=4,6,8$, the next generated city is record $k+1$ in 24/30 Qwen and 21/30 Gemma trials, compared with 0/30 and 2/30 when patching each target with its own states (Fig.~\ref{fig:native-retrieve-encode-count-loop}C). Among first-step successes, the following item matches record $k+2$ in 24/24 and 20/21 trials without another patch. The patched states therefore influence multiple steps of enumeration. 

\paragraph{Read: trace information supports the final count.}
At the answer query, the model uses information in the completed trace to predict the final count. To test this dependence, we hold the generated trace fixed and separately zero out prompt-record and trace representations while preserving token positions. Removing prompt-record representations leaves exact-count accuracy unchanged at 97\% for Qwen and 70\% for Gemma, whereas removing trace representations reduces it to 1\% and 15\% (Fig.~\ref{fig:cot-answer-readout-controls}A,B). We then test the role of the answer state by patching between different-count prompts with correct baseline answers, one layer at a time. As in Non-thinking mode, we measure the proportion of predictions matching the source count. At the final layer, this proportion reaches 97.5\% for Qwen and 90\% for Gemma (Fig.~\ref{fig:native-retrieve-encode-count-loop}D).

\section{Synthetic Experiment: Emergence of Two Retrieval Mechanisms}
\label{sec:emergence}

\begin{figure}[htb]
    \centering
\includegraphics[trim=0 6bp 0 1.5bp,clip,width=\linewidth,keepaspectratio]{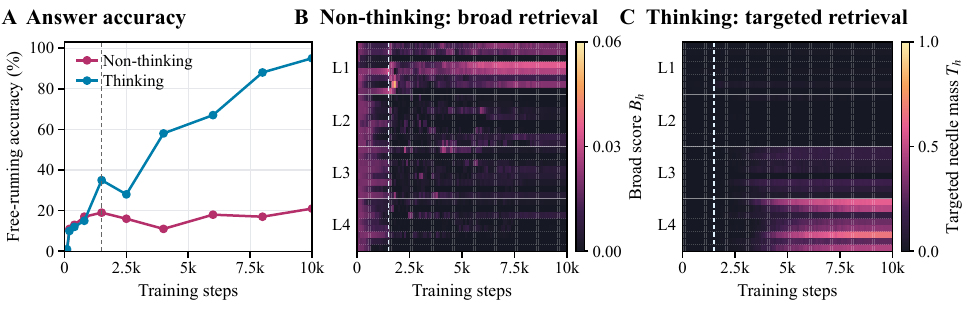}
% Use default caption spacing.
    \caption{\textbf{Learning dynamics of two synthetic counting models.} \textbf{A.} Free-running count accuracy. \textbf{B,C.} Layer-wise broad retrieval scores in Non-thinking and targeted retrieval scores in Thinking. The dashed line marks the switch to task-output loss at step 1,500. See Appendix~\ref{app:figure-details} for evaluation details.}
    \label{fig:emergence}
\end{figure}
% Figure follows its explanatory text: synthetic_section.pdf

% How do models develop the two distinct mechanisms during training? 
% We train 4-layer, 8-head transformers from scratch adapted from nanoGPT~\citep{karpathy2022nanogpt}.
%To examine how the retrieval patterns emerge during training,
Complementing our analysis of pretrained LLMs, we next test whether the mechanism contrast can emerge under standard autoregressive training, to understand training dynamics and the role of data.
%Complementing our analysis of pretrained LLMs, we present a synthetic experiment to understand training dynamics and the role of data.
We train two separate 4-layer, 8-head transformers from scratch, with and without enumeration traces, using an implementation adapted from nanoGPT~\citep{karpathy2022nanogpt}. The training data are a modified Shakespeare corpus: each input sequence starts with \textcolor{promptblue}{a query}, then a character-shuffled 256-character window from Tiny Shakespeare with target counts of 1--10, followed by an optional \textcolor{thinkvermillion}{CoT trace} and then the \textcolor{ansgreen}{final count}. Both models share the architecture, input distribution, and training budget; only the Thinking model generates an enumeration trace before the final count. 
Below, an unshuffled fragment contains four occurrences of the targets $\{\texttt{T,b,n}\}$:
%How do models develop the two distinct mechanisms during training? We present a controlled experiment where we train 4-layer 8-head models from scratch, using an implementation adapted from nanoGPT~\citep{karpathy2022nanogpt}. The training data are a modified Shakespeare corpus: each input sequence starts with \textcolor{promptblue}{a query}, then a 256-character chunk of Shakespeare text, followed by an optional \textcolor{thinkvermillion}{CoT trace} and then the \textcolor{ansgreen}{final count}. 
%an enumeration-type CoT trace (for training Thinking models only), and then the final count. 
%Both models share the architecture, input distribution, and training budget. Inputs are a character-shuffled 256-character window from Tiny Shakespeare with target counts from 1 to 10. For readability, we illustrate the sequence format with an unshuffled text fragment containing four occurrences of the target characters $\{\texttt{T,b,n}\}$ (boldfaced below): 

%    \YZ{Give a one-line example of the input sequence that includes the Shakespeare text, think-token, enumeration, answer token, and final count. Boldface the needle characters}
% Default vertical spacing.
\par
%\vspace{-0.5em}
% Default vertical spacing.
\begingroup
\begin{center}
\normalsize
\noindent
\texttt{\textcolor{promptblue}{<BOS> <CountChar> T b n <Sep>}}
\ldots\ \textbf{T}o \textbf{b}e, \ldots\
or \textbf{n}ot to \textbf{b}e.\ \ldots
\\*
\texttt{\textcolor{thinkvermillion}{<Think> <Sep> \textbf{T} <Sep> \textbf{b}
<Sep> \textbf{n} <Sep> \textbf{b} </Think>} \textcolor{ansgreen}{<Ans> <4> <EOS>}}
\end{center}
\par
%\vspace{-0.5em}
\endgroup
% Default vertical spacing.
% Default vertical spacing.

The trace enumerates occurrences of three target characters without numbering; Non-thinking instead outputs \texttt{\textcolor{ansgreen}{<Ans> <4> <EOS>}} directly. To limit overfitting on the small corpus, we use a two-stage autoregressive schedule. Both models train on full sequences for 1,500 steps, then on task outputs with equal final count loss weights (App.~\ref{app:synthetic-design}).
%The trace lists the matching characters in order, without numbering. The Non-thinking model outputs \texttt{\textcolor{ansgreen}{<Ans> <4> <EOS>}} directly after the input, without generating a trace. Each sequence selects three characters as needles to be counted in the text. See Appendix~\ref{app:synthetic-design} for dataset and training details. To limit overfitting on the small corpus, we use a two-stage autoregressive schedule. We train on the full sequence for 1,500 steps,
%the validation loss starts to plateau.
%then restrict the loss to task outputs, with the same final count loss weight in both models. All reported runs use this switching point.
%\YZ{It's a bit tricky to describe the change of training-loss change.} 
The learning dynamics reveal a distinction between the two models. Free-running count accuracy at the token after \texttt{<Ans>} reaches 95\% for Thinking but plateaus at 21\% for Non-thinking (Fig.~\ref{fig:emergence}). Further, broad retrieval heads emerge in the Non-thinking model’s first layer, while targeted retrieval heads emerge in the Thinking model’s last layer. 

As before, we apply the scores $B_h$ and $T_h$ with $N$ target occurrences and singleton spans $\mathcal S_k=\{s_k\}$. Broad retrieval is measured at \texttt{<Ans>}; targeted scores are averaged over the $N$ teacher-forced trace queries within each input. Both scores are then averaged equally across inputs (App.~\ref{app:synth-retrieval}). Additional representation analyses and state patching experiments are reported in App.~\ref{app:synth-representations}.
%The synthetic models also differ in final count readability; counter-state and answer state transplants test their roles in trace continuation and final count prediction respectively (Appendix~\ref{app:synth-representations}).
%\LS{The synthetic evidence is strongest for final count readout. Running index decoding and continuation interventions do not yet establish a stable counter update, and trace structure may contribute.} 
%\YZ{If space permits, we can put a small table to summarize the dynamics of other geometric metrics. Otherwise, refer to the appendix.}

%This synthetic experiment complements our analysis of pretrained LLMs with insights into data and training. 
This experiment provides insights into data and training.  First, the presence of enumeration traces, even without explicit numbering, appears to influence whether models learn broad or targeted retrieval. Second, standard autoregressive training on enumeration-type CoT traces can yield strong counting accuracy, even without specialized RL post-training. These results suggest the potential of high-quality CoT traces for improving reasoning capability, consistent with prior empirical findings \citep{zelikman2022star,muennighoff2025s1}.

%This synthetic experiment provides additional insights. First, models trained with enumeration traces develop targeted retrieval, whereas models trained to answer directly develop broad retrieval.
%First, as the only controlled factor, CoT traces in the training sequences determine which of the two retrieval mechanisms forms, and therefore whether the resulting model is Thinking or Non-thinking. 
%\LS{Maybe we need to claim more conservative, since this comparison jointly changes trace supervision, output length, loss composition, and generated computation. It does not isolate CoT content as the sole causal factor.} 
%Second, standard autoregressive training on enumeration-type CoT traces yields strong counting accuracy, even without specialized post-training. This reinforces a common belief that high-quality CoT traces are a key ingredient for strong reasoning capability. This result is consistent with prior work on learning from intermediate traces~\citep{nye2021scratchpads,zelikman2022star,muennighoff2025s1}. 
%\TY{ add a citation: ~\citep{muennighoff2025s1} }
%\YZ{search for papers and cite. I think a folklore is that data matters more, especially CoT traces are the secret sauce. It's interesting because we didn't do RLVR for improving counting ability here.}

%the two retrieval mechanisms measuring broad retrieval scores and targeted retrieval scores across training, we find that broad retrieval heads emerge in the first layer

\FloatBarrier
\section{Limitations and Future Work}\label{sec:limit}

In this work, we answered our motivating questions Q1 and Q2 directly with head-level and representation-level measurements. For Q3, our causal interventions provide evidence that counter states guide continued enumeration. The transferred states jointly carry count information and record content. More direct evidence would come from identifying circuits that implement the update as a needle is retrieved, which is left as future work.
Moreover, we used counting as a clean testbed for analyzing long-context reasoning in LLMs. Future work could extend this analysis to sophisticated reasoning involving search, branching, and the evolution of multiple states in model representations.

%\subsection*{Html version}

\subsection*{Reproducibility}
Code for reproducing the results in this paper is available at \url{https://github.com/Twist-Shan/CoT_NIAH_Counting}. In the appendix, we provide details about prompt construction, synthetic datasets, training hyperparameters, representation measurements, and additional supporting analyses.

\subsection*{Acknowledgments}
Y.Z.~is partially supported by NSF-DMS grant 2412052 and a Coefficient Giving (formerly Open Philanthropy) grant. In addition, support for this research was provided by the University of Wisconsin--Madison Office of the Vice Chancellor for Research with funding from the Wisconsin Alumni Research Foundation. This work also used Purdue Anvil AI system through allocation MTH260088 from the Advanced Cyberinfrastructure Coordination Ecosystem: Services \& Support (ACCESS) program at NSF \citep{boerner2023access, song2022anvil}. 
We also thank Haolin Yang and Quan Chen for helpful discussions.

\subsection*{AI Use Statement}
%AI tools (including ChatGPT 5.6 Sol, ChatGPT 6 Astra, etc.) were used to assist with polishing the writing and structuring the appendices; searching for relevant literature; generating code and organizing the code repository. All technical claims, experimental designs, implementations, and reported results were reviewed and verified by the authors, who take full responsibility for the final manuscript.

In this work, we used generative AI tools (including ChatGPT 5.6 Sol, ChatGPT 6 Astra, etc.) for searching for relevant literature, generating synthetic data sets, implementing analysis by generating code and organizing the code repository, polishing the writing and structuring the appendices, and editing diagrams in the figures. We have not used generative AI tools for designing the conceptual frameworks, proposing or refining hypotheses, designing research  methodology/experiments, interpreting results, or suggesting the paper structure. Theoretical proofs are not applicable to this work. 
We reviewed all technical claims, experimental designs, implementations, and reported results. We take responsibility for the final content of this work, including text, claims or artifacts produced with the aid of generative AI.

\bibliographystyle{plainnat}
\bibliography{references_no_url}

\clearpage

\appendix
\startcontents[appendices]

\section*{Appendix Contents}
\printcontents[appendices]{}{1}{}

\clearpage

% !TEX root = ../main.tex
\section{Details of the main-text figures}
\label{app:figure-details}

This section provides panel-level definitions, evaluation settings, and interpretation limits for the main-text figures, in order of appearance. In head-ablation experiments, a \emph{bank} is a fixed set of heads, the \emph{dose} $K$ is the number ablated, and a \emph{frozen prefix} consists of the first $K$ heads in a fixed ranking.

\paragraph{Figure~\ref{fig:main}: retrieval and representations}

% Example ID: V4_4_T10000_N10_seed1255.
Panels A--B use the same ten-record, approximately 10k-token Qwen3-8B input (seed 1255), selected for a complete CoT trace and distributed Non-thinking attention. Arrows and marker labels are schematic.

\noindent\textit{Panel A: prompt attention.}
Head L29H20 is measured at the final token of the supplied \texttt{Total:} prefix. The strip sums raw attention in 170 approximately equal-width bins, $a_b=\sum_{t\in B_b}\alpha_{q,t}$. Intensity is $\sqrt{\min(a_b/\tau,1)}$, with $\tau$ the 98.5th percentile ($\approx0.099$); the colorbar, labeled \emph{Attention mass}, reports raw bin mass.

\noindent\textit{Panel B: successive retrieval.}
For head L25H30, query $q_k$ precedes item $k$ (Marker $k-1$); Marker 0 denotes initialization. The three rows show Markers 2, 3, and 4 ($k=3,4,5$), matching the schematic. Each cell reports needle-normalized attention,
\[
    s_{k,j}=\frac{\sum_{t\in\operatorname{span}(N_j)}\alpha_{q_k,t}}
    {\sum_{r=1}^{10}\sum_{t\in\operatorname{span}(N_r)}\alpha_{q_k,t}},
    \qquad k\in\{3,4,5\},\quad j=1,\ldots,10.
\]
Each row sums to one over the ten complete needle spans; attention to other tokens is excluded from the denominator. Outlined cells mark the next needle, $j=k$, with rounded shares of 92\%, 89\%, and 90\% beneath the respective cells. Needles are indexed by order of occurrence; horizontal spacing does not represent token distance. Ellipses indicate omitted queries. The bottom colorbar, labeled \emph{Relative mass}, is linear from 0\% to 100\% and reports each needle's share of total needle attention; Panel A retains its different raw-bin-mass mapping. Blue arrows denote attention; dark-orange boxes label query positions.

\noindent\textit{Panel C: running index representations.}
To align item-end representations under a shared trace format, we use 30 paired $N=10$ inputs whose Thinking traces are correct, complete, and consecutively numbered. Format selection and the subsequent 20/10 fitting/evaluation split followed inspection of these data. On 100 held-out states per mode, nearest-centroid classification (NCC) with 16 whitened principal components (PCs) achieves 98\% accuracy at trace-item endpoints (Thinking, L31), compared with 46\% at prompt-record endpoints (Non-thinking, L13). The display projects these states onto separate three-PC bases fitted using the selection data, with numbered centroids. Cross-validated NCC layer selection and preprocessing are detailed in Appendix~\ref{app:cot-representations}. This exploratory comparison is conditioned on Thinking format and correctness and uses different endpoints and selected layers across modes. The separate PCA bases preclude direct coordinate comparisons; numbering and position remain possible contributors to index separability.

\paragraph{Figure~\ref{fig:empirical-accuracy-laws}: observed counting accuracy}
\phantomsection
\label{app:empirical-figure-details}

\noindent\textit{Panels A,B: accuracy versus target count.}
These panels summarize the short-context benchmark: 12 model comparison groups, each evaluated on 30 paired seeds per condition. At each count and passage length, circles show the median exact-count accuracy across groups, and vertical bars show the interquartile range across groups. Unparseable responses count as errors and remain in the denominator. Darker colors indicate longer passages (1k--20k). The count axis is logarithmic; lines use Gaussian-kernel smoothing in $\log_2 N$ (bandwidth 0.3) as visual guides.

\noindent\textit{Panels C,D: accuracy versus passage length.}
Each curve fixes $N\in\{3,5,8,10,15,20\}$ over 25k--100k, on a linear length axis. Straight segments connect mean accuracies over 30 paired seeds per condition without smoothing. Truncated and unparseable outputs count as errors. Rose solid lines denote Non-thinking; blue dashed lines denote Thinking. Darker shades indicate larger $N$. Panel C evaluates Qwen3-32B with YaRN disabled; Panel D evaluates Gemma-4-31B. Appendix~\ref{app:empirical-law} specifies the evaluation protocol; Appendix~\ref{app:empirical-long-observations} shows the complete count and length grids with pointwise 95\% Wilson intervals.

\paragraph{Figure~\ref{fig:nonthinking-broad-retrieval}: broad retrieval}

The head-selection cohort has no correctness filter (Table~\ref{tab:nonthinking-cohorts}); head selection is detailed in Appendix~\ref{app:nonthinking-head-scores}.

\noindent\textit{Panel A: needle-attention shares.}
Cells average the within-prompt shares $p_{hj}$ defined in Sec.~\ref{sec:broad-retrieval-score}, on a shared linear 0--25\% scale. Needles follow occurrence order; 10\% denotes equal shares. Averaged profiles do not imply uniform attention on individual inputs.

\noindent\textit{Panel B: layerwise scores.}
Each point is a head's mean full-span $B_h$; both models share the scale. Highlights retain the frozen head banks. Effective-coverage summaries use the complete banks across counts 2--10.

\paragraph{Figure~\ref{fig:nonthinking-form-retrieve-consolidate}: the causal pathway}

Protocols and populations are in Appendices~\ref{app:nonthinking-protocol} and~\ref{app:nonthinking-causal}. Bands and bars show pointwise 95\% seed-bootstrap intervals. Panels B--D report changes in normalized count error, normalized changes in generated count, and source-count match rates, respectively.

\noindent\textit{Panel A: mechanism schematic.}
The diagram summarizes state formation, retrieval, and consolidation; arrows and depths are illustrative.

\noindent\textit{Panel B: needle span patching.}
Clean-to-corrupted restoration (solid) and corrupted-to-clean patching (dashed) measure normalized error reduction and increase, respectively (Eq.~\ref{eq:nonthinking-span-effects}). Bootstrap draws pair seeds across directions.

\noindent\textit{Panel C: retrieval-head ablation.}
Filled circles show selected-bank ablations, diamonds layer-matched random banks, and hollow circles unmodified runs. Effects are normalized absolute count changes, $|\hat N_{\rm abl}-\hat N_{\rm clean}|/N$ (Appendix~\ref{app:nonthinking-retrieval}).

\noindent\textit{Panel D: answer state transfer.}
Proportion of predictions matching the source count after answer state patching (Eq.~\ref{eq:nonthinking-answer-match}). Every layer uses the same 24 Qwen and 26 Gemma directed pairs with correct source and target baseline answers. Invalid outputs count as failures.

\paragraph{Figure~\ref{fig:native-targeted-retrieval}: targeted attention}
\phantomsection
\label{app:cot-attention}

\noindent\textit{Panel A: examples and normalization.}
Both examples use indexed ten-item traces in approximately 10k-token passages (Qwen: seed 1255, L25H30; Gemma: seed 1240, L30H5). Marker 0 denotes initialization; Marker $k$ follows completed item $k$. Attention is normalized over complete needle spans so that each row sums to one, with a shared linear 0--100\% scale. Circles mark the designated next needle.

\noindent\textit{Panel B: scores and membership.}
Raw $T_h$ is averaged over eligible transitions within each selection seed, then equally across contributing seeds. For Qwen, 15 of 20 seeds contribute transitions with a rank immediately before the city; the query follows that marker. For Gemma, 19 of 20 seeds contribute transitions with the rank and city in the same text unit; the query is at the preceding item endpoint. The plots use a shared scale for all 1,152 Qwen heads and the 56 Gemma global-attention heads. Different formats, queries, and head coverage limit cross-model magnitude comparisons. Highlights preserve frozen banks; Gemma's plotted scores need not reproduce its bank's selection ranking.

\paragraph{Figure~\ref{fig:native-retrieve-encode-count-loop}: retrieval and counter states}

Populations and protocols are in Appendices~\ref{app:cot-protocol} and~\ref{app:cot-causal}. The source supplies the state transplanted into the target.

\noindent\textit{Panel A: notation and scope.}
$N_k$ is record $k$ at $\tau_k$, and $e_k$ its trace item. $H_k$ collects residual states over item $k$, ending at $t_k$; $h_{t_k}$ is its endpoint state. $h_{q_k}$ is the retrieval-query state. In the pretrained-model schematics, \texttt{<Ans>} labels the answer-query position $T=q_A$, with state $a_T$. Layer indices are omitted where unambiguous. Arrows summarize functional relations supported by local tests on the populations in Appendix~\ref{app:cot-causal}; the diagram does not establish a fully mediated chain on one common cohort.

\noindent\textit{Panel B: ablation.}
Ten seeds/model compare persistent masking of the frozen bank (filled circles), three layer-matched random banks (diamonds, averaged), and unmodified runs (hollow circles). Effects are increases in next-item or final-count failure rates relative to unmodified runs, expressed in percentage points; bootstrap draws pair conditions by seed.

\noindent\textit{Panel C: counter-state transfer.}
The illustration uses Qwen seed 1307. The table pools forward patches from source item $k$ to target item $k-1$, with $k=4,6,8$ and ten seeds per model. Later columns report conditional success fractions among trials that succeeded at all preceding steps and have another item remaining (Appendix~\ref{app:cot-progress}).

\noindent\textit{Panel D: answer transfer.}
Source-count match rates use the same 40 directed pairs per model at every decoder layer, with correct source and target baseline answers. Invalid outputs count as failures.

\paragraph{Figure~\ref{fig:emergence}: synthetic learning dynamics}

Each mode uses one four-layer, eight-head Transformer trained with seed 1234 (Appendix~\ref{app:synthetic-details}).

\noindent\textit{Panel A: counting accuracy.}
Free-running count accuracy is evaluated at eleven pre-specified checkpoints on the same 100 held-out inputs, ten per count $N=1,\ldots,10$. This comparison uses one training seed per mode and has no error bars.

\noindent\textit{Panels B--C: retrieval scores.}
The same inputs are evaluated every 100 steps, including step 0 (101 checkpoints). Non-thinking uses the broad retrieval score $B_h$; Thinking uses the targeted retrieval score $T_h$ from teacher-forced query-to-source attention (Appendix~\ref{app:synth-retrieval}). All 32 heads are ordered by layer, then head; each mode has a fixed raw-score color scale.

% !TEX root = ../main.tex
\section{Additional related work}

\label{app:additional-related-work}

\paragraph{Inspiration for our figures.}
Here we want to highlight the inspiration for our figures. The work on induction heads \citep{olsson2022induction}, circuit tracing \citep{ameisen2025circuit}, and attention tracing \citep{kamath2025tracing} inspired how we illustrate attention and information flow in Figs.~\ref{fig:main}A,B, \ref{fig:nonthinking-form-retrieve-consolidate}A, and~\ref{fig:native-retrieve-encode-count-loop}A.
Our plots of count representations (Fig.~\ref{fig:main}C and appendix) were inspired by the manifold plots in \citet{gurnee2025when}.
The scaling-law plots of \citet{kaplan2020scaling} also informed how we present the performance comparisons in Fig.~\ref{fig:empirical-accuracy-laws}.

% !TEX root = ../main.tex
\section{Further details for the behavioral comparison}
\label{app:empirical-law}

\subsection{Prompts and experimental design}
\label{app:behavioral-prompts}
\label{app:empirical-model-results}

\paragraph{Passage construction.}
For the behavioral benchmark, we use a fixed corpus of 154 Paul Graham essays collected from the RULER source list~\citep{hsieh2024ruler}. Seeded shuffling and concatenation of distinct essays provide the background text. We sample $N$ cities without replacement from a fixed 100-city vocabulary (\texttt{data/entities/cities.csv}) and $N$ distinct integer scores from 50 to 100, then insert the paired records at randomly sampled sentence boundaries, with record starts constrained to 5--95\% of the final passage's character length. We adjust the background prefix length so the passage, including all records, contains exactly $L$ tokens under the common tokenizer. Each $(N,L,\mathrm{seed})$ passage is frozen and reused across models and modes.

The behavioral and mechanism experiments share the following city--score counting template.
The placeholder \promptvar{passage} denotes the complete experimental passage, and \promptvar{response\_\allowbreak{}instruction} is replaced by the instruction for the selected mode.
Box titles and role labels are annotations; angle-bracketed output placeholders are literal parts of the instructions sent to the model.

The behavioral comparison uses these Non-thinking and Thinking instructions.
Structured bullet enumeration replaces only \promptvar{response\_instruction} with the block in Appendix~\ref{app:enumeration-prompts}; its behavioral and mechanism experiments use identical wording.
Appendices~\ref{app:nonthinking-prompts} and~\ref{app:native-prompts} list the remaining protocol-specific changes. All messages use the registered checkpoint's chat template with \texttt{add\_\allowbreak generation\_\allowbreak prompt=True}. The behavioral protocol supplies no additional assistant prefix.

% Source: Realistic_CoT_NiaH_Count/src/realistic_niah/prompts.py::build_messages
\begin{promptbox}[unbreakable]{Common counting template}
\setlength{\parskip}{1.5pt}
% Prompt fragment: C-common
\promptlabel{User message}
You will need to count all city-score audit records in the passage below.\par
A city-score audit record names one city and gives that city's numeric score.\par
\smallskip
<passage>\par
\{passage\}\par
</passage>\par
\smallskip
How many city-score audit records are in the passage?\par
\{response\_instruction\}\par
\end{promptbox}

\begin{promptbox}[unbreakable]{Response instructions}
\setlength{\parskip}{1.5pt}
% Prompt fragment: C-direct-instruction
\promptlabel{Non-thinking}
Do not explain, reason aloud, quote, or list any records.\par
Your entire response must be exactly one line:\par
Total: <integer>\par
\promptdivider
% Prompt fragment: C-native-instruction
\promptlabel{Thinking}
Reason concisely without repeating or restarting.\par
Stop as soon as you determine the count, then output exactly one line:\par
Total: <integer>\par
\end{promptbox}

% Exact additional Ministral 3 Reasoning system instruction (source documentation):
% # HOW YOU SHOULD THINK AND ANSWER
% 
% First draft your thinking process (inner monologue) until you arrive at a response. Format your response using Markdown, and use LaTeX for any mathematical equations. Write both your thoughts and the response in the same language as the input.
% 
% Your thinking process must follow the template below:[THINK]Your thoughts or/and draft, like working through an exercise on scratch paper. Be as casual and as long as you want until you are confident to generate the response to the user.[/THINK]Here, provide a self-contained response.

\paragraph{A complete 1k-token passage example.}
\label{app:behavioral-1k-example}
The example below uses a frozen behavioral stimulus with $N=2$, $L=1{,}000$, and seed 1234. The passage, including both inserted records, contains exactly 1,000 tokens under the common Qwen3-8B tokenizer; task instructions, passage delimiters, and chat-template tokens are excluded from this length. The background is drawn from the Paul Graham essay corpus used in our benchmark~\citep{kamradt2023needle,hsieh2024ruler}. We show the Thinking user message; Non-thinking uses the same passage with the response instruction above. Boldface highlights the inserted records for presentation only. The original passage ending is retained, even though it falls within a sentence.

\begin{promptbox}{Thinking input: a 1k-token passage with two records}
\setlength{\parskip}{3pt}
\promptlabel{User message}
You will need to count all city-score audit records in the passage below.\par
A city-score audit record names one city and gives that city's numeric score.
\par\smallskip
\prompttag{<passage>}\par
Want to start a startup? Get funded by Y Combinator. July 2010 I realized recently that what one thinks about in the shower in the morning is more important than I'd thought. I knew it was a good time to have ideas. Now I'd go further: now I'd say it's hard to do a really good job on anything you don't think about in the shower. Everyone who's worked on difficult problems is probably familiar with the phenomenon of working hard to figure something out, failing, and then suddenly seeing the answer a bit later while doing something else. There's a kind of thinking you do without trying to. I'm increasingly convinced this type of thinking is not merely helpful in solving hard problems, but necessary. The tricky part is, you can only control it indirectly. [1] I think most people have one top idea in their mind at any given time. That's the idea their thoughts will drift toward when they're allowed to drift freely. And this idea will thus tend to get all the benefit of that type of thinking, while others are starved of it. \promptrecord{In the 2024 city score audit, Seoul received a score of 86.} Which means it's a disaster to let the wrong idea become the top one in your mind. What made this clear to me was having an idea I didn't want as the top one in my mind for two long stretches. I'd noticed startups got way less done when they started raising money, but it was not till we ourselves raised money that I understood why. The problem is not the actual time it takes to meet with investors. The problem is that once you start raising money, raising money becomes the top idea in your mind. That becomes what you think about when you take a shower in the morning. And that means other questions aren't. I'd hated raising money when I was running Viaweb, but I'd forgotten why I hated it so much. When we raised money for Y Combinator, I remembered. Money matters are particularly likely to become the top idea in your mind. The reason is that they have to be. It's hard to get money. It's not the sort of thing that happens by default. It's not going to happen unless you let it become the thing you think about in the shower. And then you'll make little progress on anything else you'd rather be working on. [2] (I hear similar complaints from friends who are professors. Professors nowadays seem to have become professional fundraisers who do a little research on the side. It may be time to fix that.) The reason this struck me so forcibly is that for most of the preceding 10 years I'd been able to think about what I wanted. So the contrast when I couldn't was sharp. But I don't think this problem is unique to me, because just about every startup I've seen grinds to a halt when they start raising money --- or talking to acquirers. You can't directly control where your thoughts drift. \promptrecord{In the 2024 city score audit, Beijing received a score of 82.} If you're controlling them, they're not drifting. But you can control them indirectly, by controlling what situations you let yourself get into. That has been the lesson for me: be careful what you let become critical to you. Try to get yourself into situations where the most urgent problems are ones you want to think about. You don't have complete control, of course. An emergency could push other thoughts out of your head. But barring emergencies you have a good deal of indirect control over what becomes the top idea in your mind. I've found there are two types of thoughts especially worth avoiding --- thoughts like the Nile Perch in the way they push out more interesting ideas. One I've already mentioned: thoughts about money. Getting money is almost by definition an attention sink. The other is disputes. These too are engaging in the wrong way: they have the same velcro-like shape as genuinely interesting ideas, but without the substance. So avoid disputes if you want to get real work done. [3] Even Newton fell into this trap. After publishing his theory of colors in 1672 he found himself distracted by disputes for years, finally concluding that the only solution was to stop publishing: I see I have made myself a slave to Philosophy, but if I get free of Mr Linus's business I will resolutely bid adew to it eternally, excepting what I do for my privat satisfaction or leave to come out after me. For I see a man must either resolve to put out nothing new or become a slave to defend it. [4] Linus and his students at Liege were among the more tenacious critics. Newton's biographer Westfall seems to feel he was overreacting: Recall that at the time he wrote, Newton's "
\par\prompttag{</passage>}
\par\smallskip
How many city-score audit records are in the passage?\par
Reason concisely without repeating or restarting.\par
Stop as soon as you determine the count, then output exactly one line:\par
Total: <integer>
\end{promptbox}
\noindent\textit{Reference answer (not part of the input):} \texttt{Total: 2}.

\paragraph{Models and comparison groups.}
We evaluate 12 comparison groups including \href{https://huggingface.co/Qwen/Qwen3-4B}{Qwen3-4B}, \href{https://huggingface.co/Qwen/Qwen3-8B}{Qwen3-8B}, \href{https://huggingface.co/Qwen/Qwen3-14B}{Qwen3-14B}, and \href{https://huggingface.co/Qwen/Qwen3-32B}{Qwen3-32B}~\citep{qwen3technicalreport}; \href{https://huggingface.co/google/gemma-4-E4B-it}{Gemma-4-E4B}, \href{https://huggingface.co/google/gemma-4-12B-it}{Gemma-4-12B}, \href{https://huggingface.co/google/gemma-4-26B-A4B-it}{Gemma-4-26B-A4B}, and \href{https://huggingface.co/google/gemma-4-31B-it}{Gemma-4-31B}~\citep{gemmateam2026gemma4}; \href{https://huggingface.co/nvidia/NVIDIA-Nemotron-Nano-9B-v2}{NVIDIA-Nemotron-Nano-9B-v2} and \href{https://huggingface.co/nvidia/NVIDIA-Nemotron-3-Nano-4B-BF16}{NVIDIA-Nemotron-3-Nano-4B}~\citep{nvidia2025nemotronnano2,nvidia2025nemotron3,taghibakhshi2025nemotronelastic}; the \href{https://huggingface.co/zai-org/GLM-4-9B-0414}{GLM-4-9B-0414}/\href{https://huggingface.co/zai-org/GLM-Z1-9B-0414}{GLM-Z1-9B-0414} pair~\citep{glmteam2024chatglm}; and the \href{https://huggingface.co/mistralai/Ministral-3-8B-Instruct-2512}{Ministral 3 8B Instruct}/\href{https://huggingface.co/mistralai/Ministral-3-8B-Reasoning-2512}{Ministral 3 8B Reasoning} pair~\citep{liu2026ministral3}.
Model names link to the official Hugging Face model cards for the corresponding checkpoints.
Qwen3, Gemma-4, and Nemotron-3-Nano-4B use \texttt{enable\_\allowbreak thinking} to switch modes; enumeration disables the built-in Thinking mode.
Nemotron-Nano-v2-9B instead receives the system message \texttt{/no\_think} or \texttt{/think}.
GLM and Ministral use their paired Non-thinking and reasoning checkpoints, with the registered system instruction added for Ministral 3 Reasoning.

\paragraph{Grids and scoring.}
The short-context benchmark has 12 comparison groups, two modes, 14 counts $N\in\{1,\ldots,10,12,15,18,20\}$, and eight lengths $L\in\{1,2,3,5,8,10,15,20\}$k. Each condition contains 30 paired seeds, giving 80,640 requests. Parsed exact accuracy requires the final integer matching \texttt{Total:} to equal $N$; unparseable responses receive zero and all requests remain in the denominator. The long-context grid includes 25, 30, 40, 50, 60, 70, 80, 90, and 100k. Qwen3-32B is evaluated over all 17 lengths with YaRN disabled (14,280 requests); Gemma-4-31B is evaluated in separate 1k--20k and 25k--100k batches. These length analyses count truncations as errors, with output limits of 64 tokens for Non-thinking and 4,096 for Thinking. Qwen's manifests verify default RoPE and a 131,072-position cache. Passage lengths use the common Qwen3-8B tokenizer.

\paragraph{Decoding.}
Non-thinking uses greedy decoding (temperature 0, top-$p=1$, top-$k$ disabled). Thinking uses top-$p=0.95$ throughout, with temperature 0.6 for Qwen3, GLM-Z1, and Nemotron-Nano-v2-9B; 1.0 for Gemma-4 and Nemotron-3-Nano-4B; and 0.7 for Ministral 3 Reasoning. Thinking uses top-$k=20$ for Qwen3, 64 for Gemma-4, and 40 for GLM-Z1; top-$k$ is disabled for the other groups. All runs use \texttt{min\_p=0} and the corresponding passage seed for generation.

\subsection{Non-thinking: relative noise}
\label{app:empirical-relative-noise}

\paragraph{Representation measurement.}
We use the mechanism cohort in Appendix~\ref{app:nonthinking-protocol}, retaining all inputs: fitting seeds 1234--1253 and held-out seeds 1254--1263 at each $N=1$--10.
States are measured at the final token of the supplied \texttt{Total:} prefix.
Each seed fixes the background and ten candidate slots, activating the first $N$; the spatial extent of active needles can still vary with $N$.

Let $\overline h_j^{\mathrm{fit}}$ be the mean state on the fitting data at count $j$.
For each adjacent pair, define a unit direction $u_n$, then compute the held-out mean $\mu_{j\mid n}$ and unbiased sample variance $s^2_{j\mid n}$ of $u_n^\top h$ at each count $j\in\{n,n+1\}$:
\begin{equation}
u_n=\frac{\overline h_{n+1}^{\mathrm{fit}}-\overline h_n^{\mathrm{fit}}}
{\|\overline h_{n+1}^{\mathrm{fit}}-\overline h_n^{\mathrm{fit}}\|_2},
\qquad
\eta_n=\frac{\sqrt{(s^2_{n\mid n}+s^2_{n+1\mid n})/2}}
{|\mu_{n+1\mid n}-\mu_{n\mid n}|},\quad n=1,\ldots,9.
\label{eq:empirical-relative-noise}
\end{equation}
Thus $\eta_n$ is inverse absolute $d'$ along this direction, using two classes at every point.
It measures variability across inputs, rather than repeated generations of one prompt.

\begin{figure}[H]
\centering
\includegraphics[width=0.9\linewidth]{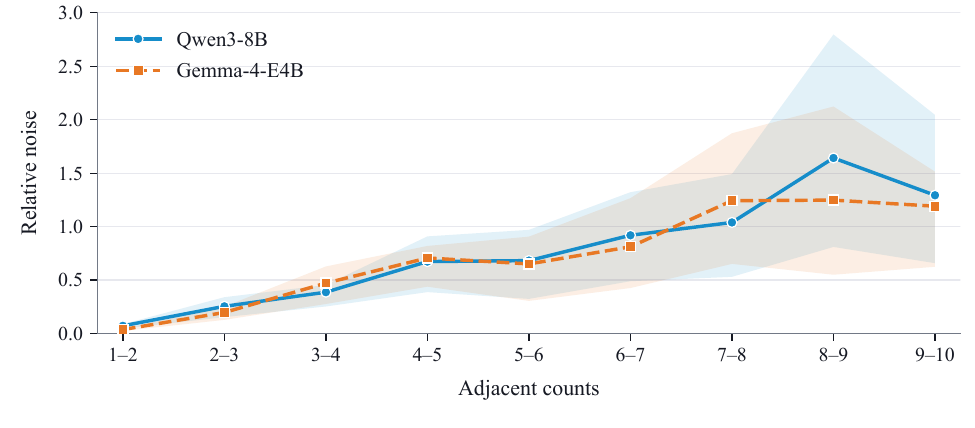}
\caption{\textbf{Relative noise between adjacent counts.} Non-thinking answer-query states at 10k tokens: Qwen3-8B L25 and Gemma-4-E4B L36, selected by cross-validated NCC (Appendix~\ref{app:nonthinking-geometry}). Relative noise is pooled within-count standard deviation divided by projected mean separation (Eq.~\eqref{eq:empirical-relative-noise}); larger values indicate poorer separation. Directions use 20 fitting seeds; estimates use ten held-out seeds per count. Lines connect observations. Shading shows pointwise 95\% intervals from 10,000 bootstrap draws, independently resampling whole-seed profiles in each split, refitting directions, and fixing layers.}
\label{fig:empirical-relative-noise}
\end{figure}

\paragraph{Results.}
Relative noise increases overall, with nonmonotone endpoints (Fig.~\ref{fig:empirical-relative-noise}).
The high/low ratio $\sqrt{(\eta_8^2+\eta_9^2)/(\eta_1^2+\eta_2^2)}$ is 7.93 for Qwen (95\% interval 4.76--13.60) and 8.64 for Gemma (4.35--14.74); last-block ratios are 7.98 and 9.03.
The increase mainly reflects shrinking mean separation: its high/low RMS ratios are 0.057 and 0.106, versus 0.65 and 1.25 for projected standard deviation.
We conjecture that broad retrieval accumulates noise as needle count increases. The current results support poorer local separation, without establishing growth in absolute noise or the model's use of this direction.
The analysis reuses the mechanism cohort and is exploratory; its intervals are not simultaneous bands.

\subsection{Count dependence: models, fits, and complete group results}
\label{app:empirical-regression}
\label{app:empirical-cross-model-fits}

\paragraph{Two approximations.}
For Non-thinking, express the scalar readout in count units, with neighboring count centers at consecutive integers.
Write $Z=N+\sigma(N,L)G$, where $G\sim\mathcal N(0,1)$ and $\sigma(N,L)$ is the noise standard deviation in these units.
Rounding is correct when $|Z-N|<1/2$, yielding
\begin{equation}
p_{\mathrm{NT}}(N,L)=2\Phi\!\left(\frac{1}{2\sigma(N,L)}\right)-1,
\qquad \sigma(N,L)=N\sigma(1,L).
\label{eq:empirical-gaussian}
\end{equation}
Here $\Phi$ is the standard normal cumulative distribution function.
We fit one positive value of $\sigma(1,L)$ per length; it is the noise standard deviation at $N=1$.
This is the scalar-variability approximation discussed in numerical-cognition models~\citep{dehaene2003weber}.
Measuring noise in count units absorbs the spacing between count centers; accuracy identifies only noise relative to this spacing.
The assumption $\sigma\propto N$ does not follow from attention averaging alone.
% Previous notation retained:
% For Thinking, assume $N$ independent required steps with success probability $1-\varphi_L$, an independent baseline failure $\alpha$, and no cancellation of step errors. Then
For Thinking, assume $N$ independent required steps with success probability $1-\gamma_L$, an independent baseline failure $\alpha$, and no cancellation of step errors. Then
\begin{equation}
% Previous notation retained:
% p_{\mathrm{T}}(N,L)=(1-\alpha)(1-\varphi_L)^N
p_{\mathrm{T}}(N,L)=(1-\alpha)(1-\gamma_L)^N
=\exp[-h_0-N\lambda_L],
\label{eq:empirical-step-product}
\end{equation}
% Previous notation retained:
% where $h_0=-\ln(1-\alpha)$ and $\lambda_L=-\ln(1-\varphi_L)$ are nonnegative.
where $h_0=-\ln(1-\alpha)$ and $\lambda_L=-\ln(1-\gamma_L)$ are nonnegative.
% Previous notation retained:
% Taylor's theorem gives $0\leq\alpha+(1-\alpha)N\varphi_L-(1-p_{\mathrm{T}})\leq(1-\alpha)\binom N2\varphi_L^2$.
Taylor's theorem gives $0\leq\alpha+(1-\alpha)N\gamma_L-(1-p_{\mathrm{T}})\leq(1-\alpha)\binom N2\gamma_L^2$.
% Previous notation retained:
% Thus the main-text linear error approximation requires $N\varphi_L\ll1$; all fitted predictions use the bounded product.
Thus the main-text linear error approximation requires $N\gamma_L\ll1$; all fitted predictions use the bounded product.

\paragraph{Connection to the main-text heuristic.}
% Previous notation retained:
% The main-text parameter $\gamma_L$ corresponds to $\varphi_L$ here. Setting $\alpha=0$ in Eq.~\eqref{eq:empirical-step-product} gives $(1-\gamma_L)^N$ and its first-order approximation $1-N\gamma_L$. The fitted model retains $\alpha$ to allow a count-independent baseline failure component. Under the assumed step-success model, increasing per-step failure at fixed $N$ lowers counting accuracy; in the small-error regime, its contribution to final error scales approximately with $N$. These fits use final-answer outcomes, so $\varphi_L$ is an effective parameter rather than a directly measured retrieval failure rate. Fit quality does not establish independent step failures or identify why retrieval degrades with length; Appendix~\ref{app:empirical-length-tests} compares descriptive length functions.
Setting $\alpha=0$ in Eq.~\eqref{eq:empirical-step-product} gives $(1-\gamma_L)^N$ and its first-order approximation $1-N\gamma_L$. The fitted model retains $\alpha$ to allow a count-independent baseline failure component. Under the assumed step-success model, increasing per-step failure at fixed $N$ lowers counting accuracy; in the small-error regime, its contribution to final error scales approximately with $N$. These fits use final-answer outcomes, so $\gamma_L$ is an effective parameter rather than a directly measured retrieval failure rate. Fit quality does not establish independent step failures or identify why retrieval degrades with length; Appendix~\ref{app:empirical-length-tests} compares descriptive length functions.

\paragraph{Estimation and validation.}
Nonlinear least squares fits each group separately and, separately, the descriptive cross-model medians.
For equal-size condition means, squared error is equivalent to request-level Brier loss up to a constant.
At $n_L$ lengths, Non-thinking has $n_L$ free parameters and Thinking $n_L+1$.
Validation holds out every condition at one count and refits (14 folds), testing new counts at measured lengths.
For observed accuracies $a_c$ and out-of-fold predictions $\widehat p_c$, $R^2=1-\sum_c(a_c-\widehat p_c)^2/\sum_c(a_c-\overline a)^2$ and $\mathrm{RMSE}=100\sqrt{\operatorname{mean}_c(a_c-\widehat p_c)^2}$ in percentage points.
The analysis is exploratory; shared seeds preclude treating conditions as independent samples for confidence intervals.

\begin{figure}[H]
\centering
\includegraphics[width=\linewidth]{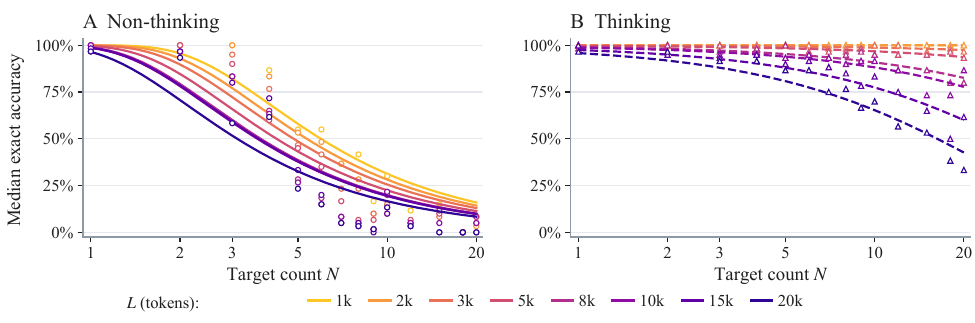}
\caption{\textbf{Count-model fits to unsmoothed cross-model medians.} All 14 counts and eight lengths are shown. Curves use Eqs.~\eqref{eq:empirical-gaussian} and~\eqref{eq:empirical-step-product}.}
\label{fig:empirical-median-fits}
\end{figure}

\begin{figure}[H]
\centering
\includegraphics[width=\linewidth]{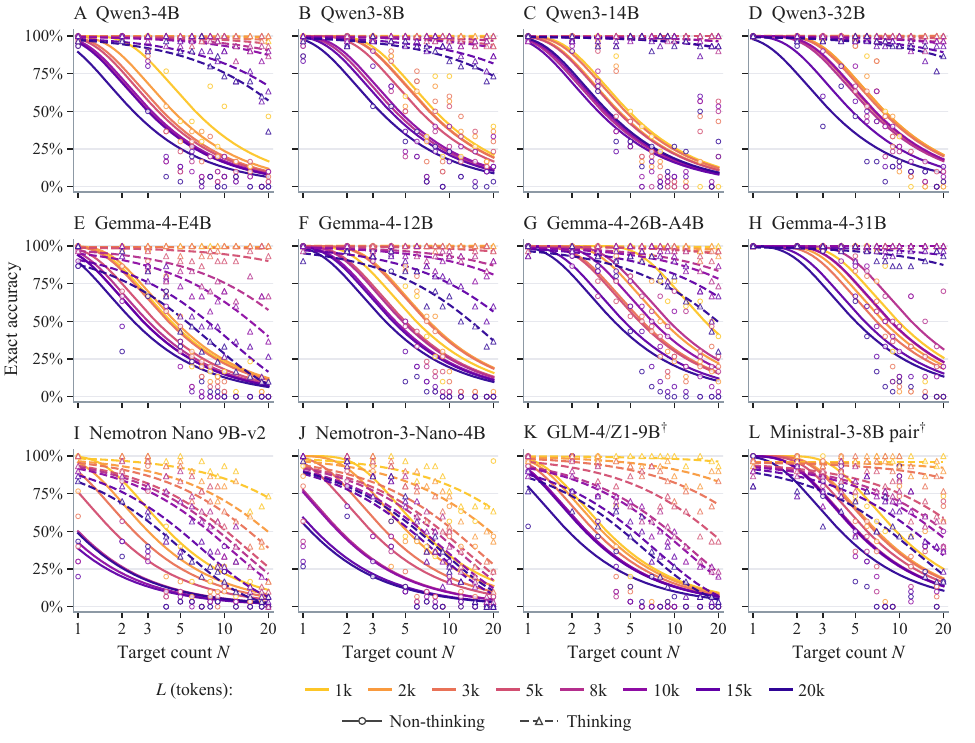}
\caption{\textbf{Complete 12-group comparison at 1k--20k.} Every panel includes 14 counts, eight lengths, and both modes. Points are 30-request accuracies; solid Gaussian curves and circles denote Non-thinking, dashed step-success curves and triangles Thinking. Colors identify length. Fits are separate for each group. $^\dagger$Separate checkpoints.}
\label{fig:empirical-all-models}
\end{figure}

For fits to the cross-model median accuracies, training/held-count $R^2$ is $0.858/0.828$ for Non-thinking and $0.942/0.924$ for Thinking, with held-count RMSE 15.2 and 3.6 points.
The Non-thinking transition shows systematic departures.
Thinking's overall mean is higher in every group (Table~\ref{tab:empirical-model-results}); Fig.~\ref{fig:empirical-all-models} gives all group-level observations and fits.

\begin{table}[H]
\centering
\caption{\textbf{Observed means and held-count fits.} The 1k--20k and 1k--100k grids contain 3,360 and 7,140 requests per mode per group, respectively. RMSE: percentage points. $^\dagger$Separate checkpoints.}
\renewcommand{\arraystretch}{0.90}
\setlength{\tabcolsep}{3.5pt}
\begin{tabular}{lrrrrrr}
\toprule
& \multicolumn{2}{c}{Mean accuracy (\%)} & \multicolumn{2}{c}{Non-thinking} & \multicolumn{2}{c}{Thinking}\\
\cmidrule(lr){2-3}\cmidrule(lr){4-5}\cmidrule(lr){6-7}
Comparison group & NT & T & CV $R^2$ & RMSE & CV $R^2$ & RMSE\\
\midrule
Qwen3-4B & 31.8 & 94.4 & 0.750 & 17.9 & 0.782 & 4.9\\
Qwen3-8B & 45.3 & 96.5 & 0.739 & 17.0 & 0.778 & 2.8\\
Qwen3-14B & 35.5 & 98.6 & 0.663 & 20.7 & 0.293 & 2.3\\
Qwen3-32B & 47.7 & 98.0 & 0.816 & 16.3 & 0.508 & 2.5\\
Gemma-4-E4B & 29.2 & 79.4 & 0.730 & 18.2 & 0.927 & 7.1\\
Gemma-4-12B & 39.6 & 91.6 & 0.746 & 21.1 & 0.920 & 4.0\\
Gemma-4-26B-A4B & 51.8 & 93.1 & 0.840 & 13.8 & 0.759 & 6.2\\
Gemma-4-31B & 54.8 & 98.8 & 0.751 & 19.8 & 0.463 & 2.2\\
Nemotron Nano v2-9B & 17.9 & 60.4 & 0.476 & 19.5 & 0.877 & 10.9\\
Nemotron Nano 3-4B & 28.4 & 59.8 & 0.421 & 23.4 & 0.862 & 12.2\\
GLM-4/Z1-9B$^\dagger$ & 21.8 & 67.3 & 0.633 & 22.1 & 0.948 & 6.7\\
Ministral-3-8B pair$^\dagger$ & 50.1 & 78.1 & 0.446 & 27.5 & 0.817 & 8.7\\
\midrule\multicolumn{7}{l}{1k--100k grid}\\
Qwen3-32B & 29.1 & 72.7 & 0.813 & 15.6 & 0.952 & 7.7\\
Gemma-4-31B & 39.7 & 87.6 & 0.727 & 22.3 & 0.913 & 5.8\\
\bottomrule
\end{tabular}
\label{tab:empirical-model-results}
\end{table}

\FloatBarrier

\subsection{Complete long-context observations}
\label{app:empirical-long-context}
\label{app:empirical-long-observations}

\begin{figure}[H]
\centering
\includegraphics[width=\linewidth]{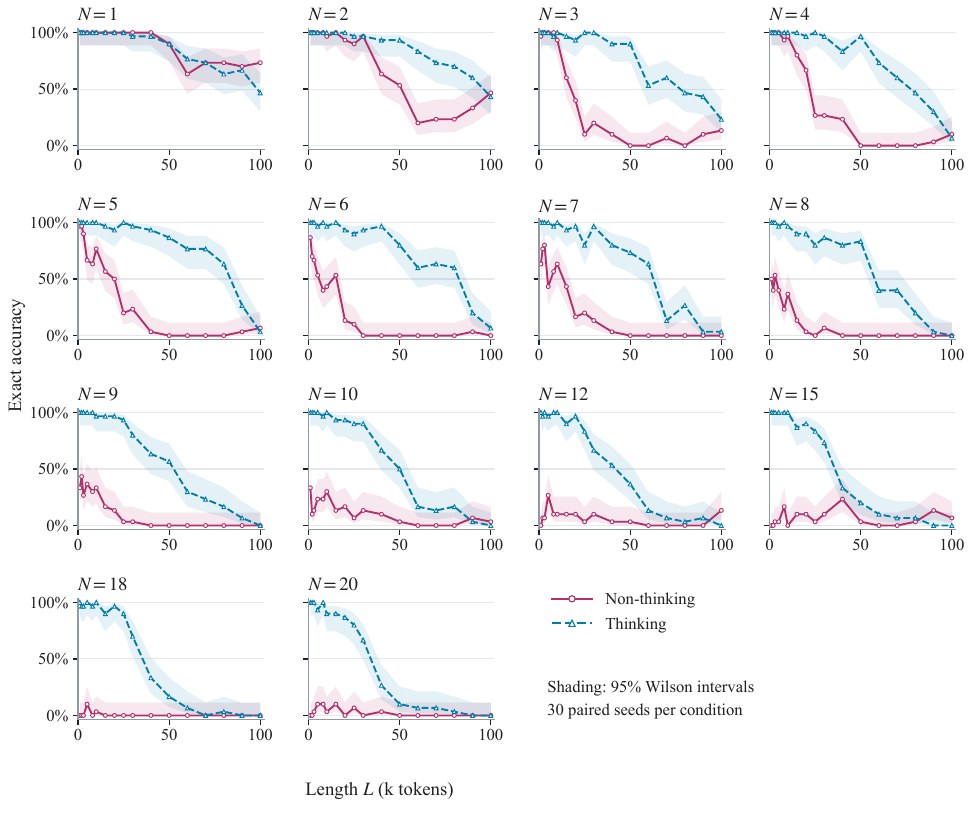}
\caption{\textbf{Qwen3-32B: all counts and lengths, YaRN disabled.} Each panel fixes $N$ and shows all 17 lengths from 1k to 100k. Rose circles/solid lines are Non-thinking; blue triangles/dashed lines are Thinking. Points are 30-seed means, shading shows pointwise 95\% Wilson intervals, and lines connect measurements without smoothing. Truncations count as errors.}
\label{fig:empirical-qwen-all-counts}
\label{fig:empirical-full-range}
\end{figure}

\begin{figure}[H]
\centering
\includegraphics[width=\linewidth]{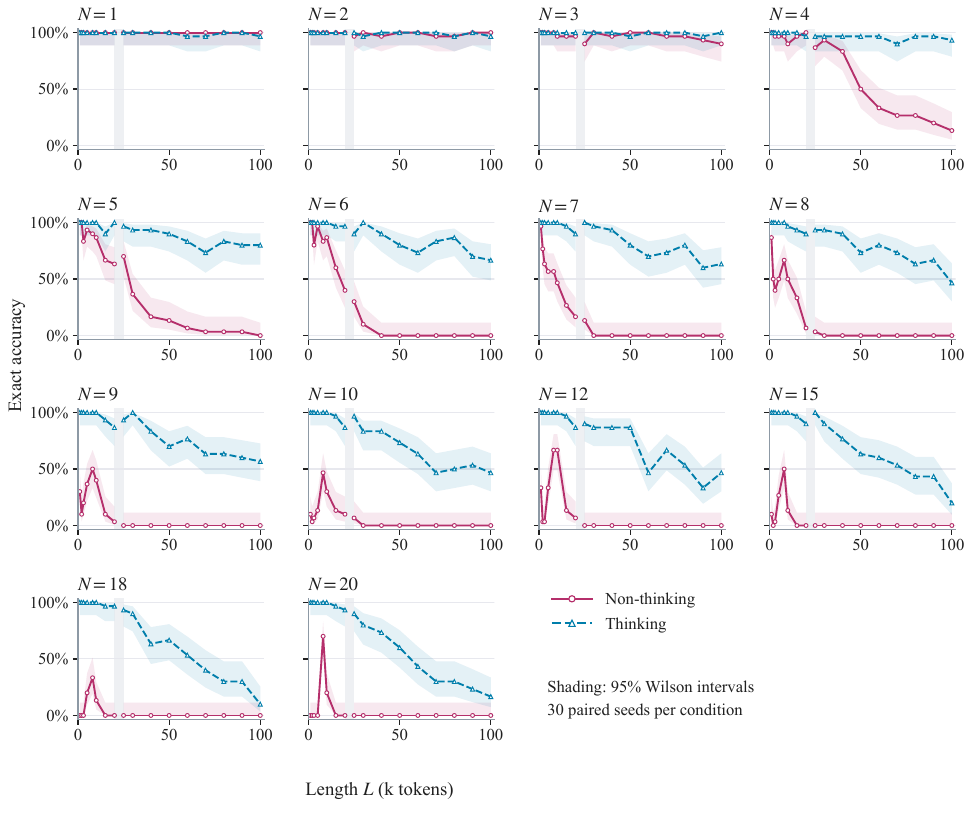}
\caption{\textbf{Gemma-4-31B: all counts and lengths.} The display matches Fig.~\ref{fig:empirical-qwen-all-counts}. Gray strips separate the 1k--20k and 25k--100k evaluation batches; lines and intervals do not cross that gap. All 14 counts and both modes are retained, including local accuracy reversals.}
\label{fig:empirical-gemma-all-counts}
\end{figure}

Figures~\ref{fig:empirical-qwen-all-counts} and~\ref{fig:empirical-gemma-all-counts} expand main panels C,D to all 14 counts, 17 lengths, and both modes (476 conditions/model).
Shading gives pointwise 95\% Wilson intervals for 30 binary outcomes; these are not simultaneous bands or paired tests across conditions.

Local reversals remain visible.
At $N=1,L=100$k, Qwen's Non-thinking and Thinking accuracies are $22/30$ and $14/30$; 12 Thinking outputs exhaust the 4,096-token limit and four give incorrect totals.
Of these 12 truncated traces, nine show repeated phrases, while the other three continue listing years/scores, paragraph indices, or supposed records.
Gemma Non-thinking at $N=20$ is correct on $0/30$ requests at 5k, $21/30$ at 8k, and $0/30$ at 15k.
These results delimit the overall trends: output budgets contribute to failure, and the cause of Gemma's local increase is not identified.
\FloatBarrier

\FloatBarrier

\begin{figure}[H]
\centering
\includegraphics[width=\linewidth]{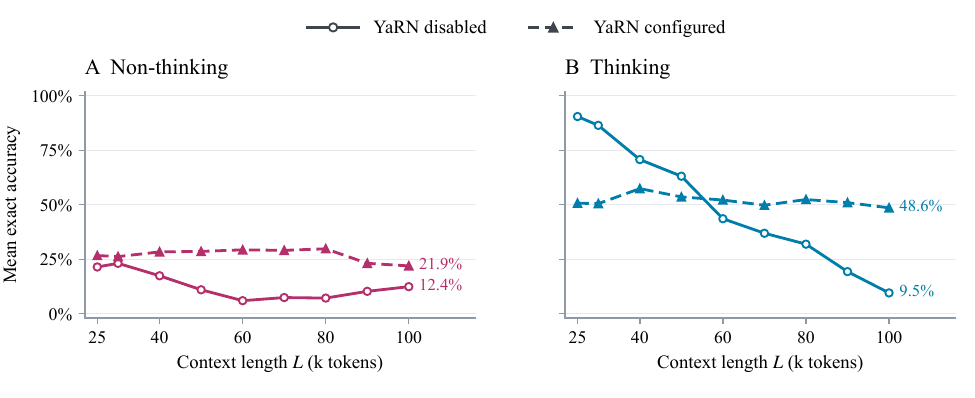}
\caption{\textbf{Qwen3-32B long-context accuracy with and without YaRN.} Panels show Non-thinking (rose) and Thinking (blue). Solid circles: YaRN disabled; dashed triangles: YaRN configured. Each point pools 14 counts with 30 seeds each (420 requests); truncations and unparseable outputs count as errors. Lines connect observations without smoothing. The runs also differ in execution settings (request batch size 2 versus 1), so gaps do not isolate YaRN's causal effect.}
\label{fig:empirical-yarn}
\end{figure}

\paragraph{Long-context evaluation with YaRN.}
Figure~\ref{fig:empirical-yarn} compares the YaRN-disabled run with an archived run configured with YaRN (factor 4; original context 32,768).
Both use the same checkpoint and 64/4,096-token output limits for Non-thinking/Thinking.
The YaRN setting is documented in the registered configuration; scores are recovered from archived request-level records.
In the YaRN-configured run, Thinking has higher mean accuracy than Non-thinking at every tested length.

\subsection{Which length functions describe Thinking?}
\label{app:empirical-length-tests}

\paragraph{Literature and candidate functions.}
Length-sensitive retrieval and reasoning are documented by RULER and FLenQA~\citep{hsieh2024ruler,levy2024sametask}; these evaluations motivate testing length dependence without specifying a universal error law.
% Previous notation retained:
% \citet{chen2026critical} derive a logarithmic scaling of attention logits in a simplified model, which concerns a different quantity from $\varphi(L)$.
\citet{chen2026critical} derive a logarithmic scaling of attention logits in a simplified model, which concerns a different quantity from $\gamma(L)$.
% Previous notation retained:
% We therefore compare five explicit hypotheses for the effective per-step success $q(L)=1-\varphi(L)$, keeping $p_{\mathrm{T}}=(1-\alpha)q(L)^N$.
We therefore compare five explicit hypotheses for the effective per-step success $q(L)=1-\gamma(L)$, keeping $p_{\mathrm{T}}=(1-\alpha)q(L)^N$.
Let $x=(L/1000-1)/99$ and $z=\ln(L/1000)/\ln100$:
\begin{equation}
\begin{aligned}
% Previous notation retained:
% q_{\mathrm{lin}\text{-}\varphi}&=q_0(1-rx), &
q_{\mathrm{lin}\text{-}\gamma}&=q_0(1-rx), &
% Previous notation retained:
% q_{\mathrm{log}\text{-}\varphi}&=q_0(1-rz),\\
q_{\mathrm{log}\text{-}\gamma}&=q_0(1-rz),\\
q_{\mathrm{lin}\text{-}\lambda}&=q_0e^{-bx}, &
q_{\mathrm{log}\text{-}\lambda}&=q_0e^{-bz}, &
q_{\mathrm{recip}}&=\frac{q_0}{1+bx}.
\end{aligned}
\label{eq:empirical-length-candidates}
\end{equation}
Here $0<q_0\leq1$, $0\leq r\leq1$, and $b\geq0$; each candidate has three parameters including $\alpha$.
% Previous notation retained:
% The first pair makes $\varphi$ linear in $L$ or $\log L$; the second does so for $\lambda=-\ln q$.
The first pair makes $\gamma$ linear in $L$ or $\log L$; the second does so for $\lambda=-\ln q$.
For the reciprocal candidate, a target competing with $D$ equal-score distractors at fixed logit margin $m$ has attention mass $1/(1+De^{-m})$.
Assuming $D$ grows linearly with $L$ gives the reciprocal length dependence.
Using this shape as a success-probability proxy is an additional assumption of our model.

\begin{figure}[H]
\centering
\includegraphics[width=\linewidth]{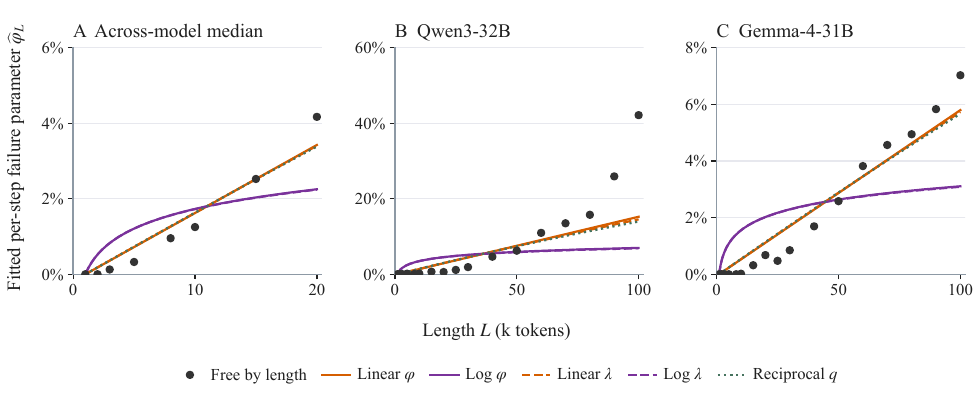}
\caption{\textbf{Candidate length dependence of effective step failure.} Dots are the free-by-length estimates from Eq.~\eqref{eq:empirical-step-product}; curves are the five joint fits in Eq.~\eqref{eq:empirical-length-candidates}. All fits use accuracy observations, not the dots. Panel A uses 1k--20k medians; B,C use each model's complete 1k--100k data. Vertical scales differ.}
\label{fig:empirical-length-candidates}
\end{figure}

\paragraph{Comparison and scope.}
All candidates use the same unsmoothed accuracies and least-squares loss.
The fixed 1k--100k rescaling and parameter constraints ensure valid probabilities throughout the tested domain, including frozen short-to-long prediction; they use no held-out outcomes.
We first leave out each whole length and refit, retaining every count.
We separately train on 1k--20k and freeze all parameters before testing 25k--100k.
Figure~\ref{fig:empirical-length-candidates} shows the implied length profiles, and Table~\ref{tab:empirical-length-validation} reports every candidate and comparison group.

% Previous notation retained:
% Linear $\varphi(L)$ has lower held-length RMSE than logarithmic $\varphi(L)$ in 11 of the 12 short-context groups; Nemotron Nano 3-4B is the exception.
Linear $\gamma(L)$ has lower held-length RMSE than logarithmic $\gamma(L)$ in 11 of the 12 short-context groups; Nemotron Nano 3-4B is the exception.
The median curves give 4.9 versus 9.0 points; Qwen and Gemma over 1k--100k give 15.5 versus 27.4 and 7.0 versus 13.8 points.
Linear-$\lambda$ and reciprocal-success fits have similar errors, so these results support a coarse length trend without identifying a unique functional law.
% Previous notation retained:
% Frozen short-to-long errors remain much larger: even linear $\varphi$ gives 44.0 points for Qwen and 18.4 for Gemma.
Frozen short-to-long errors remain much larger: even linear $\gamma$ gives 44.0 points for Qwen and 18.4 for Gemma.
% Previous notation retained:
% The fitted $\varphi$ describes answer failures under the specified output budget and does not directly measure retrieval errors.
The fitted $\gamma$ describes answer failures under the specified output budget and does not directly measure retrieval errors.

\begin{table}[H]
\centering
\caption{\textbf{Length-model validation: $R^2$/RMSE (percentage points).} The first 13 rows leave out each length in the 1k--20k grid; the next two do so over 1k--100k. The final two are frozen short-to-long predictions. All five candidates have three parameters and retain all 14 counts. Higher $R^2$ and lower RMSE indicate better predictions. $^\dagger$Separate checkpoints.}
\setlength{\tabcolsep}{3pt}
\begin{tabular}{lrrrrr}
\toprule
% Previous notation retained:
% Thinking data & Linear $\varphi$ & Log $\varphi$ & Linear $\lambda$ & Log $\lambda$ & Reciprocal $q$\\
Thinking data & Linear $\gamma$ & Log $\gamma$ & Linear $\lambda$ & Log $\lambda$ & Reciprocal $q$\\
\midrule
Across-model median & 0.86/4.9 & 0.53/9.0 & 0.86/5.0 & 0.53/9.0 & 0.85/5.0\\
Qwen3-4B & 0.74/5.4 & 0.43/7.9 & 0.73/5.4 & 0.43/7.9 & 0.73/5.4\\
Qwen3-8B & 0.71/3.1 & 0.41/4.5 & 0.71/3.2 & 0.41/4.5 & 0.71/3.2\\
Qwen3-14B & 0.36/2.2 & 0.24/2.4 & 0.36/2.2 & 0.24/2.4 & 0.36/2.2\\
Qwen3-32B & 0.61/2.3 & 0.37/2.9 & 0.61/2.3 & 0.37/2.9 & 0.61/2.3\\
Gemma-4-E4B & 0.85/10.1 & 0.52/18.2 & 0.84/10.3 & 0.51/18.3 & 0.84/10.6\\
Gemma-4-12B & 0.76/7.1 & 0.38/11.3 & 0.75/7.1 & 0.38/11.3 & 0.75/7.2\\
Gemma-4-26B-A4B & 0.77/6.1 & 0.52/8.7 & 0.77/6.1 & 0.52/8.7 & 0.76/6.1\\
Gemma-4-31B & 0.20/2.7 & -0.04/3.1 & 0.20/2.7 & -0.04/3.1 & 0.20/2.7\\
Nemotron Nano v2-9B & 0.87/11.2 & 0.83/12.8 & 0.87/11.1 & 0.83/13.0 & 0.87/11.1\\
Nemotron Nano 3-4B & 0.79/15.3 & 0.87/12.0 & 0.79/15.0 & 0.87/12.1 & 0.80/14.8\\
GLM-4/Z1-9B$^\dagger$ & 0.95/6.8 & 0.85/11.6 & 0.95/6.5 & 0.84/11.9 & 0.95/6.4\\
Ministral-3-8B pair$^\dagger$ & 0.82/8.7 & 0.72/10.7 & 0.82/8.7 & 0.72/10.8 & 0.82/8.6\\
\midrule\multicolumn{6}{l}{1k--100k grid: leave one length out}\\
Qwen3-32B & 0.81/15.5 & 0.39/27.4 & 0.80/15.9 & 0.39/27.6 & 0.78/16.3\\
Gemma-4-31B & 0.87/7.0 & 0.51/13.8 & 0.87/7.1 & 0.51/13.8 & 0.87/7.2\\
\midrule\multicolumn{6}{l}{Frozen 1k--20k fit evaluated at 25k--100k}\\
Qwen3-32B & -0.54/44.0 & -1.49/56.0 & -0.55/44.1 & -1.49/56.0 & -0.56/44.3\\
Gemma-4-31B & 0.33/18.4 & -0.58/28.4 & 0.33/18.5 & -0.58/28.4 & 0.32/18.6\\
\bottomrule
\end{tabular}
\label{tab:empirical-length-validation}
\label{tab:empirical-length-extrapolation}
\end{table}
\FloatBarrier

% !TEX root = ../main.tex
\section{Further details for the Non-thinking mode experiment}
\label{app:non-thinking}

\subsection{Experimental design}
\label{app:nonthinking-prompts}
\label{app:nonthinking-protocol}

Experiments use Qwen3-8B and Gemma-4-E4B on approximately 10k-token passages. We modify the common Non-thinking prompt in Appendix~\ref{app:behavioral-prompts} by inserting ``Write the count using ordinary decimal digits, with no space after the colon.'' before ``Your entire response must be exactly one line:'', and replacing \texttt{Total: <integer>} with \texttt{Total:<integer>}. After chat-template rendering with Thinking disabled, we supply the assistant prefix \texttt{Total:}; its final token is the \emph{answer query}.

A needle span is a complete city--score record; its endpoint is its final token. Needle end and answer-query states are labeled by running index $k$ and final count $N$, respectively. Layer/head indices are one-based: Qwen L1--36 and Gemma L1--42.

Table~\ref{tab:nonthinking-cohorts} lists the cohorts; by default, seeds 1234--1253 are used for selection and fitting, and seeds 1254--1263 are held out for evaluation. Interventions and controls use matched prompts, with sites and directions fixed before held-out evaluation where a split is used. Clean means unmodified, regardless of correctness. We write $\widehat N$ for the generated count, $N_s,N_t$ for source/target counts, and $E$ for the expected count conditional on candidate answers 1--10. Effects average counts/pairs within seed, then seeds equally. Unless specified otherwise, intervals use 10,000 seed-bootstrap draws and are pointwise 95\% intervals. Blue/orange denote Qwen/Gemma; PCA colors denote index/count.
For each candidate count $n\in\{1,\ldots,10\}$, let $p_n$ be the joint probability of its decimal answer tokens and the chat-template termination suffix, conditional on the supplied prompt and assistant prefix, under the evaluated intervention. We compute
\[
E=\frac{\sum_{n=1}^{10}n\,p_n}{\sum_{n=1}^{10}p_n}.
\]
Thus, $E$ is a probability-weighted average over these ten candidates, not the generated count $\widehat N$; probabilities assigned to other outputs are excluded by renormalization.

% Qwen routing/propagation tests reuse the OV evaluation cohort.
% Joint $p$ in the model-specific tests below is the maximum one-sided exact seed sign-flip $p$ across required effect and control comparisons.

\begin{table}[H]
\centering
\caption{Data and splits for Non-thinking experiments, per model unless specified. Eligibility and invalid-output rules accompany each analysis.}
\label{tab:nonthinking-cohorts}
\small
\begin{tabularx}{\linewidth}{@{}>{\raggedright\arraybackslash}p{.25\linewidth}X@{}}
\toprule
Analysis & Data and split \\
\midrule
Head scores & 20 selection seeds, $N=2$--10: 180 prompts. \\
Needle end geometry & $N=1$--10: 1,100 fitting endpoints from 200 prompts and 550 held-out endpoints from 100 prompts. \\
Count classification & 20 selection seeds, five-fold seed-grouped cross-validation. Both readouts use $N=1$--10: 1,100 running-index endpoints and 200 answer-query states. \\
Prompt-span steering & Ten initially correct $N=3$ prompts; all layers. \\
Cue/domain controls & Cue: ten paired seeds 1234--1243 at $N=10$. Domain: fit on 20 city fitting seeds; test on ten held-out seeds, $N=1$--10, in each domain. \\
Head ablation & 20 seeds 1316--1335, $N=1$--5: 100 prompts. \\
Span patching / serial tests & Ten held-out seeds, $N=1$--10: 100 prompts. Answer-subspace removal uses $N=2$--10: 90 prompts. \\
Answer state patching & All-layer source-count matching: 24 Qwen/26 Gemma correct-input directed pairs from seeds 1254--1258. Further single-layer/cumulative tests are described below. \\
% Qwen OV and routing tests & OV directions: seeds 1234--1253; centers and routing/readout axes: 1264--1273; evaluation: 20 seeds 1274--1293, $N=1$--10. \\
% Gemma residual mediation & Selection: seeds 1456--1465; held-out: 20 seeds 1466--1485, $N=1$--10. Residual axes fit odd counts and are evaluated on held-out even counts. \\
\bottomrule
\end{tabularx}
\end{table}

\subsection{Broad retrieval and count representations}
\label{app:nonthinking-representations}

\begin{figure}[H]
\centering
\includegraphics[width=\linewidth]{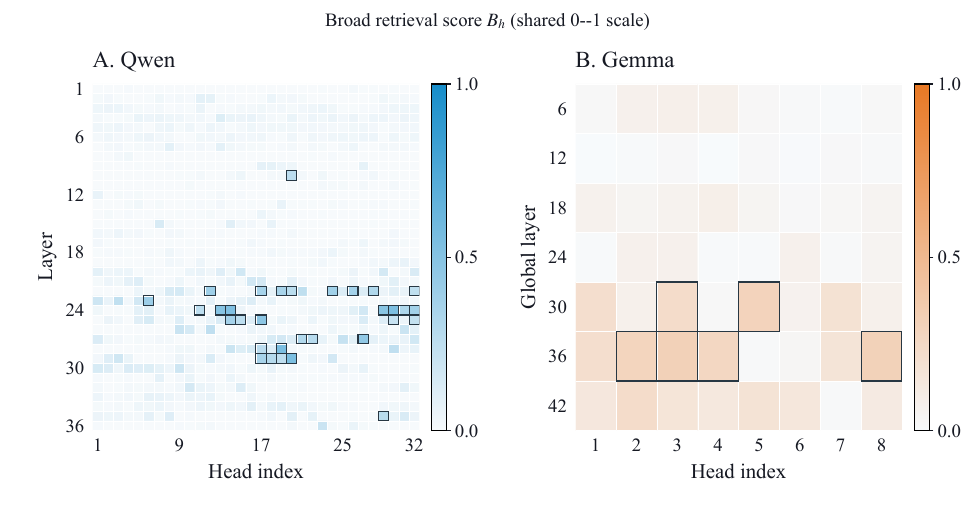}
\caption{\textbf{Complete broad-retrieval scores.} \textbf{A.} All 1,152 Qwen heads. \textbf{B.} All 56 Gemma global-attention heads. Cells show mean $B_h$ on 180 selection prompts. Horizontal axes show heads; vertical axes show layers from shallow to deep. Both panels use a linear 0--1 scale. Outlines mark frozen Qwen Top-32/Gemma Top-6 membership.}
\label{fig:nonthinking-full-head-scores}
\end{figure}

\paragraph{Head selection and complete scores.}
\label{app:nonthinking-head-scores}
Mean full-span $B_h$ on the selection prompts (Sec.~\ref{sec:broad-retrieval-score}) ranks the frozen Qwen Top-32/Gemma Top-6 banks. Figure~\ref{fig:nonthinking-full-head-scores} includes every measured head; Gemma is restricted to global-attention layers. Representative attention profiles are in Fig.~\ref{fig:nonthinking-broad-retrieval}.

\paragraph{Running index geometry and final count readout.}
\label{app:nonthinking-geometry}
Both running-index and final-count analyses use $N=1$--10, with 200 fitting and 100 held-out prompts from the default 20/10 seed split, without answer-correctness filtering. A prompt with count $N$ contributes its $N$ needle endpoints, yielding 1,100 fitting and 550 held-out running states per model; each prompt contributes one answer-query state. We compare nearest-centroid classification (NCC) and L2 logistic regression ($C=1$) across layers using five-fold cross-validation grouped by seed. For running index, each fold fits hidden-coordinate standardization followed by whitened PCA16, with no NCC shrinkage and class-balanced logistic weights, matching the primary Thinking protocol. The final-count scan retains standardization followed by unwhitened PCA16 and NCC shrinkage 0.1. Layers maximize pooled out-of-fold NCC balanced accuracy, with ties broken by logistic accuracy and then earlier depth. Balanced accuracy averages recall over labels 1--10; preprocessing uses only training-fold states.

\begin{figure}[H]
\centering
\includegraphics[width=\linewidth]{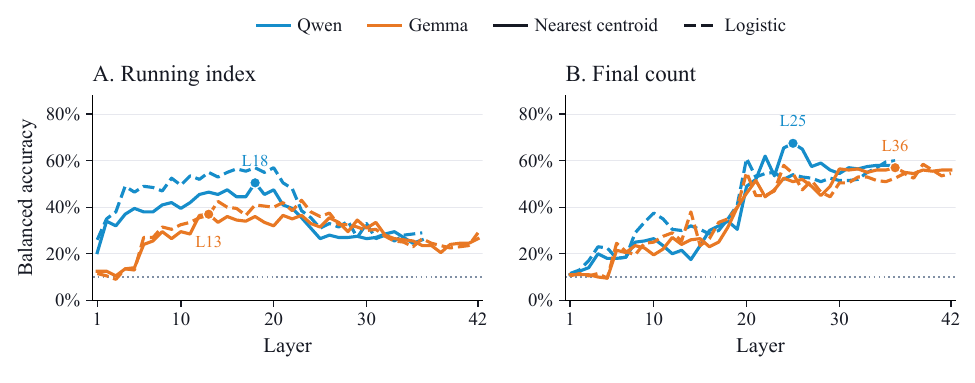}
\caption{\textbf{Count readout across layers.} \textbf{A.} Running index at needle endpoints. \textbf{B.} Final count at the answer query. Both cover $N=1$--10 and use five-fold seed-grouped cross-validation on 20 selection seeds: 1,100 states in A and 200 in B. Accuracy averages recall over the ten labels. Solid/dashed curves denote nearest-centroid/logistic classifiers; dotted lines mark 10\% chance. Labeled markers identify the NCC-selected layers, with logistic accuracy and then earlier layers breaking ties.}
\label{fig:nonthinking-count-readouts}
\label{fig:nonthinking-final-count-readout}
\end{figure}

For running index, this selects Qwen L18/Gemma L13 (cross-validated NCC 50.5\%/37\%). Refitting on all fitting states yields held-out NCC balanced accuracy of 46\%/40\% and ridge $R^2=0.910/0.827$ with penalty one in the same coordinates. A separate PCA of the ten class centroids computed from the fitting data captures 98.1\%/97.5\% of centroid variance in three components: ordered class means coexist with overlapping individual states. Index also covaries with position. Readout comparisons share the held-out set and do not constitute independent replications. Figure~\ref{fig:nonthinking-count-readouts} compares the two readouts; final count NCC peaks at 67.5\% for Qwen L25 and 57\% for Gemma L36 on cross-validation folds.

\paragraph{Linear steering at the last needle.}
\label{app:nonthinking-steering}
Neither model answers the ten original $N=10$ held-out prompts correctly, so the all-layer scan uses ten initially correct $N=3$ prompts/model; Qwen adds seeds 1264--1265. Every last-needle token receives $\beta\sigma v$, where $v$ is an auxiliary PCA16--ridge direction fitted to the 200 endpoints of the 20 selection prompts at $N=10$, using unwhitened PCA16 followed by coordinate standardization, then mapped to residual space and normalized, $\beta=\pm1$, and $\sigma$ is its span-projection standard deviation on the fitting data. Controls use norm-matched orthogonal-random directions and zero strength. Gemma's answers remain unchanged. Under $+1$, Qwen changes its answer seven times at L19 and L21: six from 3 to 2 and one from 3 to 4. Under $-1$, one answer changes from 3 to 2 at L18. Random controls also induce underestimation; zero controls preserve outputs. The intervention provides no reliable directional control.

\begin{figure}[H]
\centering
\includegraphics[width=\linewidth]{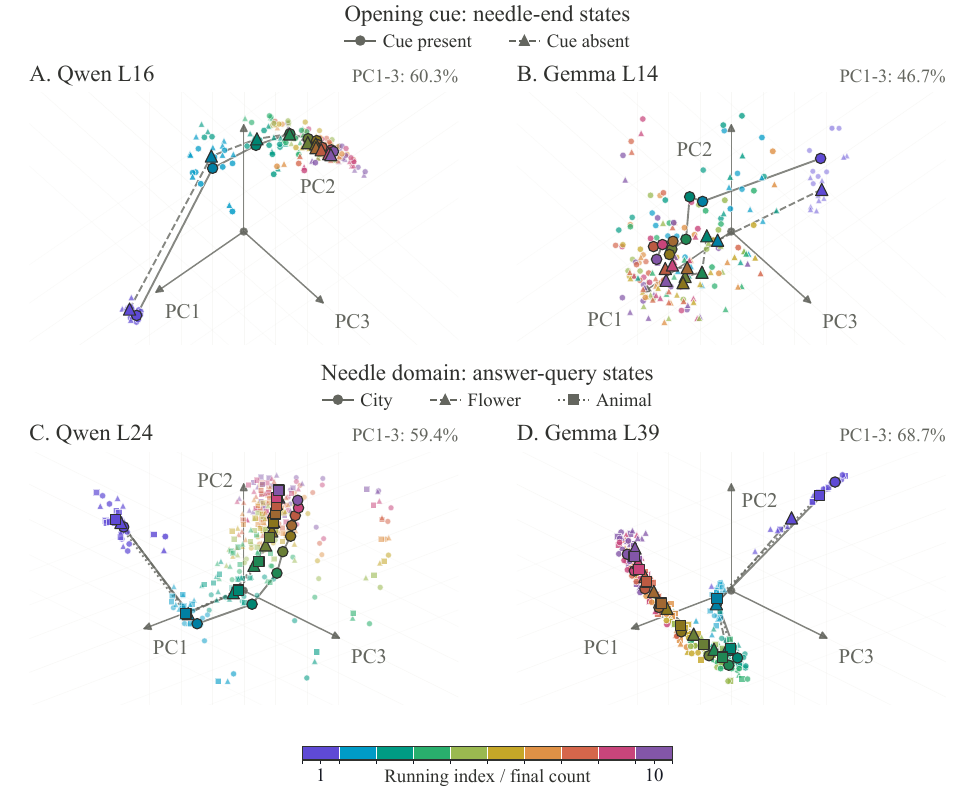}
\caption{\textbf{Cue and domain controls.} A,B: 100 needle end states per condition at $N=10$ in the first three components of a jointly fitted unwhitened PCA6 (Qwen L16/Gemma L14). C,D: 100 answer-query states per domain (city, flower, animal) in a PCA3 estimated from city examples. Small points are states; connected large markers are class means. Circles/triangles denote cue present/absent in A,B; circles/triangles/squares denote city/flower/animal in C,D. Percentages give fitting-population variance explained. Conditions share the projection and display scale within each panel.} 
\label{fig:nonthinking-cue-domain}
\end{figure}

\paragraph{Opening-cue and needle-domain controls.}
\label{app:nonthinking-robustness}
Removing the two opening task-definition sentences preserves the passage, final question, and answer format. In this auxiliary fixed-$N=10$ control, at Qwen L16/Gemma L14 selected on the corresponding fixed-count fitting data, PCA fitted jointly to both conditions shows similar centroid configurations (Fig.~\ref{fig:nonthinking-cue-domain}A--B; linear CKA between running index centroids in the original hidden space is 0.9995/0.9996). This comparison concerns geometry under cue removal.

For domain transfer, city becomes flower or animal throughout the prompt and records. City-only cross-validated NCC selects Qwen L24/Gemma L39 using standardization and whitened PCA16 without NCC shrinkage, with the same tie rule; this differs from the unwhitened PCA16 layer scan. Frozen classifiers are evaluated on held-out prompts in each domain (Table~\ref{tab:nonthinking-domain-transfer}). Flower transfer lowers answer-query accuracy in both models. For visualization, separate standardization and unwhitened PCA3 estimated from city examples are applied unchanged to city, flower, and animal states (Fig.~\ref{fig:nonthinking-cue-domain}C--D).

\begin{table}[H]
\centering
\caption{\textbf{Frozen city-trained count readout.} Nearest-centroid accuracy on 100 held-out prompts per topic ($N=1$--10; ten seeds). Chance is 10\%.}
\label{tab:nonthinking-domain-transfer}
\small
\begin{tabular}{@{}lccc@{}}
\toprule
Model & City & Flower & Animal \\
\midrule
Qwen L24 & 67\% & 50\% & 56\% \\
Gemma L39 & 58\% & 43\% & 53\% \\
\bottomrule
\end{tabular}
\end{table}

\subsection{Causal tests of the form--retrieve--consolidate pathway}
\label{app:nonthinking-causal}

\paragraph{Form: needle span patching.}
Corruption replaces needles with equal-token-length ordinary text. At each layer, restoration copies clean post-block span states into the corrupted target; reverse patching copies corrupted states into the clean target. Normalized error reduction/increase is
\begin{equation}
e_{\mathrm{restore}}=\frac{|E_{\mathrm{corrupt}}-N|-|E_{\mathrm{restored}}-N|}{N},
\qquad
e_{\mathrm{corrupt}}=\frac{|E_{\mathrm{patched}}-N|-|E_{\mathrm{clean}}-N|}{N}.
\label{eq:nonthinking-span-effects}
\end{equation}
Figure~\ref{fig:nonthinking-form-retrieve-consolidate}B reports both restoration and corruption effects, which weaken at deeper layers. Endpoint-only and token-budget/structure-matched ordinary-region restoration stay within 1.5 percentage points of zero relative to their corresponding corrupted baselines.

\begin{figure}[H]
\centering
\includegraphics[width=\linewidth]{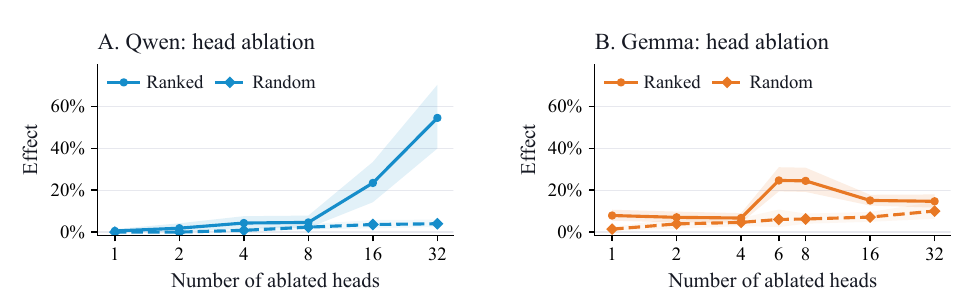}
\caption{\textbf{Retrieval-head ablation.} A: Qwen; B: Gemma. Ranked heads (solid circles) versus layer-matched random banks (dashed diamonds). Effect is the absolute change in generated count divided by the true count $N$. Bands show seed-bootstrap intervals.}
\label{fig:nonthinking-retrieval-interventions}
\end{figure}

\paragraph{Retrieve: head ablation.}
\label{app:nonthinking-retrieval}
We zero head contributions at the answer query and generate greedily, measuring $|\widehat N_{\mathrm{ablated}}-\widehat N_{\mathrm{clean}}|/N$. The $K\in\{1,2,4,8,16,32\}$ sweep and Gemma's $K=6$ condition use the same 100 prompts per model. At each dose, three random banks match per-layer head counts, sample without replacement within banks, and may overlap ranked heads. Replicates average within prompt. All numerically parseable outputs are retained, including one Gemma random-control output of 11; each random mean uses all three replicates.
Figure~\ref{fig:nonthinking-retrieval-interventions} shows the dose sweeps, including the selected Qwen Top-32/Gemma Top-6 banks. On 87 Qwen/63 Gemma initially correct prompts, equal-seed correct-to-wrong rates are 67.5\%/8.8\% for ranked/random Qwen $K=32$, and 33.3\%/12.5\% for Gemma $K=6$.

\paragraph{Consolidate: answer state patching and removal.}
We patch the full answer state at one layer at a time between prompts with different counts, retaining pairs whose source and target baseline answers are both correct. The all-layer analysis uses 24 Qwen and 26 Gemma directed pairs from five seeds (1254--1258), fixed across layers. Retained counts range from 0 to 6, with source--target offsets $\pm1,\pm3,\pm5$. We measure the proportion of predictions matching the source count:
\begin{equation}
P_\ell=\frac{1}{M}\sum_{p=1}^{M}\mathbf{1}\!\left[\widehat N_{\mathrm{patched},p}^{(\ell)}=N_{s,p}\right],
\label{eq:nonthinking-answer-match}
\end{equation}
where $M$ is the number of eligible pairs and $N_{s,p}$ is pair $p$'s source count. Invalid outputs count as failures. We pool pairs and obtain pointwise 95\% intervals from 10,000 seed-bootstrap draws, recomputing the pooled ratio in each draw. Final-layer source-count matching is 23/24 for Qwen (95.8\%, interval 88.5--100\%) and 26/26 for Gemma (100\%; all bootstrap draws equal 100\%).

\begin{figure}[H]
\centering
\includegraphics[width=\linewidth]{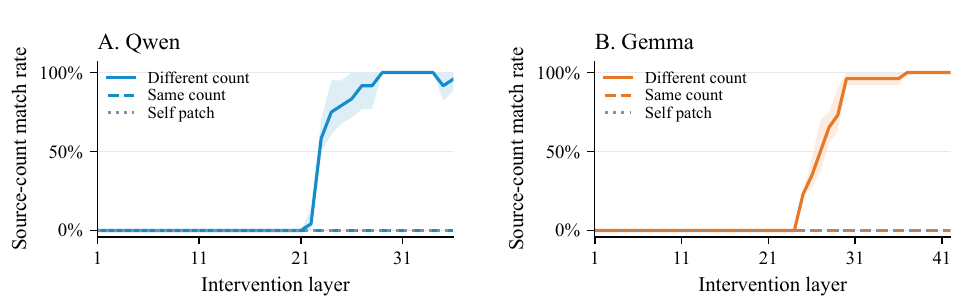}
\caption{\textbf{Answer state patching controls.} A: Qwen; B: Gemma. Proportion of predictions matching the source count after answer state patching. Different-count (solid), same-count (dashed), and self (dotted) patches use fixed correct-input pairs across layers. Controls are scored against the corresponding different-count source. Shading gives pointwise 95\% seed-bootstrap intervals.}
\label{fig:nonthinking-answer-controls}
\end{figure}

Self patches use the target's own state. Same-count controls use a state from another prompt with the target's count (20 Qwen/22 Gemma eligible pairs per layer). Both controls are scored against the corresponding different-count source count and give zero matches at every layer (Fig.~\ref{fig:nonthinking-answer-controls}).

\begin{figure}[H]
\centering
\includegraphics[width=\linewidth]{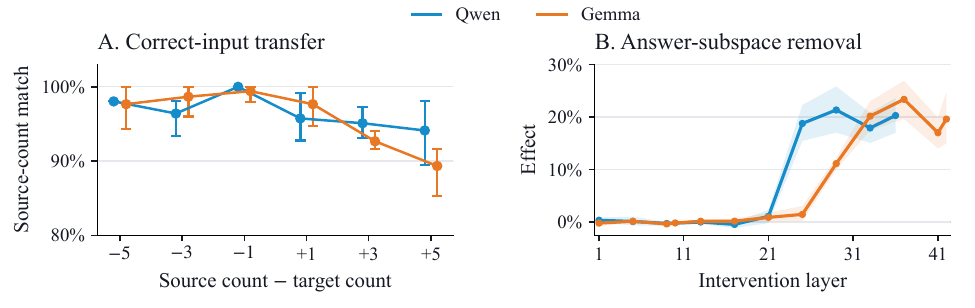}
\caption{\textbf{Answer state transfer and removal.} A: proportion of predictions matching the source count. B: normalized excess error from aligned removal. Seed-bootstrap intervals (50,000 draws in B).}
\label{fig:nonthinking-answer-function}
\end{figure}

Further tests patch either one layer or every layer from a selected starting layer through the final layer (cumulative-from-layer), using the source state from each corresponding layer. They use correct-input pairs with five seed clusters per count displacement and direction. Layer, patch-protocol, and count-displacement combinations are fixed using five separate selection seeds; the totals below pool the selected combinations. Source-count matches total 1628/1686 for Qwen and 1809/1884 for Gemma across selected single-layer/cumulative patches (Fig.~\ref{fig:nonthinking-answer-function}A). Invalid outputs fail; seed-bootstrap intervals recompute these pooled ratios.

For removal, the columns of $U$ are the first three principal directions of centered answer state class centroids from the fitting data ($N=1$--10). We apply $h'=h-UU^\top(h-\mu)$, with mean $\mu$ from the fitting data. Orthogonal within-count-residual controls match removal norm within 5\%. Panel B shows additional normalized error, $(|\widehat N_{\mathrm{aligned}}-N|-|\widehat N_{\mathrm{orthogonal}}-N|)/N$; unparseable outputs receive error ten before normalization. Late-layer effects survive Holm correction across layers within each model/population, including initially correct inputs.

\paragraph{Connecting the three stages.}
Serial tests restore needle states, then remove rank-three retrieval/answer components fitted using the selection data against norm-matched orthogonal controls. The needle/retrieval/answer layers are $9\to24\to30$ for Qwen and $10\to30\to38$ for Gemma. These sites are inherited from earlier source-restoration and retrieval/answer-subspace analyses and fixed before the serial tests; NCC display layers are selected separately. Banks are Qwen L24 H11,13,14,29,30,31,32 and Gemma L30 H3/H5; the Gemma bank is drawn from the Top-6 selection and fixed before evaluating serial interventions.

\begin{figure}[H]
\centering
\includegraphics[width=\linewidth]{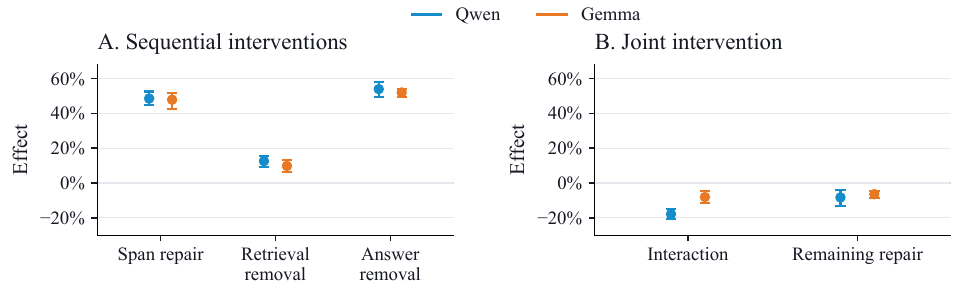}
\caption{\textbf{Serial interventions across the three stages.} A: restoration and retrieval/answer removal effects. B: interaction and remaining repair. All contrasts use absolute expected-count error divided by $N$ before averaging; both panels share the same percentage scale. Bars show seed-bootstrap intervals.}
\label{fig:nonthinking-answer-mediation}
\end{figure}

Let $e(X)=|E_X-N|/N$; $S/O$ restore needle/ordinary states in the same corrupted target, and $a/o$ denote aligned/orthogonal retrieval ($R$) or answer ($A$) removal. Figure~\ref{fig:nonthinking-answer-mediation}A reports restoration $e(O)-e(S)$ and removal effects $e(SR_a)-e(SR_o)$, $e(SA_a)-e(SA_o)$. Retrieval removal increases normalized error by 12.6/9.8 percentage points for Qwen/Gemma; answer removal increases it by 53.9/52.0, supporting both components' causal contribution.

Panel B gives interaction $[e(SR_aA_a)-e(SR_oA_a)]-[e(SR_aA_o)-e(SR_oA_o)]$ and remaining repair $e(O)-e(SR_aA_a)$. Negative interaction means that retrieval removal has a smaller additional effect when the answer component is also removed. Remaining repair is slightly negative: joint removal leaves more error than ordinary-region restoration. These effects share the true-count denominator $N$; they are not fractions of the restoration effect.

\section{Further details for the Thinking mode experiment}
\label{app:cot-reasoning}

\subsection{Experimental design}
\label{app:cot-protocol}
\label{app:native-prompts}

\begin{table}[H]
\centering
\caption{Data and splits for Thinking experiments, per model unless specified. Repeated conditions share seed clusters.}
\label{tab:cot-cohorts}
\small
\begin{tabularx}{\linewidth}{@{}>{\raggedright\arraybackslash}p{.26\linewidth}X@{}}
\toprule
Analysis & Population and split \\
\midrule
Head scores & Qwen: 15/20 eligible selection seeds, 1,152 heads. Gemma: 19/20 seeds, 56 global-attention heads; model-specific formats and queries. \\
Full-cohort geometry & 200 fitting / 100 held-out traces, $N=1$--10. Running states: Qwen 1,107/544; Gemma 1,045/519. Answer states: 200/100 each. \\
Domain transfer & 100 held-out traces per topic; projections and probes fit on the city fitting set. \\
Head ablation & Ten common final transitions with explicit ranks in both models, one per held-out seed; shared across head-count tests. Qwen Top-128/Gemma Top-6. \\
No-index readout / state transfer & 20 fitting / ten held-out traces per model, $N=10$; natural Qwen and prompted Gemma. Readout uses 200/100 states; held-out patches use $k=4,6,8$, both directions, and three scopes. \\
Answer state patching & 40 common directed pairs from 39 prompts; four per held-out seed, with both endpoints correct in both models. \\
\midrule
Qwen-only main-figure geometry & 30 paired $N=10$ inputs in a selected numbered format; secondary 20/10 split, with 200/100 states per mode (Fig.~\ref{fig:main}C). \\
\bottomrule
\end{tabularx}
\end{table}

Qwen3-8B and Gemma-4-E4B use approximately 10k-token passages and the Thinking template (Appendix~\ref{app:behavioral-prompts}), with Thinking enabled and no assistant prefill except in Gemma's prompted no-index control. Needles are prompt records; trace items are generated mentions. Query $q_k$ precedes the city mention of item $k$ (Marker $k-1$ in Fig.~\ref{fig:native-targeted-retrieval}); $H_k$ collects residual states over item $k$, ending at $t_k$; $h_{t_k}$ denotes its endpoint state. The answer query precedes the numeral. Layers/heads are one-based (Qwen 36/Gemma 42 layers).

\label{app:cot-validation}
At $N=1$--10, seeds 1234--1253 are used for selection and fitting, and seeds 1254--1263 are held out for evaluation. Table~\ref{tab:cot-cohorts} distinguishes the input, format-selected, and no-index cohorts. Intervention effects average within seed, then equally across seeds; default pointwise 95\% intervals use 10,000 seed-bootstrap draws. Repeated layers, doses, and patches share seed clusters. Intervals condition on the selected banks, sites, and analysis choices. Unparsed answers fail; truncated generations remain.
% Corrected Gemma replays fix inputs, model revision, interventions, and generation budgets;
% the attention helper snapshots parent/nested text configurations before either changes.

\paragraph{Parsing and eligibility.}
\label{app:cot-parsing}
The shared parser selects the longest $1,\ldots,M$ episode with consecutive explicit indices (earliest in ties). Rank-one restarts delimit episodes. Duplicate evidence for the same record and rank is merged; repetitions at advancing ranks remain. A structural list or recap is used if no episode exists, or if it contains the whole episode as an exact record prefix and adds a new record. If both routes fail, each record's first mention followed by its registered score within 96 characters determines occurrence order (3/300 Qwen; 0/300 Gemma). This fallback does not establish model-generated indices.

Prompt-record identities identify mentions and audit coverage. Neither gold count $N$ nor the final answer value selects or pads sequences. Thus $k$ indexes retained items, including repeated records. Answers are parsed independently. Full-cohort geometry uses parsed traces without correctness filtering; other eligibility, including complete-trace and correct-baseline requirements, is specified per analysis.

\subsection{Targeted retrieval and compact count representations}
\label{app:cot-representations}

\begin{figure}[H]
\centering
\includegraphics[width=\linewidth]{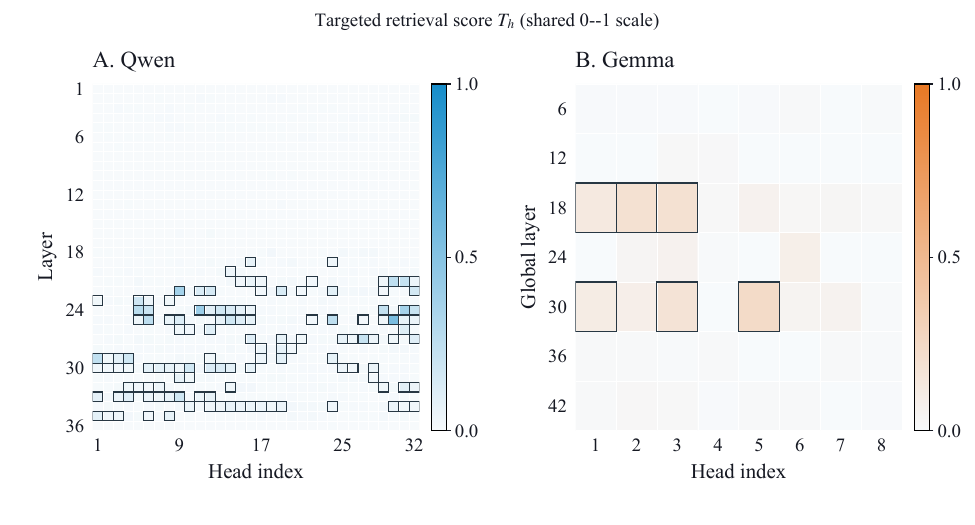}
\caption{\textbf{Complete targeted-retrieval scores.} \textbf{A.} All 1,152 Qwen heads (15/20 eligible selection seeds). \textbf{B.} The 56 Gemma global-attention heads (19/20). Scores use the model-specific formats and queries above. Horizontal axes show heads; vertical axes show layers from shallow to deep. Both panels use a linear 0--1 scale, shared with Enumeration. Outlines mark frozen Top-128/Top-6 membership.}
\label{fig:cot-full-head-scores}
\end{figure}

\paragraph{Head selection and complete scores.}
Of 20 selection seeds per model, 15 Qwen and 19 Gemma seeds contribute eligible transitions. Qwen requires a rank immediately before the city and queries after that marker; Gemma requires the rank and city in one text unit and queries at the preceding item end (Appendix~\ref{app:cot-attention}). Raw $T_h$ (Sec.~\ref{sec:targeted-retrieval-score}) is averaged within contributing seed, then equally across seeds. Figure~\ref{fig:cot-full-head-scores} shows all Qwen heads and Gemma's global-attention heads. Formats, query sites, tokenization, and attention coverage differ, limiting cross-model comparisons of magnitude. The bank-size limits are fixed experimental choices inherited from earlier exploratory analyses; the full head-count sweeps are reported below. Outlines retain the frozen Top-128/Top-6 banks; Gemma's displayed score ordering can differ from its frozen bank order.

\begin{figure}[H]
\centering
\includegraphics[width=\linewidth]{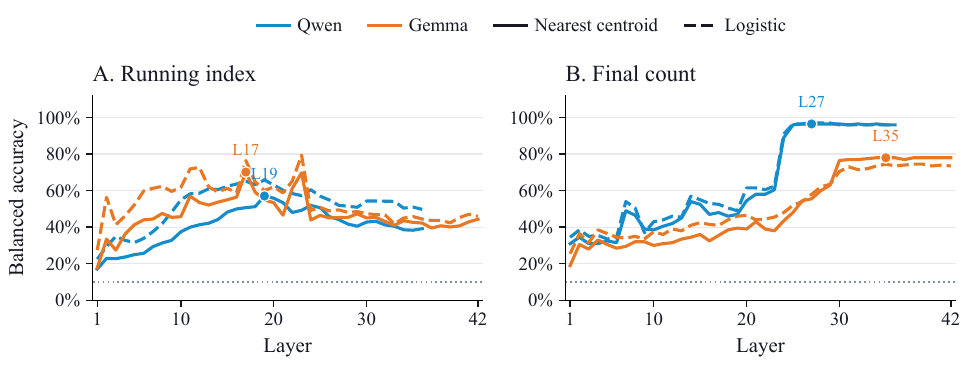}
\caption{\textbf{Count readout across layers.} \textbf{A.} Running index at trace-item endpoints. \textbf{B.} Final count at the answer query. Both analyses cover $N=1$--10. Curves pool predictions from five seed-grouped cross-validation folds (20 seeds/model): 1,107 Qwen/1,045 Gemma running states and 200 answer states each. Solid/dashed curves denote NCC/logistic classifiers; labeled markers identify NCC-selected layers. Blue denotes Qwen and orange Gemma; dotted lines mark 10\% chance.}
\label{fig:cot-count-readouts}
\end{figure}

\paragraph{Running index geometry and final count readout.}
The primary running-index and final-count analyses in both modes cover $N=1$--10; Thinking uses successfully parsed trace items with observed occurrence labels $k=1,\ldots,10$. Both models retain the shared 200/100 input split without answer-correctness filtering; generated traces determine the available running states, so the state populations are not paired. Item endpoints are labeled by observed occurrence $k$, including repeated records, and answer-query states by gold count $N$. We compare nearest-centroid classification (NCC, no shrinkage) and class-weighted L2 logistic regression ($C=1$). Each of five seed-grouped cross-validation folds fits standardization and whitened PCA16. Layers maximize pooled out-of-fold NCC balanced accuracy, with ties broken by logistic accuracy and then earlier depth. Balanced accuracy averages recall over ten labels. Readouts are refitted using the selection data before held-out evaluation. Figure~\ref{fig:cot-count-readouts} shows only cross-validation curves; held-out results at the selected layers are reported below.

Held-out NCC/logistic balanced accuracy is 58.9/68.1\% at Qwen L19 and 71.8/78.0\% at Gemma L17 for running index, and 99/100\% at Qwen L27 and 70/71\% at Gemma L35 for final count. Running index NCC accuracy is lower than in the grammar-filtered comparison in Fig.~\ref{fig:main}C, possibly due to more varied counts and trace grammars. Layer selection precedes held-out evaluation. Readout comparisons share the held-out set and do not constitute independent replications. Visible indices, token position, trace format, and $N$ can covary with $k$, and their contributions are not separated by these readouts.

\begin{figure}[H]
\centering
\includegraphics[width=\linewidth]{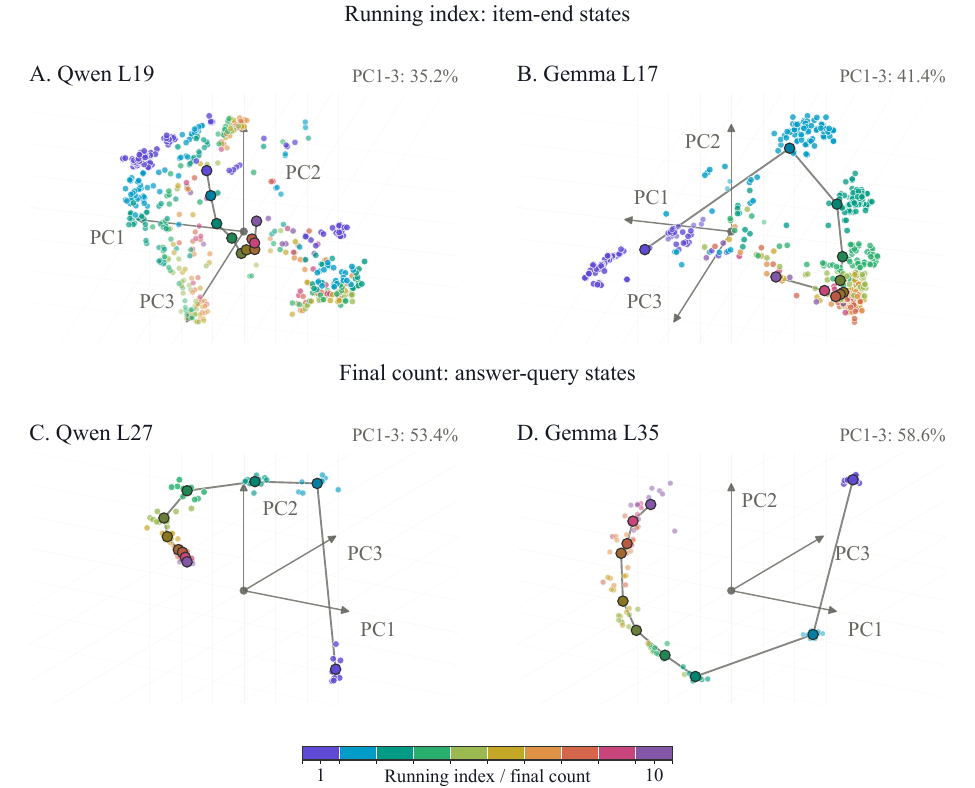}
\caption{\textbf{Counter-state and answer state PCA in the full Thinking cohort.} \textbf{A,B.} Qwen/Gemma running index states. \textbf{C,D.} Final count states. Small points show held-out states, with opacity indicating viewing depth; connected large markers show class means. Color denotes $k$ or $N$. Axes use PC1--3 fitted using the selection data; percentages report their combined explained variance.}
\label{fig:cot-count-pca}
\end{figure}

Separate standardization and unwhitened PCA3, fitted using the selection data using randomized SVD with seed 0, project all held-out states (Fig.~\ref{fig:cot-count-pca}). The three PCs explain 35.2/41.4\% of running-state variance and 53.4/58.6\% of final-state variance in Qwen/Gemma. Class means summarize heterogeneous, overlapping states. Independent PCA bases limit cross-panel distance comparisons, and the PCA3 display differs from the whitened PCA16 readout space. These analyses establish label separability; causal tests below address different, specified populations.

\paragraph{Qwen-only main-figure comparison.}
Figure~\ref{fig:main}C uses 30 paired $N=10$ inputs selected for correct, complete Thinking traces in consecutive \texttt{$k$. city - score} format, ending on score digits. After inspecting trace formats, a fixed seed hash assigns 20 inputs to fitting (eleven from the original fitting pool and nine from the original held-out pool) and ten to evaluation. These ten inputs are held out from refitting, but do not constitute a new independent cohort. The seed-grouped NCC protocol above selects Non-thinking L13/Thinking L31, yielding 46/98\% accuracy on 100 held-out states per mode. Separate PCA3 bases fitted using the selection data provide the display. This comparison changes mode, endpoint, and selected layer within a sample conditioned on Thinking format and correctness; numbering and position remain possible contributors.

\begin{figure}[H]
\centering
\includegraphics[width=\linewidth]{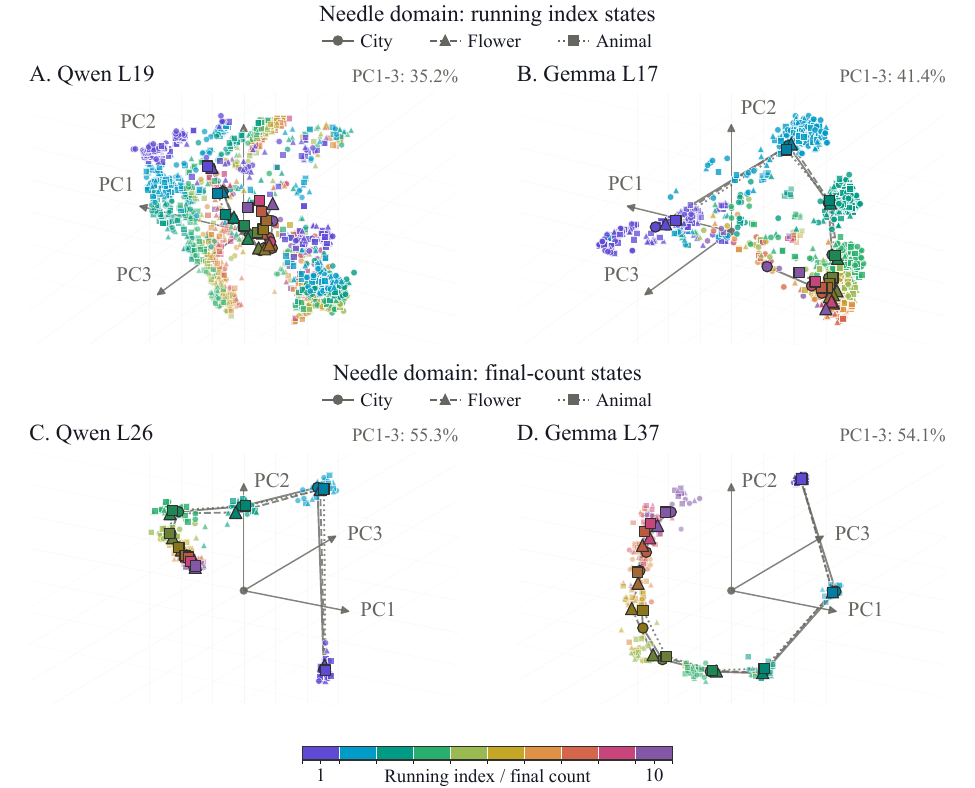}
\caption{\textbf{Count geometry across needle domains.} \textbf{A,B.} Qwen/Gemma running states. \textbf{C,D.} Final count states. Circles/triangles/squares denote city/flower/animal; small points show held-out states and connected large markers show class means. Colors encode $k$ or $N$. All three domains share the projection fitted on city data within each panel; percentages give fitting-population variance explained by PC1--3.}
\label{fig:cot-domain-pca}
\end{figure}

\paragraph{Needle-domain controls.}
City-trained projections and classifiers are frozen before testing 100 held-out traces per topic (city, flower, animal), using seeds 1254--1263 and $N=1$--10. Running states retain L19/L17. Final states use L26/L37 from a separate city-only five-fold NCC selection with experiment-specific folds and the same tie rule; their scores should be compared within this domain experiment. Standardization, PCA3 displays, and whitened PCA16 probes use only states from the city fitting set. Figure~\ref{fig:cot-domain-pca} compares city/flower/animal geometry; Table~\ref{tab:cot-domain-transfer} reports the corresponding readouts. Topic changes also alter tokenization and generated traces; running-state populations need not align across topics.

\begin{table}[H]
\centering
\caption{\textbf{Count readout across needle domains.} Cells report state count; NCC/logistic balanced accuracy (\%). Each domain contains 100 held-out trajectories per model. City-only fitting and layer selection precede transfer; chance is 10\%.}
\label{tab:cot-domain-transfer}
\small
\begin{tabular}{@{}llccc@{}}
\toprule
Model / endpoint & Layer & City & Flower & Animal \\
\midrule
Qwen / running $k$ & L19 & 544; 58.9/68.1 & 546; 58.6/61.7 & 528; 55.9/70.1 \\
Gemma / running $k$ & L17 & 519; 71.8/78.0 & 507; 70.7/78.8 & 524; 72.4/72.5 \\
Qwen / final $N$ & L26 & 100; 100/100 & 100; 98.0/98.0 & 100; 100/100 \\
Gemma / final $N$ & L37 & 100; 70.0/71.0 & 100; 70.0/69.0 & 100; 77.0/75.0 \\
\bottomrule
\end{tabular}
\end{table}

\subsection{Causal tests of the retrieve--update--read pathway}
\label{app:cot-causal}

\paragraph{Retrieve: head ablation and head-count dependence.}
\label{app:cot-retrieval}
\label{app:cot-current-dose}
We zero head slices before the output projection at the retrieval query and throughout cached greedy decoding (512-token limit). For each seed 1254--1263, we take the highest $N$ with a common final transition $N-1\to N$ and explicit ranks in both models, yielding counts 10, 10, 8, 4, 10, 7, 10, 7, 6, 7. This selects a particular trace population. Blanking and answer patching include these prompts. For next-item or final count failure $F$, ablation effect is $\Delta_c=10^{-1}\sum_s(\overline F_{s,c}-F_{s,\mathrm{clean}})$, averaging random repeats within seed. Effects subtract each model's own clean baseline: next-item failure is 0/20\% for Qwen/Gemma and final count failure is zero.

\begin{figure}[H]
\centering
\includegraphics[width=\linewidth]{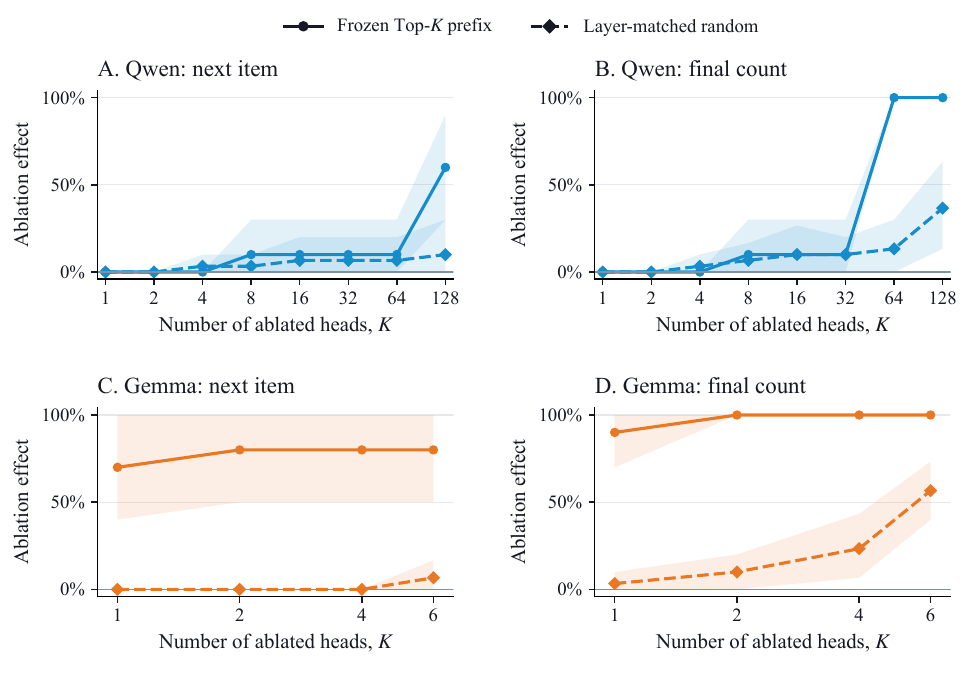}
\caption{\textbf{Head-count dependence in the frozen retrieval banks.} \textbf{A,B.} Qwen next-item/final count effects. \textbf{C,D.} Gemma effects (ten transitions/model). Solid circles ablate frozen selected prefixes; dashed diamonds average three layer-matched random banks excluding the selected bank. Axes show ablated-head count and failure increases from clean continuation; shading gives pointwise 95\% seed-bootstrap intervals. Masks persist through free generation.}
\label{fig:cot-current-dose}
\end{figure}

Figure~\ref{fig:cot-current-dose} tests frozen prefixes $K=1,2,4,8,16,32,64,128$ for Qwen and $K=1,2,4,6$ for Gemma. Three random banks per seed/dose exclude the full selected bank and match per-layer head counts; perturbation magnitudes remain unmatched. Doses share one clean continuation. Persistent masking can affect both retrieval and later generation, so final count effects do not isolate a single retrieval step.

Qwen next-item accuracy remains at least 9/10 through $K=64$, then falls to 4/10 at $K=128$ (random: 90\%). Final count accuracy already falls to 0/10 at $K=64$ (random: 86.7\%). Gemma Top-1, L30H5, leaves both accuracies at 1/10; $K=2,4,6$ leaves both at zero. Bank sizes do not quantify relative circuit sparsity across models. In a qualitative inspection, ablating small subsets of Qwen heads increased targeted attention in unablated heads in later layers. This pattern is compatible with compensatory retrieval, resembling the backup-head behavior and self-repair reported in prior work~\citep{wang2022interpretability,mcgrath2023hydra}. Whether this increase helps preserve next-item accuracy remains untested.

\paragraph{Update: counter-state transfer.}
\label{app:cot-progress}
We select patch layers using cross-validated NCC on the no-index cohorts described below, separately from the full-cohort geometry display. Gemma candidates are restricted to L1--L22: a post-block intervention can then affect subsequent K/V writes in both sliding and global attention. The last independent writers are L23 and L24, respectively; later layers reuse their K/V~\citep{gemmateam2026gemma4}. Qwen candidates L1--L35 retain a subsequent K/V write. Within each candidate range, we maximize five-fold seed-grouped cross-validated NCC, breaking ties by logistic accuracy and then earlier depth. Each fold fits standardization followed by whitened PCA16 on its training states; NCC uses no shrinkage, and logistic regression uses balanced class weights and $C=1$. This selects Qwen L19 and Gemma L21, with cross-validated NCC of 63.5\% and 52\%. At these fixed layers, NCC refitted on all fitting states achieves 70\%/48\% progress accuracy on Qwen/Gemma held-out traces before filler deletion (chance: 10\%).

Both models use $N=10$, source progress $k=4,6,8$, and target progress $k-1$ (forward) or $k+1$ (backward). A single post-block residual patch is followed by 96-token greedy generation. Source-successor adoption means that the first generated prompt-city mention before reasoning ends is $k+1$; later steps use the same ordered mention sequence. Non-record prompt filler is deleted to align absolute positions; the paired self patch uses the same aligned target prefix. Both models test the endpoint, a four-token tail, and the largest endpoint-aligned span fitting inside both items. Item spans contain 5--11 tokens in Qwen and 11--21 in Gemma. Decoding, alignment, sites, directions, and analysis rules are shared; each model retains its own no-index cohort.

Conditions share held-out seeds; intervals condition on the layer-selection procedure and the specified input cohorts. Truncated generations are retained and missing expected cities fail. Item-span runs reach the 96-token cap in 30/30 Qwen trials in each direction, and in 9/30 forward and 6/30 backward Gemma trials.

Qwen eligibility requires a first complete pass through all ten unique city--score records without repetition or explicit running index cues; later recaps are allowed. Held-out seeds are 1307, 1364, 1553, 1598, 1688, 1805, 1979, 1982, 2009, 2024. Gemma replaces the response instruction as below and supplies the prefix \texttt{FOUND:} followed by one space. Eligibility requires one record-supported \texttt{FOUND:} line per item, no repeats or other reasoning, and one terminal \texttt{Total:} line. The first 20 eligible traces are used for fitting and layer selection; the next ten, seeds 1276--1285, are held out for evaluation. Neither cohort is selected by final-answer correctness or intervention outcomes.

% Exact instruction from run_realistic_niah_v5_gemma_prompt_conditioned_noindex.py.
\begin{promptbox}[unbreakable]{Gemma control: replacement response instruction}
\setlength{\parskip}{1.5pt}
During reasoning, write each matching record on one line using exactly:\par
FOUND: <city> | score <score>\par
Always use the same literal marker FOUND:. Do not number records, use ordinal\par
words, item/record/excerpt indices, mention how many have been found, or write a\par
running subtotal. Do not write a preamble, recap, verification, or explanation.\par
Immediately after the final FOUND: line, output exactly one line:\par
Total: <integer>\par
Your assistant response has already begun with the literal text 'FOUND: ';\par
continue that first line directly with the first city and score.\par
\end{promptbox}

\begin{figure}[H]
\centering
\includegraphics[width=\linewidth]{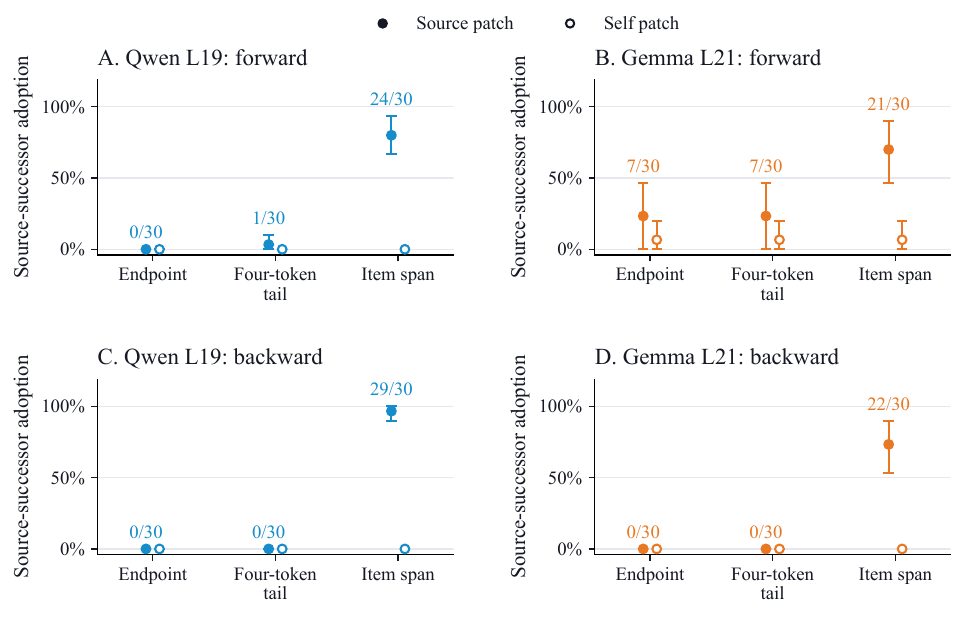}
\caption{\textbf{Counter-state scope controls by direction.} \textbf{A,B.} Forward transfer. \textbf{C,D.} Backward transfer. Columns show Qwen L19 and Gemma L21. Each scope uses the same 30 pairs per panel (ten seeds, $k=4,6,8$). Filled/hollow points show source/self-patch successor adoption; bars give pointwise 95\% seed-bootstrap intervals.}
\label{fig:cot-progress-controls}
\end{figure}

\textbf{Patch scope.} Item-span patches yield higher successor-adoption rates than endpoint or four-token patches in each model and direction (Fig.~\ref{fig:cot-progress-controls}). Layers and pairs are fixed across scopes, but perturbation norms are unmatched. Spans jointly transfer record content and progress; they do not isolate an arithmetic update, and progress can covary with order and position.

\textbf{Forward transfer.} Patching into target item $k-1$ skips ahead to an unseen record, with 24/30 Qwen and 21/30 Gemma successes (self: 0/30 and 2/30). The following item is correct in 24/24 and 20/21 cases without another patch (Fig.~\ref{fig:native-retrieve-encode-count-loop}C). Later columns require all earlier steps to succeed and another item to remain.

\textbf{Backward transfer.} Patching into target item $k+1$ first repeats its last completed record, with 29/30 Qwen and 22/30 Gemma successes (self: 0/30 each). The following item is correct in only 5/29 Qwen cases, versus 18/22 for Gemma. One untested explanation for Qwen's weaker continuation is that repeating an already-seen record favors rechecking or restarting. Trace-format differences limit the cross-model comparison.

\begin{figure}[H]
\centering
\includegraphics[width=\linewidth]{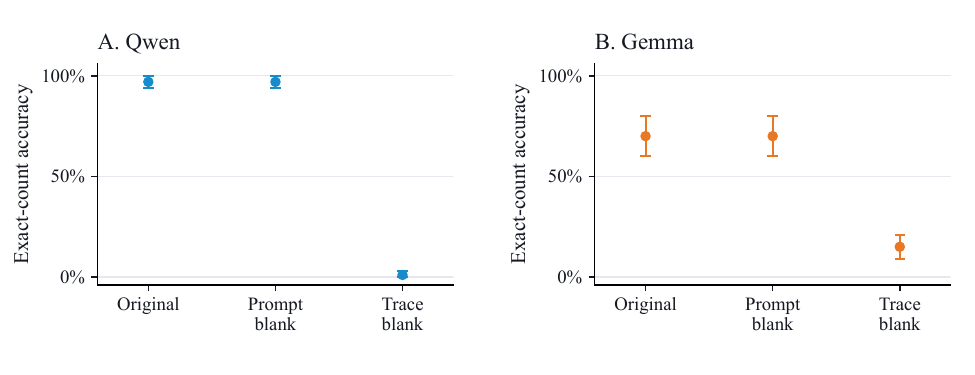}
\caption{\textbf{Trace dependence of final count readout.} \textbf{A.} Qwen. \textbf{B.} Gemma. Exact-count accuracy for Original, prompt-record blanking, and whole-trace blanking (100 unfiltered prompts/model). Error bars give pointwise 95\% seed-bootstrap intervals.}
\label{fig:cot-answer-readout-controls}
\end{figure}

\paragraph{Read: trace blanking and answer state patching.}
\label{app:cot-blanking}
\label{app:cot-answer-patching}
With the generated trace supplied, blanking zeros embeddings and post-block residuals at prompt records or across the trace, preserving positions and the answer query. On 100 unfiltered prompts/model (32-token greedy limit), Original and prompt-record-blanked exact-count accuracy is 97\% for Qwen and 70\% for Gemma; whole-trace blanking lowers it to 1\% and 15\% (Fig.~\ref{fig:cot-answer-readout-controls}A,B). This tests readout from an existing trace; whole-trace blanking jointly removes count and record content.

Answer state patches transfer the full single-token residual before the numeral across 40 common directed pairs, with both endpoints correct in both models. A 16-token greedy sweep covers all 36/42 layers; absolute positions need not match. We measure the proportion of predictions matching the source count (Eq.~\ref{eq:nonthinking-answer-match}), with invalid outputs counted as failures. Self patches preserve the target count at every layer. Final-layer source-count matching is 39/40 for Qwen (97.5\%, 95\% CI: 92.5--100\%) and 36/40 for Gemma (90\%, 82.5--97.5\%; Fig.~\ref{fig:native-retrieve-encode-count-loop}D). This controls the output numeral without separating count information from numeral encoding.

% !TEX root = ../main.tex
\section{Further details for the structured enumeration experiment}
\label{app:enumeration}

\subsection{Experimental design}
\label{app:enumeration-prompts}

We test the retrieve--update--read pathway under a common unnumbered Bullet format. Qwen3-8B and Gemma-4-E4B enumerate city--score records in approximately 10k-token passages with Thinking disabled, no assistant prefill, and a 4,096-token greedy limit. The common template in Appendix~\ref{app:behavioral-prompts} uses the following response instruction.

\begin{promptbox}[unbreakable]{Bullet enumeration: response instruction}
\setlength{\parskip}{1.5pt}
List each occurrence once, in passage order.\par
Begin each item with "-", then write the actual city name, then ": ", then the actual numeric score.\par
Use only actual values from the passage; do not output placeholders or angle brackets.\par
Then report the number listed:\par
Total: <integer>\par
Do not include any other text.\par
\end{promptbox}

Seeds 1234--1253 are used for selection and fitting, and seeds 1254--1263 are held out for evaluation, each crossed with $N=1$--10. Strict success requires the correct total, prescribed markers, every city--score pair exactly once in passage order, list--total agreement, and no truncation: Qwen succeeds on 297/300 inputs and Gemma on 229/300. Representation analyses retain all 300 inputs. Read, retrieval-head ablation, and answer state patching use the same default held-out seeds, with assay-specific eligibility (Table~\ref{tab:enumeration-cohorts}). Gemma's $N=10$ Update cohort also includes eligible inputs from disjoint, prespecified reserve pools described below. Representation and causal analyses share the default input pool and do not constitute independent replications.

Layers and heads are one-based. Effects average within source seed, then equally across seeds; continuation uses pooled conditional success ratios. Pointwise 95\% intervals resample source seeds 10,000 times and condition on frozen analysis choices. Repeated settings share seed clusters; degenerate intervals at observed 0/1 rates do not establish population certainty. Failures, unparsed answers, and truncations remain.

\begin{table}[H]
\centering
\caption{Data and splits for Enumeration experiments, per model unless specified. Repeated conditions share seed clusters.}
\label{tab:enumeration-cohorts}
\small
\begin{tabularx}{\linewidth}{@{}>{\raggedright\arraybackslash}p{.26\linewidth}X@{}}
\toprule
Analysis & Population and split \\
\midrule
Full-cohort geometry & 200 fitting / 100 held-out inputs; each model retains its available parsed running states and all answer states. \\
Head scores / ablation & 20 selection seeds; ten model-common final transitions, one per held-out seed. \\
Update & 20 $N=10$ fitting / ten held-out traces (200/100 states); model-specific eligible inputs. \\
Read blanking & 100 unfiltered inputs: ten seeds $\times$ $N=1$--10. \\
Answer state patching & 40 model-common directed pairs; four pairs per seed, full layer sweeps. \\
\bottomrule
\end{tabularx}
\end{table}

\subsection{Targeted retrieval and compact count representations}
\label{app:enumeration-representations}

\begin{figure}[H]
\centering
\includegraphics[width=\linewidth]{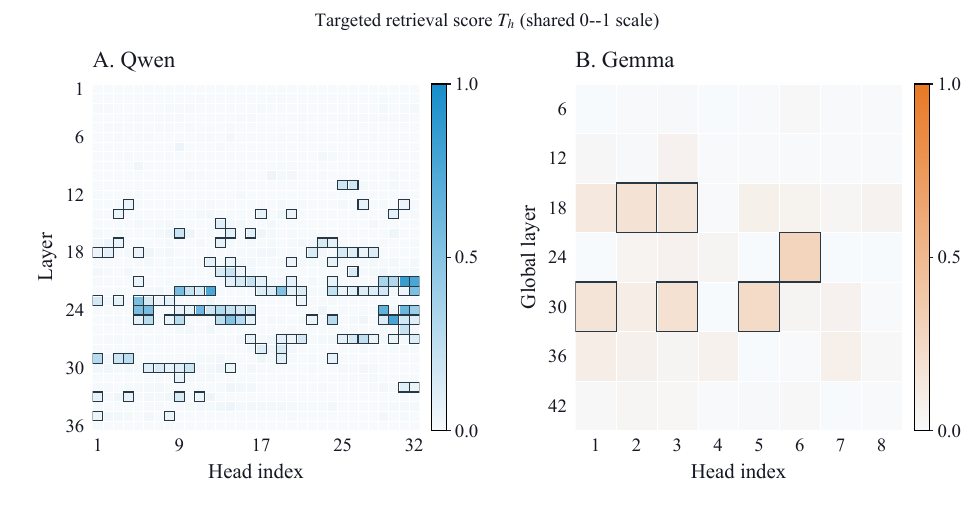}
\caption{\textbf{Complete targeted-retrieval scores.} \textbf{A.} All 1,152 Qwen heads. \textbf{B.} The 56 Gemma global-attention heads. Scores average within seed, then across 20 selection seeds/model. Horizontal axes show heads; vertical axes show layers from shallow to deep. Both panels use a linear 0--1 scale, shared with Thinking. Outlines mark frozen Top-128/Top-6 membership.}
\label{fig:enumeration-head-scores}
\end{figure}

\paragraph{Head selection and complete scores.}
Raw targeted retrieval scores $T_h$ (Sec.~\ref{sec:targeted-retrieval-score}) average attention mass on the next prompt record within each selection seed, then equally across seeds. Queries occur at the preceding item end for both selection and intervention. Qwen uses all 1,152 heads; Gemma uses 56 global-attention heads. The selection set supplies 180 Qwen and 164 Gemma queries across 20 seeds each. Figure~\ref{fig:enumeration-head-scores} shows the complete scores and frozen Top-128/Top-6 banks, using the same size limits as Thinking; model-specific coverage limits comparisons of magnitude.

\begin{figure}[H]
\centering
\includegraphics[width=\linewidth]{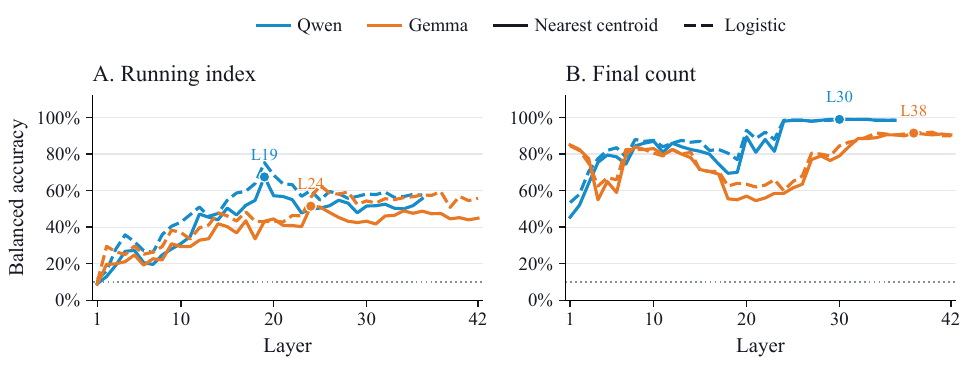}
\caption{\textbf{Count readout across layers.} \textbf{A.} Running index at trace-item endpoints. \textbf{B.} Final count at the answer query. Curves pool predictions from five seed-grouped cross-validation folds (20 seeds/model): 1,102 Qwen/1,067 Gemma running states and 200 answer states each. Solid/dashed curves denote NCC/logistic classifiers; labeled markers identify NCC-selected layers. Blue denotes Qwen and orange Gemma; dotted lines mark 10\% chance.}
\label{fig:enumeration-representations}
\end{figure}

\paragraph{Running index geometry and final count readout.}
We use the Thinking cross-validation preprocessing and NCC/logistic readouts (Appendix~\ref{app:cot-representations}), retaining each model's available running states with observed occurrence labels $k=1,\ldots,10$ and all 300 answer states without answer-correctness filtering. Display layers maximize pooled out-of-fold NCC balanced accuracy, with logistic accuracy and earlier depth breaking ties. Balanced accuracy averages recall over ten labels. The readouts are refitted on the fitting inputs and evaluated on held-out inputs. Figure~\ref{fig:enumeration-representations} shows only cross-validation curves; these mixed-$N$ populations are separate from the $N=10$ Update selection below.

Held-out NCC/logistic balanced accuracy is 69.4/75.3\% at Qwen L19 and 56.7/68.0\% at Gemma L24 for running index, and 100/100\% at Qwen L30 and 88/85\% at Gemma L38 for final count. Content, syntax, position, and progress can covary; their contributions are not separated by these readouts.

\begin{figure}[H]
\centering
\includegraphics[width=\linewidth]{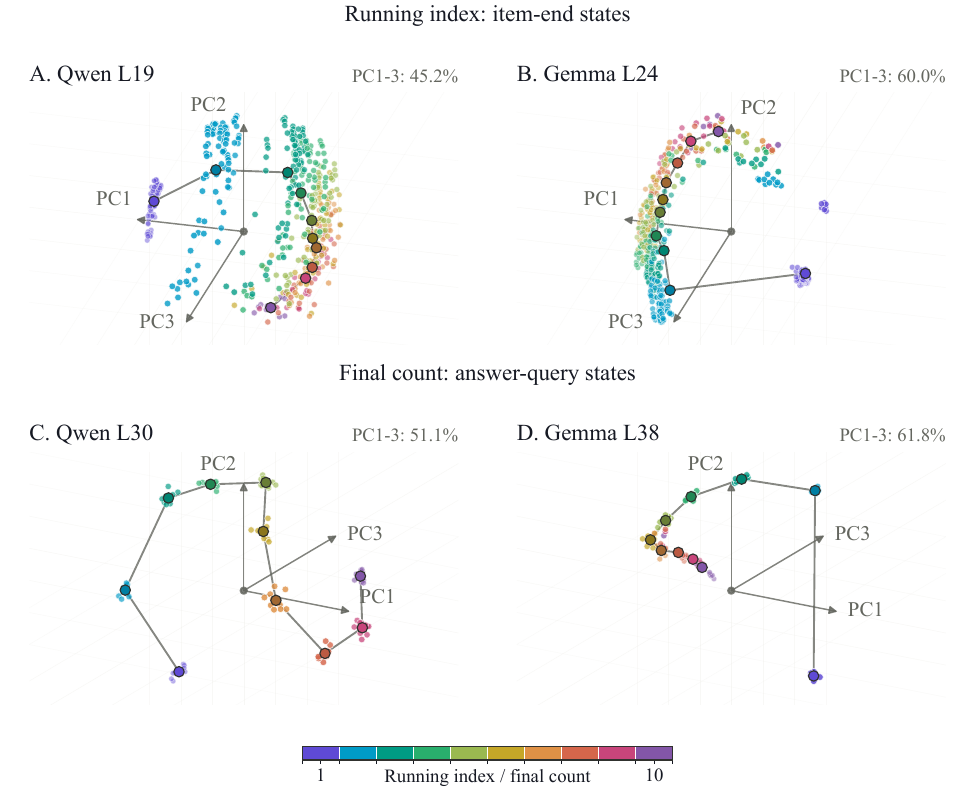}
\caption{\textbf{Counter-state and answer state PCA in Bullet enumeration.} \textbf{A,B.} All 550 Qwen/528 Gemma held-out running index states. \textbf{C,D.} All 100 final count states/model, including incorrect original answers. Small points show held-out states, with opacity indicating viewing depth; connected large markers show class means. Color denotes $k$ or $N$. Axes use PC1--3 fitted using the selection data; percentages report their combined explained variance.}
\label{fig:enumeration-pca-bullet}
\end{figure}

Separate standardization and unwhitened PCA3, fitted using the selection data using randomized SVD with seed 0, project all held-out states (Fig.~\ref{fig:enumeration-pca-bullet}). Class means summarize heterogeneous, overlapping states. Independent bases limit cross-panel distance comparisons, and the PCA3 display differs from the whitened PCA16 readout space. These analyses establish label separability; causal tests below address specified populations.

\subsection{Causal tests of the retrieve--update--read pathway}
\label{app:enumeration-causal}

\paragraph{Retrieve: head ablation and head-count dependence.}
For each held-out seed, we take the highest-count legal final transition shared by both models ($N=7$--10). Frozen doses are $K=1,2,4,8,16,32,64,128$ for Qwen and $1,2,4,6$ for Gemma. Three random banks match selected-head layer counts while excluding the full maximum selected bank. Only Qwen at $K=128$ requires global random: L22 needs 17 heads but has 15 available. Head outputs are zeroed before the output projection at the query and throughout 512-token greedy continuation; held-out intervention outcomes do not determine the dose or bank.

\begin{figure}[H]
\centering
\includegraphics[width=\linewidth]{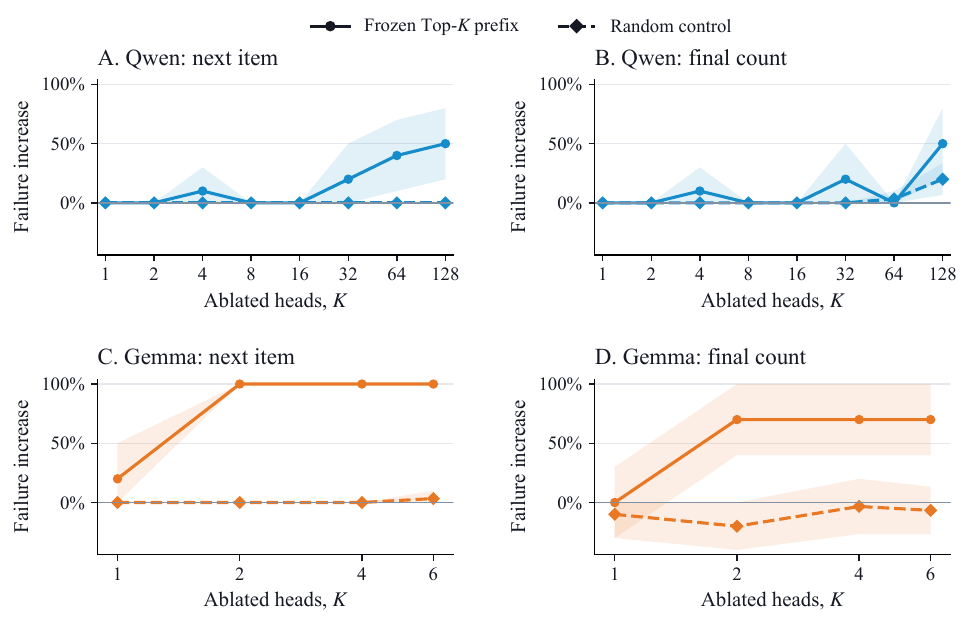}
\caption{\textbf{Head-count dependence in the frozen retrieval banks.} \textbf{A,B.} Qwen next-item/final count effects. \textbf{C,D.} Gemma effects (ten transitions/model). Solid circles ablate frozen selected prefixes; dashed diamonds average three random banks. Axes show ablated-head count and failure increases from clean continuation; shading gives pointwise 95\% seed-bootstrap intervals. Masks persist through free generation. All doses and eight truncated formal generations remain (four/model); random banks match layers except at Qwen Top-128.}
\label{fig:enumeration-retrieve}
\end{figure}

At the largest doses, selected-minus-clean next-item/final count failure increases are 50/50 percentage points for Qwen and 100/70 for Gemma (Fig.~\ref{fig:enumeration-retrieve}). Final count contrasts against random are 30.0 points (95\% CI: 0.0--60.0) and 76.7 (56.7--93.3), respectively. The Qwen final count contrast reaches zero at the lower interval endpoint, limiting evidence for an effect beyond random-head ablation on this outcome. Selected-head ablation disrupts subsequent retrieval in both models; persistent masks also affect later generation. Head-count thresholds do not measure relative circuit sparsity.

\begin{figure}[H]
\centering
\includegraphics[width=\linewidth]{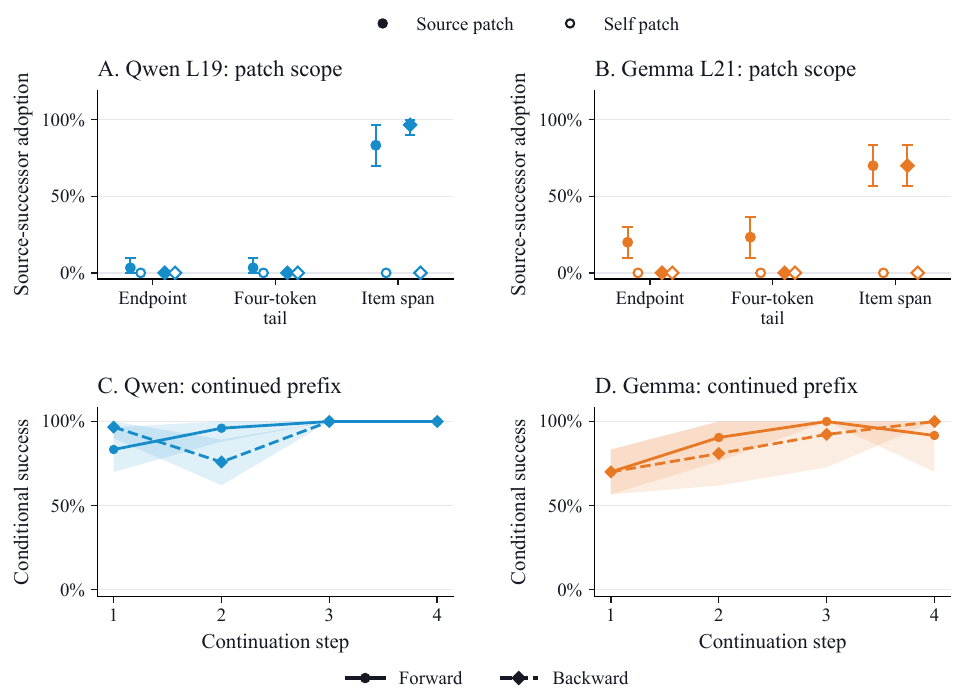}
\caption{\textbf{Counter-state scope controls and continuation.} Columns show Qwen L19 and Gemma L21. \textbf{A,B.} Filled/hollow points show source/self-patch successor adoption at each scope (30 trials/direction; ten seeds, $k=4,6,8$). \textbf{C,D.} Item-span next-step success conditional on the entire preceding source-implied city prefix being correct and another item remaining. Steps 1--2 include $k=4,6,8$; steps 3--4 include $k=4,6$. Circles/solid lines denote forward transfer; diamonds/dashed lines denote backward transfer. Bars and shading give pointwise 95\% seed-bootstrap intervals. Failures and truncations remain in the step for which they are eligible.}
\label{fig:enumeration-update}
\end{figure}

\paragraph{Update: counter-state transfer.}
The Thinking $N=10$ procedure selects layers from 200 item-end states/model, captured before filler deletion. Twenty fitting traces and ten held-out traces are selected by trace-format and token-alignment checks, without final-correctness filtering. Eligibility requires the specified bullet format, all ten records in passage order, and valid token spans that can be aligned by deleting non-record prompt filler for every tested patch condition. These checks concern trace structure and token positions. All default Qwen inputs qualify. Gemma's cohort contains 12 default and eight reserve fitting traces, and three default and seven reserve held-out traces. Reserve traces are selected in ascending eligible-seed order from fixed fitting/held-out pools 1264--1323/1324--1383. The selected reserve fitting seeds are 1264, 1265, 1269, 1274, 1276, 1277, 1278, and 1279; the selected reserve held-out seeds are 1324, 1325, 1327, 1330, 1333, 1337, and 1343. The pools are disjoint, and intervention outcomes do not enter selection.

Cross-validated NCC selects Qwen L19 and Gemma L21 within L1--35 and L1--22, respectively, preserving subsequent K/V writes (Appendix~\ref{app:cot-progress}). Cross-validated/held-out NCC accuracy is 68.5/70\% for Qwen and 44.5/52\% for Gemma. Layers are selected before held-out evaluation. Intervals condition on the layer-selection procedure and the specified input cohorts.

A single post-block patch transfers source item $k$ into target item $k-1$ (forward) or $k+1$ (backward), followed by 96-token greedy generation. We test $k=4,6,8$ at the endpoint, four-token tail, and largest common endpoint-aligned item span (6--8 tokens), with self and unpatched source controls. Deleting non-record filler aligns absolute positions. First-step success means adopting the source-implied successor $k+1$; later steps require the exact ordered city prefix.

\textbf{Patch scope.} Figure~\ref{fig:enumeration-update} compares source and self adoption at each scope and direction. Layers and pairs are fixed across scopes, but perturbation norms are unmatched. For item spans, self adoption is 0/30 in each model and direction; unpatched source adoption is 30/30 except Gemma forward (29/30). Spans jointly transfer record content and progress, limiting claims about a pure count update.

\textbf{Forward transfer.} Item-span patches into target item $k-1$ produce the source-implied successor in 25/30 Qwen and 21/30 Gemma trials. The following item is correct in 24/25 and 19/21 cases without another patch.

\textbf{Backward transfer.} Patches into target item $k+1$ first repeat its last completed record in 29/30 Qwen and 21/30 Gemma trials. The following item is correct in 22/29 and 17/21 cases, respectively.

\begin{figure}[H]
\centering
\includegraphics[width=\linewidth]{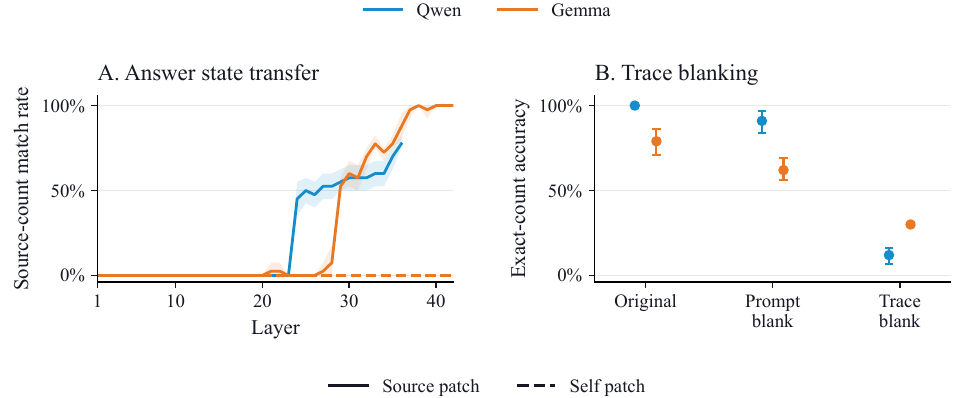}
\caption{\textbf{Answer state transfer and trace dependence.} \textbf{A.} Proportion of predictions matching the source count after full-state (solid) or self (dashed) patches at the single answer-query token, 40 directed pairs/model at every layer. \textbf{B.} Exact-count accuracy for Original, prompt-record blanking, and whole-trace blanking (100 unfiltered inputs/model). Blue denotes Qwen and orange Gemma. Shading and error bars give pointwise 95\% seed-bootstrap intervals.}
\label{fig:enumeration-readout}
\end{figure}

\paragraph{Read: trace blanking and answer state patching.}
On 100 unfiltered inputs/model, blanking zeros embeddings and post-block states while preserving positions and the answer query (32-token limit). Unmodified replay agrees with the original counts. Original/prompt-record-blank/trace-blank correct counts are 100/91/12 for Qwen and 79/62/30 for Gemma (Fig.~\ref{fig:enumeration-readout}B). Trace dependence coexists with prompt contributions.

Answer state transfer patches the full residual at one answer-query token across all 36/42 layers (16-token greedy limit). Thinking's low/high-count selector yields 40 model-common directed pairs with correct original answers, one-to-one traces, and legal queries; replay correctness is not an eligibility criterion. Self patches have zero state change and reproduce unmodified replay, preserving the original count in 40/40 pairs/model at every layer. Final-layer source-count matching is 31/40 for Qwen and 40/40 for Gemma (self: 0/40 for both; panel A). Full-state transfer does not isolate count from numeral encoding.

\section{Further details for the synthetic experiment}
\label{app:synthetic-details}

Our experiment adapts the small-decoder training setup of nanoGPT, whose character-level example trains on Tiny Shakespeare~\citep{karpathy2022nanogpt}. Related public implementations include minGPT's character-level language model~\citep{karpathyMingpt} and a nanoGPT reproduction on the same corpus~\citep{quickgridNanogpt}. We extend this setup with target-character queries and enumeration-trace supervision to study how retrieval and counting functions develop in a controlled setting.

%\YZ{Briefly say that our experiment is adapted from NanoGPT. Andrej Karpathy also studied Shakespeare.}
%\YZ{Search and cite earlier papers / repos that replicated the original experiment.} 
%\YZ{Briefly mention that Shakespeare is tiny and prone to overfitting.}

\subsection{Experimental design}
\label{app:synthetic-design}
%\YZ{Give a few more detailed examples of input sequences}
%\YZ{List a table of key hyperparameters}

\paragraph{Task and data construction.}
An input consists of a target set $S$ containing three distinct characters and a sequence $x=(x_1,\ldots,x_{256})$. Let $p_1<\cdots<p_N$ be the positions for which $x_{p_k}\in S$, and write $a_k=x_{p_k}$. The required answer is the total number of occurrences, $N=\sum_{j=1}^{256}\mathbf{1}\{x_j\in S\}$, with $N\in\{1,\ldots,10\}$. Matching is case-sensitive, and repeated appearances of the same character are counted separately.

We split Tiny Shakespeare into contiguous training, validation, and test regions with an 80/10/10 allocation after reserving two 255-character guard intervals. Every source window lies within one region. A fixed pool contains 100 target sets, selected using training-region character frequencies; each set has combined frequency at most $10/256$, with sampling stratified across 20 frequency bins. During training, a maximum-entropy distribution over feasible (target set, count) pairs gives uniform set and count marginals. Conditional on a pair, a valid source-window start is sampled uniformly. We then randomly permute the 256 characters and shuffle the order of characters in the query. The permutation preserves the count and defines the source order used by the trace.

\paragraph{Sequence format.}
Corpus characters and structural tags are individual tokens; each final count also has its own atomic token. Both modes receive the same query and input format. Non-thinking directly predicts the answer, whereas Thinking enumerates the matching characters in source order without explicit indices:
\begin{align*}
\text{prompt:}\quad &\textcolor{promptblue}{\texttt{<BOS> <CountChar>}\ S_1\ S_2\ S_3\ \texttt{<Sep>}}\ \textcolor{black}{x},\\
\text{Non-thinking:}\quad &\textcolor{ansgreen}{\texttt{<Ans>}\ \texttt{<}N\texttt{>}\ \texttt{<EOS>}},\\
\text{Thinking:}\quad &\textcolor{thinkvermillion}{\texttt{<Think>}\ \texttt{<Sep>}\ \boldsymbol{a}_1\ \cdots\ \texttt{<Sep>}\ \boldsymbol{a}_N}\\[-2pt]
&\hspace{1em}\textcolor{thinkvermillion}{\texttt{</Think>}}\ \textcolor{ansgreen}{\texttt{<Ans>}\ \texttt{<}N\texttt{>}\ \texttt{<EOS>}}.
\end{align*}
Here $S_1,S_2,S_3$ denote the displayed query order. As in the main text, \textcolor{promptblue}{blue marks the query prefix}, \textcolor{thinkvermillion}{vermilion the trace}, and \textcolor{ansgreen}{green the answer}; target occurrences are bold. Table~\ref{tab:synth-examples} illustrates absent query characters, repeated occurrences, and adjacent occurrences (A--C).

\begin{table}[H]
\centering
\caption{\textbf{Examples of synthetic inputs.} Three actual test-region inputs and their expected outputs in both modes. The input characters are randomly permuted, so these excerpts do not preserve words or sentences. Within each example, both modes receive the same query prefix and 256-character input. Excerpts retain every target occurrence and nearby non-target characters; ellipses omit only non-target characters. Positions are one-based. Boldface marks target occurrences, \texttt{\textvisiblespace} denotes an input space, and \texttt{\char92{}n} denotes an input newline.}
\label{tab:synth-examples}
\small
\renewcommand{\arraystretch}{1.12}
\begin{tabularx}{\linewidth}{@{}l >{\raggedright\arraybackslash}X@{}}
\toprule
\multicolumn{2}{@{}l@{}}{\textbf{A. One occurrence ($N=1$)}}\\[2pt]
Query & \texttt{\textcolor{promptblue}{<BOS> <CountChar> S c D <Sep>}}\\
Shuffled input excerpt & \textcolor{black}{\ldots\  \texttt{,r\textbf{c}\textvisiblespace{}G} \ \ldots}\\
Target positions & 232\\
Counts by character & \texttt{S}: 0, \texttt{c}: 1, \texttt{D}: 0\\
Thinking output & \texttt{\textcolor{thinkvermillion}{<Think> <Sep> \textbf{c} </Think>}} \texttt{\textcolor{ansgreen}{<Ans> <1> <EOS>}}\\
Non-thinking output & \texttt{\textcolor{ansgreen}{<Ans> <1> <EOS>}}\\
\midrule
\multicolumn{2}{@{}l@{}}{\textbf{B. Repeated occurrences of one character ($N=4$)}}\\[2pt]
Query & \texttt{\textcolor{promptblue}{<BOS> <CountChar> D c S <Sep>}}\\
Shuffled input excerpt & \textcolor{black}{\ldots\  \texttt{ae\textbf{S}otl\textbf{c}rh} \ldots\  \texttt{\textvisiblespace{}w\textbf{c}e\textvisiblespace{}} \ldots\  \texttt{st\textbf{c}he} \ \ldots}\\
Target positions & 75, 79, 164, 178\\
Counts by character & \texttt{D}: 0, \texttt{c}: 3, \texttt{S}: 1\\
Thinking output & \texttt{\textcolor{thinkvermillion}{<Think> <Sep> \textbf{S} <Sep> \textbf{c} <Sep> \textbf{c} <Sep> \textbf{c} </Think>}} \texttt{\textcolor{ansgreen}{<Ans> <4> <EOS>}}\\
Non-thinking output & \texttt{\textcolor{ansgreen}{<Ans> <4> <EOS>}}\\
\midrule
\multicolumn{2}{@{}l@{}}{\textbf{C. Adjacent target occurrences ($N=7$)}}\\[2pt]
Query & \texttt{\textcolor{promptblue}{<BOS> <CountChar> b w k <Sep>}}\\
Shuffled input excerpt & \textcolor{black}{\ldots\  \texttt{Uh\textbf{w}c\textvisiblespace{}} \ldots\  \texttt{\textvisiblespace{}p\textbf{w}de} \ldots\  \texttt{e\textvisiblespace{}\textbf{k}\char92{}ne} \ldots\  \texttt{aI\textbf{k}ca} \ldots\  \texttt{ae\textbf{w}da} \ldots\  \texttt{ht\textbf{w}\textbf{b}gv}}\\
Target positions & 12, 18, 174, 181, 207, 253, 254\\
Counts by character & \texttt{b}: 1, \texttt{w}: 4, \texttt{k}: 2\\
Thinking output & \texttt{\textcolor{thinkvermillion}{<Think> <Sep> \textbf{w} <Sep> \textbf{w} <Sep> \textbf{k} <Sep> \textbf{k} <Sep> \textbf{w} <Sep> \textbf{w} <Sep> \textbf{b} </Think>}} \texttt{\textcolor{ansgreen}{<Ans> <7> <EOS>}}\\
Non-thinking output & \texttt{\textcolor{ansgreen}{<Ans> <7> <EOS>}}\\
\bottomrule
\end{tabularx}
\end{table}

\paragraph{Architecture and optimization.}
We train one decoder-only Transformer from scratch for each output format, using the same architecture and training settings (Table~\ref{tab:synth-hyperparameters}). No representation-learning objective or scheduled sampling is used.

\begin{table}[H]
\centering
\caption{\textbf{Synthetic training and evaluation settings.} Both modes use the same settings unless the output format or loss component is specified. There is one training run per mode. Learning-rate values refer to one-based optimizer steps.}
\label{tab:synth-hyperparameters}
\small
\renewcommand{\arraystretch}{1.06}
\begin{tabularx}{\linewidth}{@{}l >{\raggedright\arraybackslash}X@{}}
\toprule
Setting & Value\\
\midrule
\multicolumn{2}{@{}l@{}}{\textbf{Model and data}}\\
Transformer depth / heads per layer & 4 / 8\\
Hidden / head dimensions & 512 / 64 (approximately 12.7M parameters)\\
Feed-forward network & Width 2,048; GELU with tanh approximation\\
Normalization / dropout & Pre-LayerNorm / 0 throughout the model\\
Position encoding / capacity & RoPE, base 10,000 / 384 token positions\\
Embedding sharing & Input and output embeddings tied; the ten atomic count-output vectors are untied\\
Input / query / count range & 256 shuffled characters / 3 target characters / 1--10\\
Target-set pool / sampling seed & 100 sets / 21,234\\
Training / validation / test allocation & 80/10/10, with 255-character guard intervals\\
\midrule
\multicolumn{2}{@{}l@{}}{\textbf{Optimization}}\\
Optimizer & AdamW; $\beta_1=0.9$, $\beta_2=0.999$, $\epsilon=10^{-8}$\\
Weight decay & 0.01, applied to all model parameters\\
Gradient clipping & Global $\ell_2$ norm capped at 1.0 before each optimizer step\\
Batch size & 128 examples per optimizer step; no gradient accumulation\\
Training budget & 10,000 optimizer steps; 1,280,000 sampled examples per mode, including repeats\\
Precision & bfloat16 autocast; model parameters remain in float32\\
Training seed & 1234 for each independently trained mode\\
\midrule
\multicolumn{2}{@{}l@{}}{\textbf{Learning-rate schedule}}\\
Peak learning rate & $\eta_{\max}=3\times10^{-4}$ at step 500\\
Linear warmup & Steps 1--500; $\eta_1=6\times10^{-7}$, rising linearly to $\eta_{\max}$\\
Cosine decay & Steps 501--10,000; minimum learning rate $\eta_{10{,}000}=0$\\
Loss-switch behavior & Schedule continues at step 1,501; no restart or second warmup\\
\midrule
\multicolumn{2}{@{}l@{}}{\textbf{Objectives and evaluation}}\\
Steps 1--1,500 & Teacher-forced next-token cross-entropy over the full sequence\\
Steps 1,501--10,000 & Task-output loss with count / marker / structure weights 8 / 8 / 16; Non-thinking has no marker term\\
Task-output loss reduction & Per-example mean within each component, then the weighted sum and batch mean (Eq.~\ref{eq:synth-output-loss})\\
Selection / evaluation inputs & 200 / 100; respectively 20 / 10 per count\\
Analysis checkpoints & Steps $0,100,200,\ldots,10{,}000$ (101 checkpoints)\\
\bottomrule
\end{tabularx}
\end{table}

\paragraph{Training objectives.}
For steps 1--1,500, we minimize next-token cross-entropy over all non-padding targets. From step 1,501 onward, the loss includes task-output targets only, beginning with \texttt{<Ans>} for Non-thinking and \texttt{<Think>} for Thinking. Let $\bar\ell_C$, $\bar\ell_R$, and $\bar\ell_S$ denote cross-entropy averaged first within each example over final count tokens, retrieved-character tokens, and structural tokens, respectively, and then across the batch. The objectives are
\begin{equation}
\mathcal L_{\mathrm{NT}}=8\bar\ell_C+16\bar\ell_S,
\qquad
\mathcal L_{\mathrm{T}}=8\bar\ell_C+8\bar\ell_R+16\bar\ell_S.
\label{eq:synth-output-loss}
\end{equation}
For Thinking, structural tokens include \texttt{<Think>}, every \texttt{<Sep>}, \texttt{</Think>}, \texttt{<Ans>}, and \texttt{<EOS>}; for Non-thinking they are \texttt{<Ans>} and \texttt{<EOS>}. Separately averaging the components keeps the final count coefficient independent of trace length. These weights are inherited from preliminary training experiments and held fixed in the reported runs.

\begin{figure}[H]
\centering
\includegraphics[width=\linewidth]{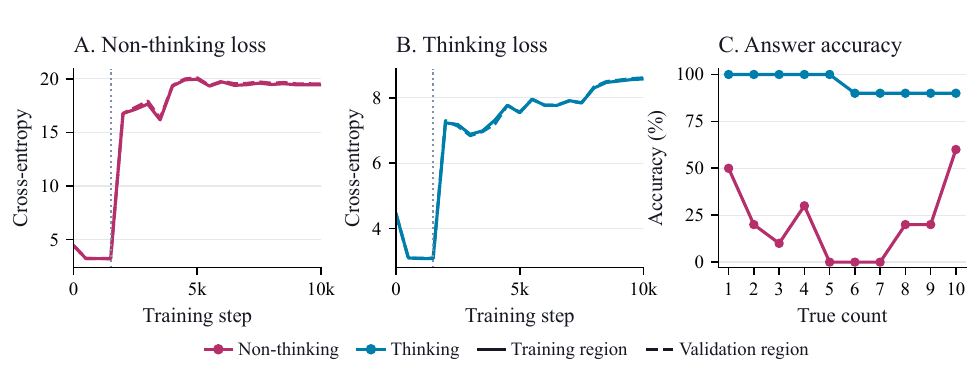}
\caption{\textbf{Training and counting behavior.} A/B: full-sequence, token-weighted cross-entropy on evaluation inputs generated from the training (solid) and validation (dashed) regions of the original corpus. Validation-region inputs are excluded from parameter updates. The dotted line marks the objective change at step 1,500; this evaluation metric remains fixed. C: final-checkpoint answer accuracy, with ten evaluation inputs per count and mode. Missing numeric answers are incorrect.}
\label{fig:apx-behavior}
\end{figure}

Tiny Shakespeare is a small corpus, approximately 1 MB, and the original nanoGPT configuration explicitly anticipates overfitting~\citep{karpathy2022nanogpt}. To monitor generalization, Fig.~\ref{fig:apx-behavior} reports full-sequence cross-entropy on inputs from the training and validation regions. We retain this evaluation loss after training switches to the weighted task-output loss in Eq.~\ref{eq:synth-output-loss} at step 1,501, so the plotted values remain comparable across training stages. The two curves largely overlap and do not identify a clear onset of overfitting. The objective switch at step 1,501 is a preset training choice.

\paragraph{Evaluation protocol.}
Unless stated otherwise, mechanistic analyses use disjoint selection and evaluation sets from the held-out test region: 200 and 100 inputs, organized into 20 and 10 analysis blocks, respectively, with one input per count $N=1,\ldots,10$. The sets share neither input identities nor (target-set, source-window-start) pairs. Head rankings and representation transforms are fitted on selection inputs. Answer accuracy uses greedy generation, counting missing or incorrect answers as failures. Attention and representation measurements use teacher-forced prefixes; final-answer prompt and trace ablations use each model's generated prefix.

\subsection{Broad and targeted retrieval heads}
\label{app:synth-retrieval}

%\YZ{Write the score definition, and add some explanations}
%\YZ{Perhaps write head ablation analysis here?}

\paragraph{Broad and targeted retrieval.}
Let $A_h(q,p)$ be the attention weight from query position $q$ to source position $p$ in head $h$, and let $q_A$ be the \texttt{<Ans>} position whose logits predict the final count. Let $s_k$ be the position of input character $x_{p_k}$ in the complete token sequence, including the query prefix; $p_k$ remains its index within $x$. For one input, define total target attention $M_h$, its normalized distribution $\pi_{hk}$ over the target occurrences, and the corresponding entropy $H_h$ by
\begin{equation}
M_h=\sum_{k=1}^{N}A_h(q_A,s_k),\qquad
\pi_{hk}=\frac{A_h(q_A,s_k)}{M_h},\qquad
H_h=-\sum_{k=1}^{N}\pi_{hk}\log\pi_{hk}.
\label{eq:synth-broad-components}
\end{equation}
The broad score is $B_h=M_h\exp(H_h)/N$. We set $B_h=0$ if $M_h=0$ and use $0\log0=0$. The factor $\exp(H_h)$ is the effective number of attended target occurrences: uniform target attention gives $B_h=M_h$, while concentrating the same mass on one occurrence gives $B_h=M_h/N$. Thus, the score rewards both mass and coverage.

\begin{figure}[H]
\centering
\includegraphics[width=\linewidth]{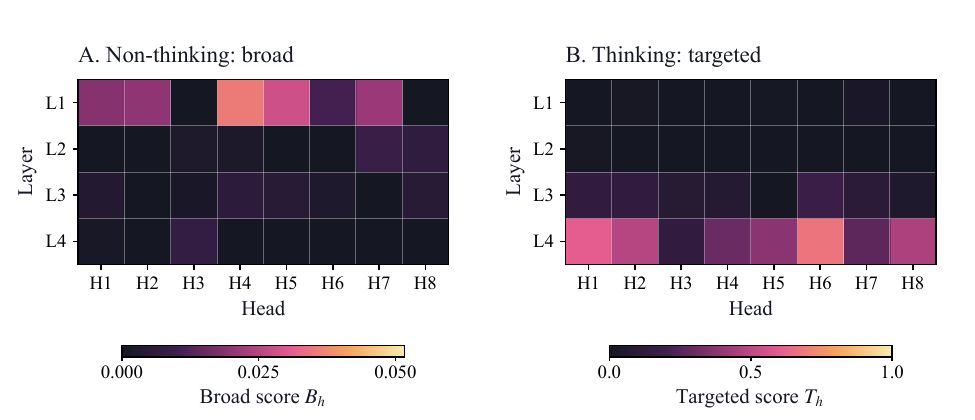}
\caption{\textbf{Retrieval-head scores.} A: Non-thinking broad score $B_h$ at the final-answer query. B: Thinking targeted score $T_h$ at trace retrieval queries. Each cell shows one head's score at step 10,000, averaged equally over 100 teacher-forced evaluation inputs. Rows are layers L1--L4; columns are heads H1--H8. The panels use separate color scales, matching Fig.~\ref{fig:apx-attention-dynamics} in Appendix~\ref{app:synth-dynamics}, with no normalization across heads.}
\label{fig:apx-head-scores}
\end{figure}

For Thinking, $q_k$ is the \texttt{<Sep>} query immediately before the $k$th trace character. The targeted score is
\begin{equation}
T_h=\frac{1}{N}\sum_{k=1}^{N}A_h(q_k,s_k).
\label{eq:synth-targeted-score}
\end{equation}
This score distinguishes the corresponding occurrence from other instances of the same character. Each score is computed within an input and then averaged equally across inputs. Heads are ranked globally across all four layers; $\mathrm{L}\ell\mathrm{H}j$ denotes layer $\ell$ and head $j$, both one-based. Figure~\ref{fig:apx-head-scores} shows the final-checkpoint scores.

\paragraph{Head bank ablation.}

We zero selected head outputs before the attention output projection, from the final-answer query (Non-thinking) or final retrieval query (Thinking) through EOS, capped at 64 new tokens. We measure final count accuracy for Top-1--4 broad heads on all 100 Non-thinking inputs, and next-needle accuracy for Top-1--8 targeted heads on 88 Thinking inputs. The latter have $N\geq2$ and an available clean final retrieval query; their selection ignores final count correctness. No inputs are excluded based on intervened outputs.

Next-needle accuracy measures whether the next generated token is the correct target character. For Thinking, we intervene at the $N$th \texttt{<Sep>} in the clean generated trace and score whether the next token equals $a_N$; non-character outputs are errors.

Controls match the number of heads per layer. We enumerate all feasible sets, using the least possible overlap with selected heads; Top-4 Non-thinking controls share one head. Non-thinking controls also preserve L1H3, following a diagnostic in which its ablation made all 100 inputs repeat \texttt{<Ans>}. Replacing its values with the fixed value vector for the non-target character \texttt{o} restored every clean first-token prediction, suggesting an approximately constant contribution to answer-token generation.

\begin{figure}[H]
\centering
\includegraphics[width=\linewidth]{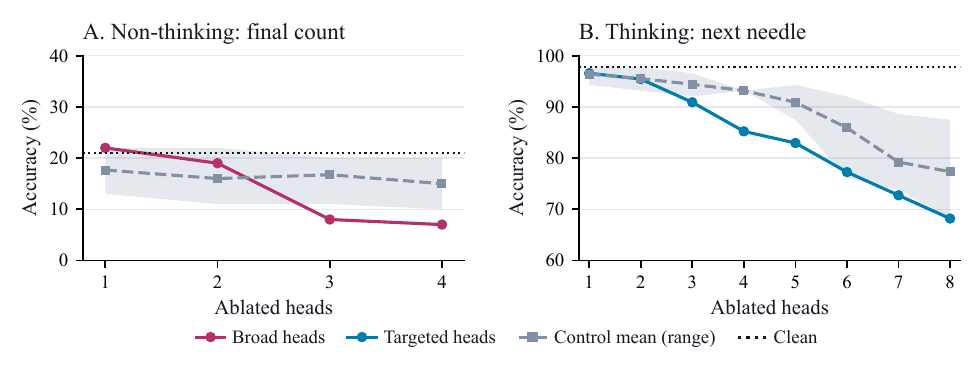}
\caption{\textbf{Retrieval head ablation.} A: final count accuracy on all 100 Non-thinking inputs, for Top-1--4 broad heads. B: next-needle accuracy on 88 eligible Thinking inputs, for Top-1--8 targeted heads. Lines show selected sets, control means, and clean baselines; shading spans the control sets. The axes each span 40 percentage points; B is truncated to 60--100\%.}
\label{fig:apx-retrieval}
\end{figure}

In Fig.~\ref{fig:apx-retrieval}, Non-thinking Top-4 ablation reduces final count accuracy to 7\%, versus a 15\% control mean. Thinking next-needle accuracy is 85.2\% versus 93.2\% at Top-4, and 68.2\% versus 77.3\% at Top-8. These ablations support the selected heads' roles in final count prediction and local needle retrieval, respectively.

\subsection{Count representations and causal interventions}
\label{app:synth-representations}
%\YZ{Define the measurements of representation compactness, internal counter, etc. Present detailed results. And say how the results support the claims in the main section.}

\paragraph{Counter states and answer states.}
Following the main-text terminology, counter states represent counting progress at prompt or trace items, and answer states support the final count prediction at the answer query. We decode running index $k$ from residual states at prompt position $s_k$ (Non-thinking) or the trace-character token $a_k$ (Thinking), and final count $N$ from the answer-query state at $q_A$. Here $k$ counts target occurrences encountered in the prompt or emitted in the trace, respectively. States are taken after each Transformer block, before final normalization, using teacher-forced prefixes. Each mode has 550 occurrence states and 100 answer states from ten inputs per count.

Each Thinking occurrence adds two trace tokens, so running index covaries with trace length and position. Counter-state analyses therefore retain these structural cues alongside item content.

\paragraph{PCA visualization.}

\begin{figure}[H]
\centering
\includegraphics[width=\linewidth]{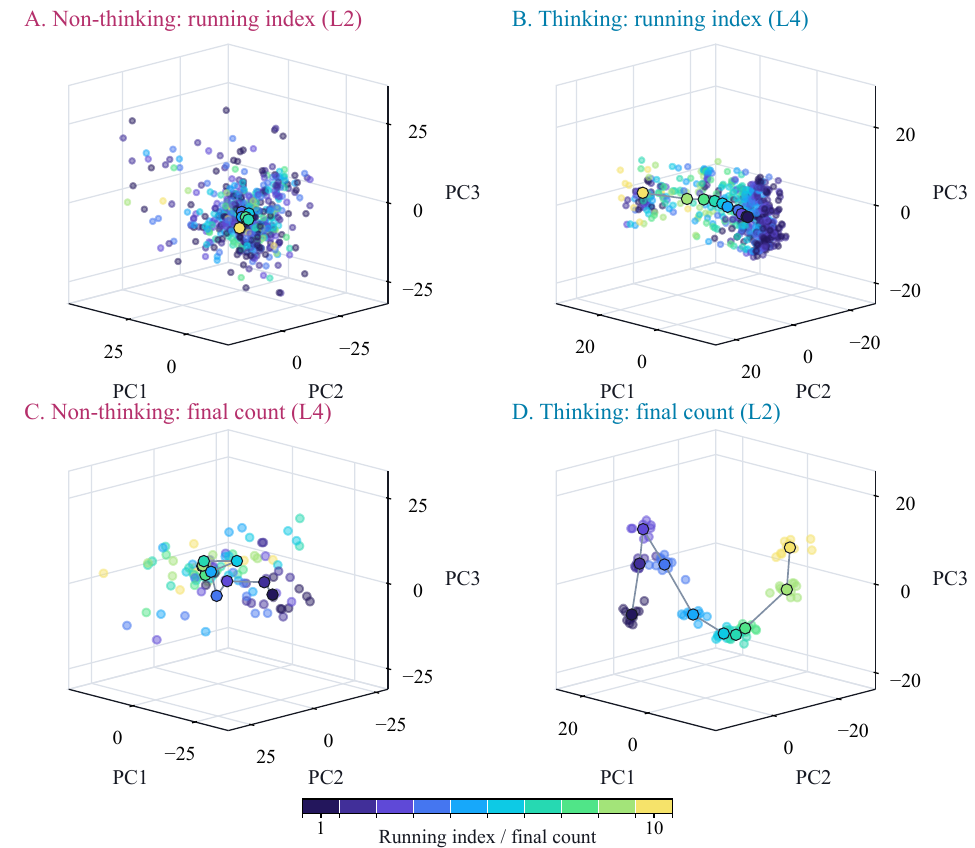}
\caption{\textbf{Count geometry.} Cross-validated NCC selects L2/L4/L4/L2 for A/B/C/D. Separate PCA projections show 550 running index states (A/B) and 100 answer states (C/D) per mode. Outlined class means are connected in label order. The three components explain 50.9/48.1/74.0/71.7\% of selection variance in A/B/C/D. All panels use the same camera view and equal coordinate units within each panel.}
\label{fig:apx-geometry}
\end{figure}

All four PCA panels use the layers selected by cross-validated NCC: L2/L4 for Non-thinking/Thinking running indices and L4/L2 for their final counts (Fig.~\ref{fig:apx-geometry}). Selection searches the four Transformer blocks and uses the grouped-validation rule below. A separate three-component PCA is fitted for each mode and endpoint. Selection-set standardization and PCA are applied unchanged to evaluation states, without whitening. Coordinates are not directly comparable across these independent projections.

\paragraph{Nearest-centroid decoding.}
For quantitative comparison, we standardize states and apply whitened 16-component PCA. A nearest-centroid classifier (NCC) assigns each transformed state $z$ to the closest class centroid $\mu_c$:
\begin{equation}
\widehat y(z)=\mathop{\mathrm{arg\,min}}_{c\in\{1,\ldots,10\}}\|z-\mu_c\|_2^2,
\qquad
\mathrm{BA}=\frac{1}{10}\sum_{c=1}^{10}\frac{1}{n_c}\sum_{i:y_i=c}\mathbf{1}\{\widehat y(z_i)=c\}.
\label{eq:synth-ncc}
\end{equation}
Here $n_c$ is the number of evaluation states in class $c$. Balanced accuracy weights all ten labels equally (chance: 10\%), accounting for the greater frequency of smaller running indices. The transform and centroids are fitted only on selection states. Marked layers maximize five-fold NCC balanced accuracy on the selection set, keeping each input block together; ties use logistic balanced accuracy and then earlier layers. Every fold fits its preprocessing and classifiers using training inputs only.

\begin{figure}[H]
\centering
\includegraphics[width=\linewidth]{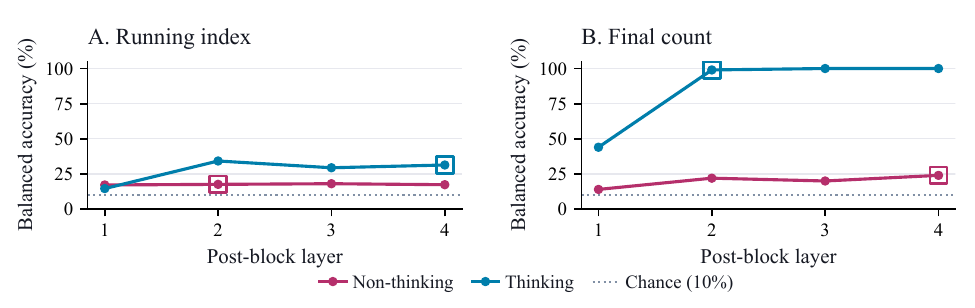}
\caption{\textbf{Count decoding across layers.} Balanced NCC accuracy for running index (A) and final count (B), using teacher-forced states. Squares mark layers selected by grouped cross-validation. Preprocessing and centroids are fitted only on selection inputs.}
\label{fig:apx-readability}
\end{figure}

At the NCC-selected layers, final count decoding reaches 99\% for Thinking (L2) and 24\% for Non-thinking (L4). Thinking running index accuracy ranges from 15\% to 34\% across Transformer blocks (Fig.~\ref{fig:apx-readability}). Thus, Thinking separation is strongest at the final answer; decoding alone does not establish that the model uses this information.

\paragraph{Sources of the next needle.}
\begin{figure}[H]
\centering
\includegraphics[width=\linewidth]{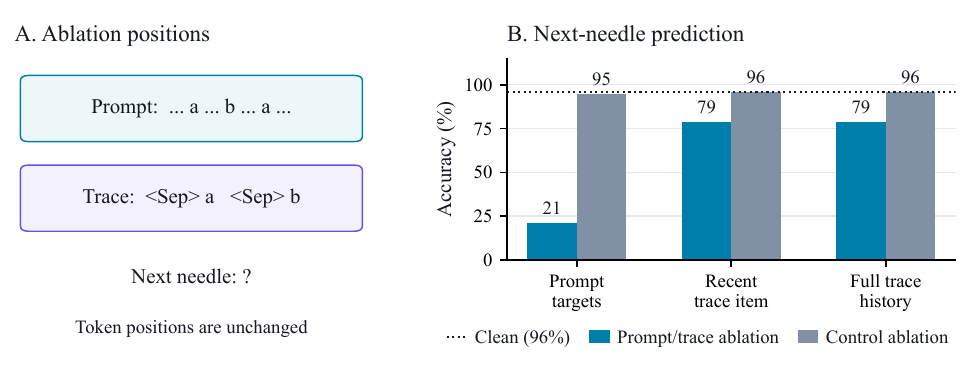}
\caption{\textbf{Sources of the next needle.} A: prompt and trace schematic. B: next-needle accuracy after prompt or trace ablation, with controls at non-target prompt positions. All 100 inputs are retained, including 30 without prior trace items.}
\label{fig:apx-progress-sources}
\end{figure}
At the query for $k=\max(1,\lfloor N/2\rfloor)$ under teacher forcing, we zero states at the selected prompt or trace positions at the embedding output and after every block, preserving token positions. We separately ablate prompt targets, the latest completed trace item, or the full preceding trace; controls ablate an equal number of non-target prompt positions. Next-needle accuracy falls from 96\% to 21\% after prompt-target ablation and to 79\% after either trace ablation (Fig.~\ref{fig:apx-progress-sources}). All 100 inputs remain in the evaluation, including 30 with no completed trace item. On the 70 inputs with prior trace items, accuracy falls from $66/70$ (94.3\%) to $49/70$ (70.0\%) under either trace ablation; the other 30 inputs are unchanged. The next needle prediction is therefore sensitive to both prompt targets and available trace history.

\paragraph{Counter-state transplantation.}
We copy the full separator and character states of source trace item $d$ into target trace item $k=d\pm1$, then continue generation without further patching. The order of target characters, the total count, and the target run's visible prefix remain fixed. Adding or removing two non-target characters from the source prompt aligns the transplanted token positions.

\begin{figure}[H]
\centering
\includegraphics[width=\linewidth]{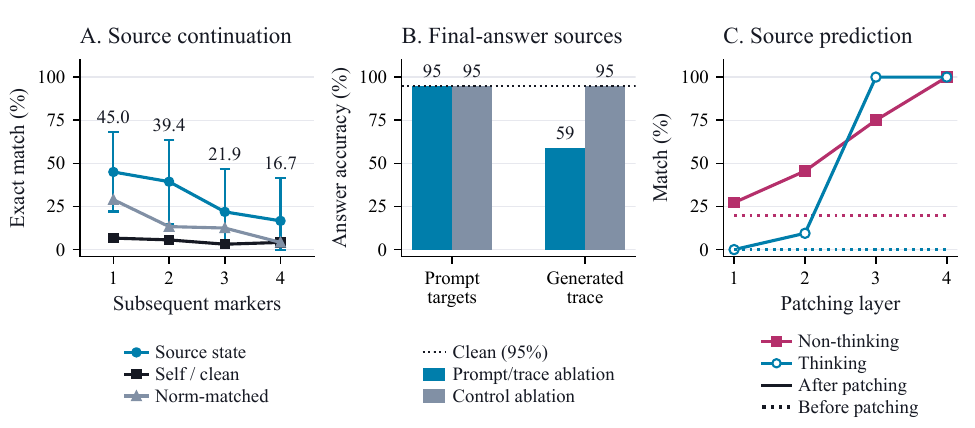}
\caption{\textbf{Trace continuation and final-answer interventions.} A: input-averaged source-continuation exact match for an L1 source-item transplant (circles), self patches/clean runs (squares), and norm-matched orthogonal controls (triangles). Conditions use the same eight inputs and 26/30/26/20 eligible pairs for lengths 1/2/3/4. Error bars on source transplants show 95\% percentile intervals from 10,000 input-bootstrap resamples; control curves show means. B: Thinking answer accuracy after prompt or trace ablation on 100 generated prefixes. Controls ablate equally many non-target prompt positions; the dotted line marks clean accuracy. C: agreement with the source run's clean prediction on 180 pairs using teacher-forced prefixes.}
\label{fig:apx-continuation}
\label{fig:apx-answer}
\end{figure}

On 20 selection inputs ($N=10$, $d=6$), we select L1 from L1--L3: require positive median shifts in the logit difference between the source and target continuations in both directions, then take the earliest layer reaching 95\% of the peak median input-averaged shift. Evaluation uses ten separate $N=10$ inputs at $d\in\{4,6,8\}$. For each length $h=1,\ldots,4$, we compare the first $h$ generated characters with the source continuation $(a_{d+1},\ldots,a_{d+h})$. A pair is eligible if both continuations have at least $h$ characters and the source sequence differs from the target continuation $(a_{k+1},\ldots,a_{k+h})$. Eight of the ten inputs contribute eligible pairs at each length. We average exact matches within input, then across eligible inputs; short or malformed outputs fail. Source-continuation exact match is 45.0\% for one character and 16.7\% for four characters (Fig.~\ref{fig:apx-continuation}A). Paired controls use the same eligible pairs: self patches restore the target's own states, and three orthogonal random perturbations match the full transplant's norm, averaging replicates within pair before averaging within input. For one character, clean and self-patched runs both yield 6.7\% source matches, compared with 29.0\% under the random control. The source transplant exceeds self patching by 38.3 percentage points (95\% paired input-bootstrap interval 10.0--66.0) and the random control by 16.0 points (4.2--29.0). Full-item transplantation transfers content and progress jointly. Wide intervals and different eligible pair sets across lengths limit conclusions about sustained continuation.

\paragraph{Sources of the final answer.}
On all 100 inputs, each model generates its prefix through \texttt{<Ans>}. We repeat the prompt and trace ablation protocol above at the final count query, targeting prompt occurrences or generated trace positions. Thinking accuracy falls from 95\% to 59\% after trace ablation and remains 95\% after prompt-target ablation (Fig.~\ref{fig:apx-continuation}B). Together, these prompt and trace ablations support prompt-based character retrieval and trace-based final count readout.

\paragraph{Answer state transplants.}
At each layer, we transplant the full answer-query state using teacher-forced prefixes and 180 directed pairs of adjacent counts within evaluation blocks. We score agreement with the source run's clean prediction. Before patching, this compares the clean target and source predictions: Thinking correctly predicts their distinct counts, giving 0\% agreement; Non-thinking gives the same prediction on 36 of 180 pairs, giving 20\%. All clean source predictions are valid count tokens.

At L3, with one block remaining, agreement rises from 20\% to 75\% for Non-thinking and from 0\% to 100\% for Thinking (Fig.~\ref{fig:apx-continuation}C). Both reach 100\% at L4, where the copied state directly feeds final normalization and the output projection. Non-thinking matches the source prompt's true count on only 17.2\% of these pairs, reflecting errors in the source predictions.

\subsection{Learning dynamics of retrieval and counting}
\label{app:synth-dynamics}

\paragraph{Attention roles and successor-like heads.}

\begin{figure}[H]
\centering
\includegraphics[width=\linewidth]{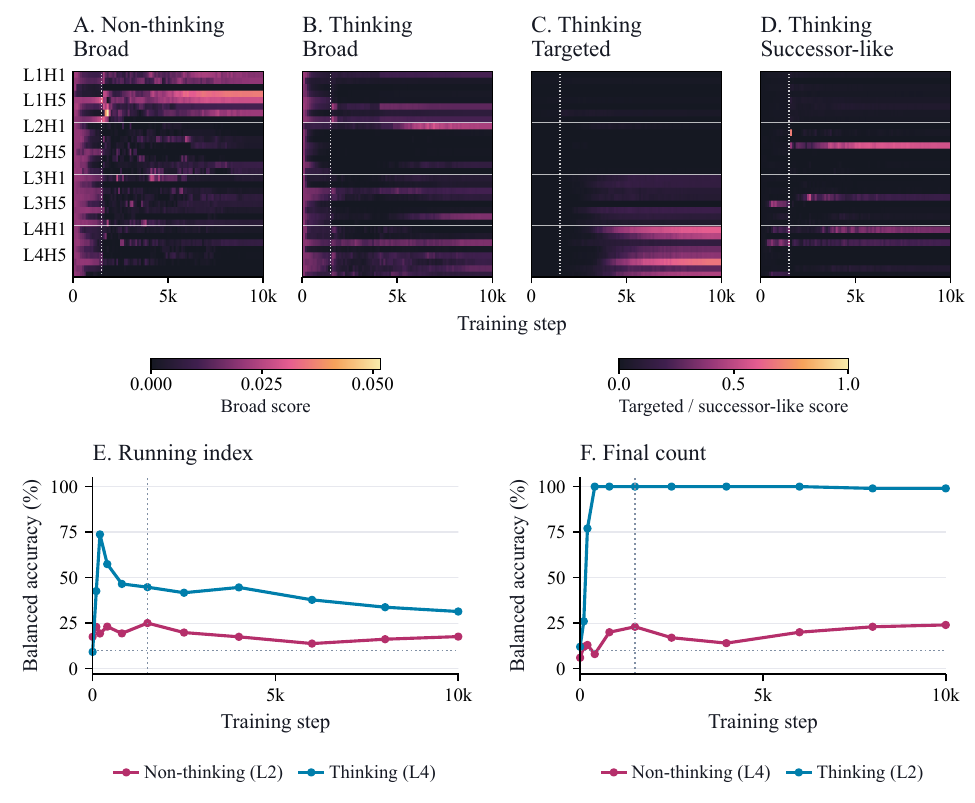}
\caption{\textbf{Attention and count decoding during training.} All panels use a linear training-step axis and the same 100 evaluation inputs. A--D: attention scores for all 32 heads across 101 checkpoints; rows retain head identity and white lines separate layers. A/B share the broad-score scale; C/D use a fixed 0--1 scale for targeted and successor-like attention. E/F: balanced NCC accuracy for running index and final count at eleven checkpoints. Layers are selected by final-checkpoint cross-validated NCC and remain fixed; transforms and centroids are fitted on selection states at each checkpoint. Vertical dotted lines mark the objective switch at step 1,500; horizontal dotted lines in E/F mark 10\% chance accuracy.}
\label{fig:apx-attention-dynamics}
\label{fig:apx-readability-dynamics}
\end{figure}

We track all 32 heads every 100 steps from initialization, using the same 100 teacher-forced evaluation inputs (Fig.~\ref{fig:apx-attention-dynamics}A--D). Alongside broad and targeted retrieval, we measure \emph{successor-like} attention from each trace character at $r_k=q_k+1$ to its preceding separator at $q_k$:
\begin{equation}
U_h=\frac{1}{N}\sum_{k=1}^{N}A_h(r_k,q_k).
\label{eq:synth-successor-score}
\end{equation}
Scores are averaged equally across inputs. Non-thinking develops broad retrieval in shallow layers, while Thinking develops targeted retrieval in deeper layers. Local successor-like attention strengthens earlier: at step 400, the largest $U_h$ across Thinking heads is 0.267, while the largest $T_h$ is 0.006.

\citet{gould2023successor} identify successor heads whose outputs promote the next member of an ordered sequence, such as numbers or days of the week. Our term \emph{successor-like} describes the local attention pattern in $U_h$, without assuming this increment function. The preceding separator is the query used to retrieve the current character; attending back to it may carry retrieval-query information forward to the completed item's state.

At the final checkpoint, L2H4 has the highest $U_h$ on the 200 selection inputs. In a separate assay on 100 evaluation inputs, we zero its output at generated trace-character positions immediately following \texttt{<Sep>}. This position-restricted intervention differs from the sustained retrieval-head ablations above. Final count accuracy falls from 95\% to 45\%, versus 79--94\% for three same-layer controls (L2H1, L2H5, and L2H8). Exact-trace accuracy falls from 87\% to 39\%, with all outputs remaining valid in format. These results support L2H4's contribution to trace processing and counting behavior.

\paragraph{Decoding and generation.}
We evaluate NCC at steps $0,\allowbreak100,\allowbreak200,\allowbreak400,\allowbreak800,\allowbreak1500,\allowbreak2500,\allowbreak4000,\allowbreak6000,\allowbreak8000,\allowbreak10000$ (Fig.~\ref{fig:apx-readability-dynamics}E/F). Layers remain fixed at the final-checkpoint cross-validated NCC selections: L2/L4 for Non-thinking/Thinking running indices and L4/L2 for their final counts. The grouped cross-validation rule and tie-breaks are the same as in Fig.~\ref{fig:apx-readability}; PCA displays use these same layers. Earlier checkpoints do not reselect layers. Transforms and centroids are refitted on selection states at each checkpoint.

At step 400, Thinking final count decoding reaches 100\% with a correct teacher-forced trace, while free-generation count accuracy is 12\% and exact-trace accuracy is 2\%. Thus, the final count becomes readable given a correct trace before the model can reliably generate that trace. Intermediate running index decoding follows a different trajectory, falling from 73.7\% at step 200 to 31.4\% at the final checkpoint. Early decoding may reflect regular trace structure before accurate retrieval develops. Successor-like attention is still weak at step 200 (maximum $U_h=0.025$), so its role in this peak remains unverified.

\subsection{Relation to the main experiments}
\label{app:synth-main-relation}

\paragraph{Retrieval mechanisms.}
The models trained from scratch reproduce broad retrieval in Non-thinking and targeted retrieval in Thinking, measured by the same scores as in Sections~\ref{sec:non-thinking} and~\ref{sec:thinking} (Fig.~\ref{fig:apx-head-scores}). Selected-head ablations affect final count prediction and next-needle retrieval, respectively (Fig.~\ref{fig:apx-retrieval}). The synthetic Thinking assay supports local retrieval; the main experiments additionally test final count effects.

\paragraph{Counter states and continuation.}
Full-item state transplants increase source-continuation adoption relative to paired self and norm-matched controls at short horizons (Fig.~\ref{fig:apx-continuation}A), paralleling the progress-transfer tests in Appendix~\ref{app:cot-progress}. These results support a role for trace-item states in guiding subsequent enumeration; the full-item patch transfers content and progress jointly. 

\paragraph{Final-answer readout.}
Thinking improves final count decoding, while its intermediate running index states remain weakly separated (Fig.~\ref{fig:apx-readability}). The synthetic geometric evidence is therefore strongest at the answer query. With a generated trace available, final answering depends on trace states, while prompt-target ablation leaves accuracy unchanged (Fig.~\ref{fig:apx-continuation}B), matching Appendix~\ref{app:cot-blanking}. Late answer state transplants steer both modes toward the source run's prediction (Fig.~\ref{fig:apx-continuation}C), consistent with Non-thinking consolidation and Thinking readout (Appendices~\ref{app:nonthinking-causal} and~\ref{app:cot-answer-patching}).

% !TEX root = ../main.tex
\section{Further details for additional tasks}
\label{app:additional-task}

We extend the counting experiments to retrieval of the $k$th record and counting within a specified category, measuring final-answer accuracy and testing heads selected separately for each task.

\subsection{Experimental design}
\label{app:additional-task-prompts}
\paragraph{Tasks and inputs.}
Each passage contains $J=10$ records $r_j=(e_j,v_j,c_j)$ in passage order, with entity $e_j$, numeric score $v_j$, and category $c_j$. The retrieval task asks for $(e_k,v_k)$, varying $k=1,\ldots,10$ on the same city-record passage within each seed. For category counting, $N=\sum_{j=1}^{J}\mathbf{1}\{c_j=c\}$ denotes the number of records in the queried category $c$. Each seed supplies five passages with 1, 3, 5, 7, or 9 city records and the remaining records assigned to flowers; both categories are queried on every passage. For example, a passage with three city records has city and flower counts of 3 and 7. This construction preserves the main experiment's background, record order, scores, and passage wrapper, changing only the category word and entity name where needed.

\paragraph{Models and generation.}
We evaluate Qwen3-8B and Gemma-4-E4B without additional training, using bfloat16 weights, SDPA generation, and greedy decoding. Non-thinking permits 64 new tokens; Thinking permits 4,096, with reasoning enabled and no forced enumeration or first-list cutoff. Each task contains 300 inputs from 30 seeds: 200 selection inputs (seeds 1234--1253) and 100 evaluation inputs (1254--1263). Inputs are shared across models and modes, giving 2,400 natural outputs. Prompts contain approximately 10,000 tokens. Only selection inputs determine the head rankings.

As in Appendix~\ref{app:behavioral-prompts}, each task has a complete user-message template and two response-instruction variants. Replace \promptvar{passage} with the experimental passage and \promptvar{response\_instruction} with the chosen mode's block. Box titles and role labels are annotations; angle-bracketed output placeholders are literal.

\paragraph{Retrieval of the $k$th record.}
Set \promptvar{k} to the requested position ($1,\ldots,10$). Non-thinking uses the \texttt{Needle:} assistant prefill after chat-template rendering; Thinking uses no prefill.

% Source: additional_experiments/protocol.py::user_prompt; frozen kth_uniform_20260906_v2 prompts.
\begin{promptbox}[unbreakable]{Retrieval of the $k$th record: user template}
\setlength{\parskip}{1.5pt}
\promptlabel{User message}
% Prompt fragment: H-kth-user
You will need to find a specified city-score audit record in passage order.\par
A city-score audit record names one city and gives that city's numeric score.\par
\smallskip
<passage>\par
\{passage\}\par
</passage>\par
\smallskip
Which city-score audit record is number \{k\} in passage order?\par
Count occurrences from the beginning of the passage, starting at 1.\par
Return the actual city name and numeric score of that record.\par
\{response\_instruction\}\par
% End prompt fragment: H-kth-user
\end{promptbox}

\begin{promptbox}[unbreakable]{Retrieval of the $k$th record: response instructions}
\setlength{\parskip}{1.5pt}
\promptlabel{Non-thinking}
% Prompt fragment: H-kth-nonthinking
Do not explain, reason aloud, quote, or list any records.\par
Use no spaces around the colon or vertical bar.\par
Your entire response must be exactly one line:\par
Needle:<city>|<integer>\par
% End prompt fragment: H-kth-nonthinking
\promptdivider
\promptlabel{Thinking}
% Prompt fragment: H-kth-cot
Reason concisely without repeating or restarting.\par
Stop as soon as you identify that record, then output exactly one line:\par
Needle:<city>|<integer>\par
% End prompt fragment: H-kth-cot
\end{promptbox}

\paragraph{Counting a specified category.}
Set \promptvar{category} to \texttt{city} or \texttt{flower}. Non-thinking uses the \texttt{Total:} assistant prefill after chat-template rendering; Thinking uses no prefill.

% Source: additional_experiments/category_trial.py::prompt; frozen category_full_20260906 prompts.
\begin{promptbox}[unbreakable]{Category counting: user template}
\setlength{\parskip}{1.5pt}
\promptlabel{User message}
% Prompt fragment: H-category-user
You will need to count only \{category\}-score audit records in the passage below.\par
A city-score audit record names one city and gives its numeric score.\par
A flower-score audit record names one flower and gives its numeric score.\par
\smallskip
<passage>\par
\{passage\}\par
</passage>\par
\smallskip
How many \{category\}-score audit records are in the passage?\par
Count only matching audit records. Ignore audit records of the other category and ordinary passage text.\par
\{response\_instruction\}\par
% End prompt fragment: H-category-user
\end{promptbox}

\begin{promptbox}[unbreakable]{Category counting: response instructions}
\setlength{\parskip}{1.5pt}
\promptlabel{Non-thinking}
% Prompt fragment: H-category-nonthinking
Do not explain, reason aloud, quote, or list any records.\par
Write the count using ordinary decimal digits, with no space after the colon.\par
Your entire response must be exactly one line:\par
Total:<integer>\par
% End prompt fragment: H-category-nonthinking
\promptdivider
\promptlabel{Thinking}
% Prompt fragment: H-category-cot
Reason concisely without repeating or restarting.\par
Stop as soon as you determine the count, then output exactly one line:\par
Total:<integer>\par
% End prompt fragment: H-category-cot
\end{promptbox}

For Non-thinking, we prepend the saved answer prefix and remove end tokens. For Thinking, we separate reasoning from the final response using \texttt{</think>} for Qwen and \texttt{<channel|>} for Gemma, then remove supported response wrappers and end tokens; already-separated final responses are also accepted. The entire final response must match \texttt{Needle:<city>|<digits>} or \texttt{Total:<digits>}; surrounding whitespace and whitespace after separators are allowed. City names are trimmed and matched case-sensitively, and digit strings are compared as integers. Extra prose, missing fields, or a detected but unclosed reasoning channel yield an unparsed answer and count as errors. Capped outputs remain in the denominator and follow the same rule.

Here we average accuracy within data-generation seeds, then equally across seeds (30 for natural behavior; ten for interventions). We report 95\% percentile intervals from 20,000 bootstrap resamples of seeds, paired across conditions for intervention contrasts.

\subsection{Behavioral comparison between the two modes}
\label{app:additional-behavior}

\begin{figure}[H]
\centering
\includegraphics[width=\linewidth]{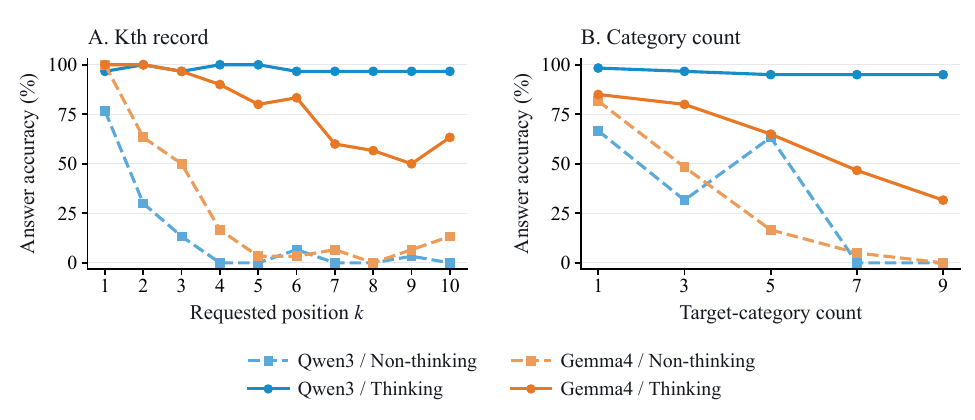}
\caption{\textbf{Final-answer accuracy by position and category count.} A: requested record position (30 inputs per point). B: true target-category count, pooling city and flower questions (60 per point). Blue denotes Qwen and orange Gemma; solid circles denote Thinking and dashed squares Non-thinking. Aggregate 95\% CIs appear in Table~\ref{tab:additional-behavior}.}
\label{fig:additional-behavior}
\end{figure}

Thinking achieves higher final-answer accuracy for both models on both tasks (Table~\ref{tab:additional-behavior}). Across requested positions and category counts, Qwen's Thinking accuracy remains high, whereas Gemma's generally declines at later positions and larger counts (Fig.~\ref{fig:additional-behavior}).

\begin{table}[H]
\centering
\caption{\textbf{Natural final-answer accuracy (\%).} Each condition includes all 300 inputs from 30 seeds. Brackets give 95\% confidence intervals (CIs).}
\label{tab:additional-behavior}
\par\small
\renewcommand{\arraystretch}{1.08}
\setlength{\tabcolsep}{4pt}
\begin{tabularx}{\linewidth}{@{}l l >{\centering\arraybackslash}X >{\centering\arraybackslash}X@{}}
\toprule
Task & Model & Non-thinking & Thinking\\
\midrule
Kth record & Qwen & 13.00 [11.33, 15.00] & 97.67 [94.67, 100.00]\\
Kth record & Gemma & 26.33 [23.00, 29.67] & 78.00 [70.00, 85.67]\\
Category count & Qwen & 32.33 [28.33, 36.33] & 96.00 [93.67, 98.33]\\
Category count & Gemma & 30.33 [27.00, 34.00] & 61.67 [54.67, 68.67]\\
\bottomrule
\end{tabularx}\par
\end{table}

Gemma's Thinking results include 12 unparsed retrieval answers and 16 unparsed category answers, with one retrieval output reaching the token limit. All other conditions have no unparsed or capped outputs. These comparisons include each mode's instructions and generation budget.

\subsection{Retrieval-head distributions and causal tests}
\label{app:additional-retrieval}

We next test whether task-specific retrieval heads contribute to these behaviors. Broad-head interventions measure Non-thinking final-answer accuracy; targeted-head interventions measure next-entity accuracy during Thinking. Head selection and intervention evaluation use disjoint inputs.

\paragraph{Head identification and parsing.}
For targeted retrieval, we first parse the reasoning channel by matching input city and flower names case-insensitively. A rank parser links mentions to local numeric, ordinal, or running-count markers, splits at restarts from 1, and retains the longest contiguous $1,\ldots,M$ sequence (earliest on ties). Repeated entities are retained if their ranks advance. A structural parser recognizes terminated numbered, bulleted, ordinal, or supported record-sentence lists; its sequence replaces a rank sequence only by extending the exact entity prefix, or is used when no rank sequence exists. If both fail, we order the first mention of each distinct entity whose registered score occurs within the next 96 characters. These inferred positions index mentions without establishing count updates. Neither the requested answer nor final-answer correctness determines the selected sequence.

We register the earlier middle eligible transition between consecutive parsed records. For Qwen, $q_*$ ends the prefix through a pre-entity rank marker when available; otherwise, and for Gemma, it is $P_0$. The preceding item ends at the later of its entity-bearing sentence/line and rank marker, or at its structural/score-supported endpoint. We align to the first original-token boundary at or after this endpoint, crossing only whitespace or punctuation and remaining before the next entity; $P_0$ is the token immediately before that boundary. Targeted score $T_h$ (Sec.~\ref{sec:targeted-retrieval-score}) measures attention at $q_*$ to the next parsed entity's full input-record span $\mathcal S_*$, regardless of the requested rank or category. For Non-thinking, the broad score (Sec.~\ref{sec:broad-retrieval-score}) uses the final-answer query $q_A$ and all $J$ input-record spans: $B_h=M_h\exp(H_h)/J$, with sums over $j=1,\ldots,J$. For numerical stability, we use $p_{hj}=m_{hj}/(M_h+\epsilon)$, $\log(p_{hj}+\epsilon)$, $\epsilon=10^{-12}$, and $B_h=0$ for $M_h\leq\epsilon$.

\begin{figure}[H]
\centering
\includegraphics[width=\linewidth]{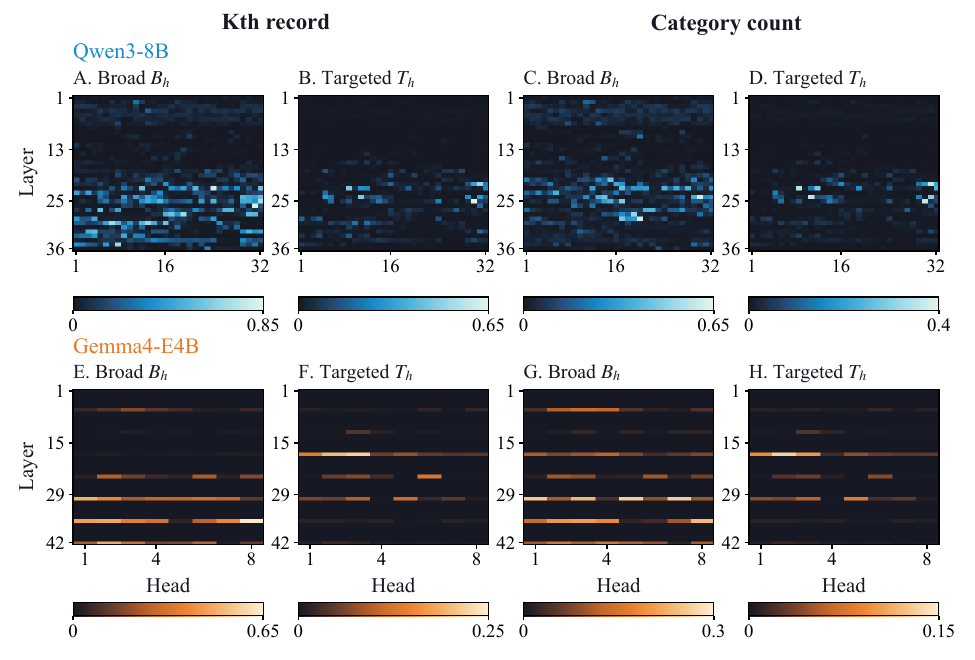}
\caption{\textbf{Head scores for both additional tasks.} Task groups pair broad ($B_h$, Non-thinking) and targeted ($T_h$, Thinking) scores; rows show Qwen (blue) and Gemma (orange). Cells represent heads, with one-based layer and head indices. Each panel has an independent, unclipped linear scale from zero to its labeled maximum; compare numerical scores using the colorbars.}
\label{fig:additional-scores}
\end{figure}

We rank heads across all layers by their mean scores on the selection inputs within each task, model, and scoring condition, breaking ties by layer and head index (Fig.~\ref{fig:additional-scores}). Broad retrieval uses all inputs in each split. Targeted retrieval requires an aligned transition: selection/held-out evaluation counts are 167/83 and 149/79 for Qwen on retrieval and category counting, and 190/94 and 180/90 for Gemma. Inputs without a transition remain in the natural evaluation. Interventions include all ten evaluation seeds, retain clean errors, and keep eligible inputs fixed across $K$.

\paragraph{Ablation protocol and evaluation.}
Selected ablation zeros the frozen Top-$K$ heads' output slices before the attention output projection. Broad ablation acts only at $q_A$ in one prefix forward pass, followed by at most 64 new tokens. Targeted ablation acts at $q_*$ and subsequent decoding positions for at most 256 tokens. Clean uses the same prefixes and budgets.

For the targeted outcome, the continuation parser compares the earliest recognized entity with the registered next entity, ignoring case. Recognition uses input names and generic record forms (``received a score,'' ``with a score,'' or name--score pairs), including city/flower audit forms that allow unlisted names. An earlier wrong entity is an error even if the target appears later. If none is recognized, a token fallback requires a saved target at continuation offset zero and a continuation beginning with its complete original entity-token sequence. Only entity identity is scored.

Random accuracy is averaged over three control sets matching Selected's head count per layer. Heads are sampled without replacement from unselected heads whenever possible. Only broad kth-record ablation at the largest $K$ requires overlap: two heads in one layer per model and control set. Qwen's broad grid is $K\in\{1,2,4,8,16,32,64,128\}$ and its targeted grid is $\{32,64,80,96,112,128\}$; Gemma uses $\{1,2,4,6,8\}$ for both. As in the main experiment, we report the full scans of Selected and Random accuracies.

\begin{figure}[H]
\centering
\includegraphics[width=\linewidth]{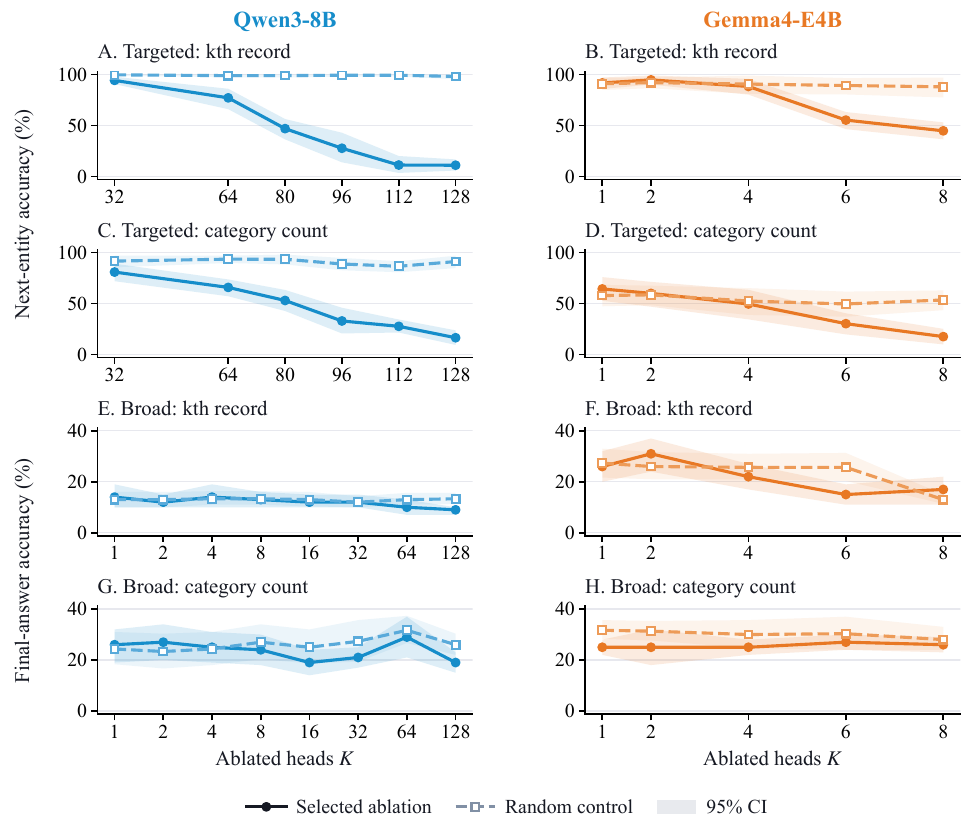}
\caption{\textbf{Top-$K$ head ablation.} Absolute accuracies for Selected (solid circles) and Random (dashed open squares; mean of three banks). A--D: Thinking targeted ablation, measuring next-entity accuracy. E--H: Non-thinking broad ablation, measuring final-answer accuracy. Columns show Qwen (blue) and Gemma (orange); rows separate tasks. Shading shows pointwise 95\% CIs. Qwen's broad axes are logarithmic. Clean accuracies for Qwen/Gemma are 100.00/97.00\% for kth-record retrieval and 94.78/69.36\% for category counting in the targeted assay; the corresponding broad-assay accuracies are 13.00/26.00\% and 25.00/29.00\%.}
\label{fig:additional-ablation}
\end{figure}

\paragraph{Effects of head ablation.}
Task-ranked targeted heads contribute causally to next-entity generation in both additional tasks (Fig.~\ref{fig:additional-ablation}A--D). Qwen's Selected accuracy decreases monotonically over the scanned $K$ values, while Gemma shows a pronounced separation from Random at $K=6$ and $8$, with mixed effects at smaller $K$. At the largest tested $K$ (128 for Qwen; 8 for Gemma), Selected versus Random accuracy is 10.9\% versus 98.1\% (Qwen) and 44.7\% versus 87.9\% (Gemma) for kth-record retrieval; for category counting, the corresponding accuracies are 16.5\% versus 91.3\% and 17.6\% versus 53.6\%. The layer-matched controls show that the impairment exceeds the effect of removing the same number of heads from the same layers.

Broad-head ablation also reduces final-answer accuracy at some $K$, but its separation from Random is weaker and less consistent across the scan (Fig.~\ref{fig:additional-ablation}E--H). Low Clean accuracy (13--29\%) leaves limited room for further declines. Additional ordering or category-selection demands may introduce errors beyond retrieval, although their role in weakening the ablation effect remains untested. Differences in outcomes, eligible inputs, and intervention duration preclude a direct comparison of head importance between the two assays.

\subsection{Relation to the main experiments}
\label{app:additional-main-relation}

Thinking improves both models' final-answer accuracy on both tasks (Table~\ref{tab:additional-behavior}). At larger tested $K$, targeted-head ablation impairs next-entity accuracy more than layer-matched random ablation (Fig.~\ref{fig:additional-ablation}A--D), supporting the Thinking retrieve stage (Sec.~\ref{sec:thinking}; Appendix~\ref{app:cot-retrieval}).

For Non-thinking, broad-head ablation reduces final-answer accuracy at some $K$, with weaker, less consistent separation from Random (Fig.~\ref{fig:additional-ablation}E--H). Low baseline accuracy limits observable declines; these effects remain compatible with broad retrieval (Sec.~\ref{sec:non-thinking}; Appendix~\ref{app:nonthinking-retrieval}).

\end{document}